\documentclass{article}

\PassOptionsToPackage{numbers, compress}{natbib}
\usepackage{fullpage}

\usepackage[utf8]{inputenc} 
\usepackage[T1]{fontenc}    
\usepackage{hyperref}       
\usepackage{url}            
\usepackage{booktabs}       
\usepackage{amsfonts}       
\usepackage{nicefrac}       
\usepackage{microtype}      

\usepackage{amsmath, amsfonts, amssymb, amsthm, amsbsy, amscd, bm, bbm,mathrsfs}
\RequirePackage[numbers]{natbib}
\usepackage{cleveref}       

\usepackage{epsfig}
\usepackage{subfig}
\usepackage{epstopdf}
\usepackage{epic,rotating,color}
\usepackage{geometry}
\usepackage{multicol}
\usepackage{pdfpages}
\usepackage{wrapfig}
\usepackage{cases}
\usepackage{color}
\usepackage[normalem]{ulem}
\usepackage{ctable} 
\usepackage{tabularx}
\usepackage{caption}
\usepackage{mathtools}
\usepackage{enumitem}
\usepackage{float}
\usepackage{csquotes}
\usepackage{rotating}
\usepackage{csquotes}
\usepackage{epsf}
\usepackage{fancyhdr}
\usepackage{graphics}
\graphicspath{{figs/}}

\usepackage{color}
\usepackage{bbm}
\usepackage{hhline}
\usepackage{algorithm}
\usepackage{algorithmicx}
\usepackage[noend]{algpseudocode}

\newcommand{\norm}[1]{\left\lVert#1\right\rVert}

\usepackage{color}
\usepackage{comment}
\usepackage{ulem}
\usepackage{soul}
\definecolor{applegreen}{rgb}{0.55, 0.71, 0.0}
\definecolor{amber}{rgb}{1.0, 0.49, 0.0}

\title{Differentially Private Bayesian Neural Networks on Accuracy, Privacy and Reliability}

\author{Qiyiwen Zhang\thanks{Equal contribution} \and Zhiqi Bu$^*$ \and Kan Chen \and Qi Long}

\newcommand{\w}{\bm w}
\newcommand{\x}{\bm x}

\newtheorem{theorem}{Theorem}
\newtheorem{othertheorem}{othertheorem}[section]
\newtheorem{lemma}[othertheorem]{Lemma}

\theoremstyle{definition}
\newtheorem{definition}[othertheorem]{Definition}
\theoremstyle{remark}
\newtheorem{remark}[othertheorem]{Remark}
\newtheorem{assumption}[othertheorem]{Assumption}
\theoremstyle{definition}

\date{}
\begin{document}

\maketitle
\vspace{-0.95cm}
\begin{center}
    \large
    University of Pennsylvania
    \\
	\texttt{\{zhangqi, zbu, kanchen, qlong\}@upenn.edu}
\end{center}

\begin{abstract}


Bayesian neural network (BNN) allows for uncertainty quantification in prediction, offering an advantage over regular neural networks that has not been explored in the differential privacy (DP) framework. We fill this important gap by leveraging recent development in Bayesian deep learning and privacy accounting to offer a more precise analysis of the trade-off between privacy and accuracy in BNN. We propose three DP-BNNs that characterize the weight uncertainty for the same network architecture in distinct ways, namely DP-SGLD (via the noisy gradient method), DP-BBP (via changing the parameters of interest) and DP-MC Dropout (via the model architecture). Interestingly, we show a new equivalence between DP-SGD and DP-SGLD, implying that some non-Bayesian DP training naturally allows for uncertainty quantification. However, the hyperparameters such as learning rate and batch size, can have different or even opposite effects in DP-SGD and DP-SGLD.

Extensive experiments are conducted to compare DP-BNNs, in terms of privacy guarantee, prediction accuracy, uncertainty quantification, calibration, computation speed, and generalizability to network architecture. As a result, we observe a new tradeoff between the privacy and the reliability. When compared to non-DP and non-Bayesian approaches, DP-SGLD is remarkably accurate under strong privacy guarantee, demonstrating the great potential of DP-BNN in real-world tasks.

\end{abstract}

\section{Introduction}


Deep learning has exhibited impressively strong performance in a wide range of classification and regression tasks. However, standard deep neural networks do not capture the model uncertainty and fail to provide the information available in statistical inference, which is crucial to many applications where poor decisions are accompanied with high risks. As a consequence, neural networks are prone to overfitting and being overconfident about their prediction, reducing their generalization capability and more importantly, their reliability. From this perspective, Bayesian neural network (BNN) \cite{mackay1992practical,mackay1995probable,buntine1991bayesian,neal2012bayesian} is highly desirable and useful as it characterizes the model's uncertainty, which on one hand offers a reliable and calibrated prediction interval that indicates the model's confidence \cite{welling2011bayesian,graves2011practical,blundell2015weight,kuleshov2018accurate,maronas2018offline}, and on the other hand reduces the prediction error through the model averaging over multiple weights sampled from the learned posterior distribution. For example, networks with the dropout \cite{srivastava2014dropout} can be viewed as a Bayesian neural network by \cite{gal2016dropout}; the dropout improves the accuracy from $57\%$ \cite{zeiler2013stochastic} to $63\%$ \cite{srivastava2014dropout} on CIFAR100 image dataset and $69.0\%$ to $70.4\%$ on Reuters RCV1 text dataset \cite{srivastava2014dropout}. In another example, on a genetics dataset where the task is to predict the occurrence probability of three alternative-splicing-related events based on RNA features. The performance of `Code Quality' (a measure of the KL divergence between the target and the predicted probability distributions) can be improved from 440 on standard network to 623 on BNN \cite{xiong2011bayesian}. 

In a long line of research, much effort has been devoted to making BNNs accurate and scalable. These approaches can be categorized into three main classes: (i) by introducing random noise into gradient methods (e.g. SG-MCMC \cite{wang2015privacy}) to quantify the weight uncertainty; (ii) by considering each weight as a distribution, instead of a point estimate, so that the uncertainty is described inside the distribution; (iii) by introducing randomness on the network architecture (e.g. the dropout) that leads to a stochastic training process whose variability characterizes the model's uncertainty. To be more specific, we will discuss these methods including the Stochastic Gradient Langevin Descent (SGLD) \cite{li2016preconditioned}, the Bayes By Backprop (BBP) \cite{blundell2015weight} and the Monte Carlo Dropout (MC Dropout) \cite{gal2016dropout}.

Another natural yet urgent concern on the standard neural networks is the privacy risk. The use of sensitive datasets that contain information from individuals, including medical records, email contents, financial statements, and photos, has incurred serious risk of privacy violation. For example, using a person's ZIP code, date of birth, and gender from Public Use Microdata Sample (PUMS), an anonymously de-identified health data, allows an attacker to re-identify Governor William Weld \cite{sweeney1997weaving,rocher2019estimating}. For another example, the sale of Facebook user data to Cambridge Analytica \cite{cadwalladr2018revealed} leads to the \$5 billion fine to the Federal Trade Commission for its privacy leakage. 
As a gold standard to protect the privacy, the differential privacy (DP) has been introduced by \cite{dwork2006calibrating} and widely applied to deep learning \cite{abadi2016deep,bu2020deep,bagdasaryan2019differential,phan2017adaptive,ryffel2018generic,bu2021fast,xie2018differentially}, due to its mathematical rigor. 

\begin{wrapfigure}{r}{0.45\textwidth}
\centering
\vspace{-0.2cm}
    \includegraphics[width=0.4\textwidth]{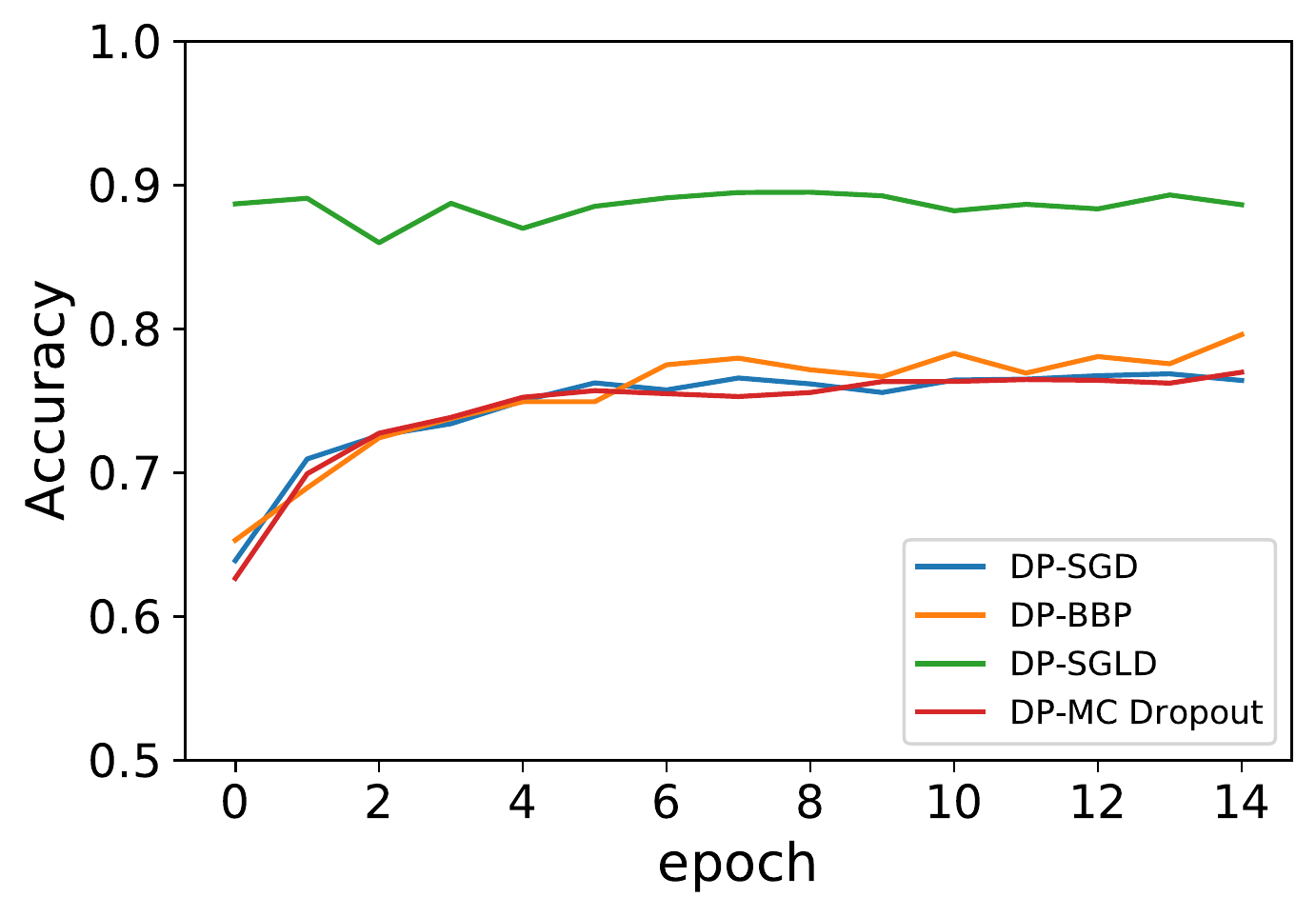}    
    \vspace{-0.35cm}
    \caption{Accuracy of DP gradient methods with Gaussian priors on MNIST. Settings in \Cref{sec:classification}.}
    \label{accuracy_epoch}
    \vspace{-0.5cm}
\end{wrapfigure}
Although both uncertainty quantification and privacy guarantee have drawn increasing attention, most existing work studied these two perspectives separately. Previous arts either studied DP Bayesian linear models \cite{wang2015privacy,zhang2016differential} or studied DP-BNN using SGLD but only for the accuracy measure without uncertainty quantification. In short, to the best of our knowledge, no existing deep learning models have equipped with the differential privacy and the Bayesian uncertainty quantification simultaneously.

\textbf{Our Contributions\quad}
To bridge this important gap, we leverage state-of-the-art Bayesian neural networks \cite{li2016preconditioned,gal2016dropout,blundell2015weight} and privacy accounting methods \cite{abadi2016deep,canonne2020discrete,bu2020deep} to accomplish the following goals:
\begin{itemize}
    \item[1.] We propose three distinct DP-BNNs that all use the DP-SGD (stochastic gradient descent) but characterize the weight uncertainty in distinct ways, namely DP-SGLD (via the noisy gradient method), DP-BBP (via changing the parameters of interest), and DP-MC Dropout (via the model architecture).
    \item[2.] In particular, we establish the precise connection between the Bayesian gradient method, DP-SGLD and the non-Bayesian method, DP-SGD. Through a rigorous analysis, we show that DP-SGLD is a sub-class of DP-SGD yet the training hyperparameters (e.g. learning rate and batch size) have very different impacts on the performance of these two methods.
    \item[3.] We empirically evaluate DP-BNNs through the classification and regression tasks, under various measures: for example, DP-SGLD can substantially outperform others in terms of prediction accuracy and uncertainty qualification. But unlike DP-BBP, the DP-SGLD offers no analytic posterior distribution and thus incurs high storage memory (less scalable to large models). Further pros and cons of each method are extensively discussed in \Cref{sec4: dp and bnn}.
\end{itemize}


\section{Differentially Private Neural Networks} \label{sec2:dp framework}
In this work, we consider $(\epsilon,\delta)$-DP and also use $\mu$-GDP as a tool to compose the privacy loss $\epsilon$ iteratively. We first introduce the definition of $(\epsilon,\delta)$-DP in \cite{dwork2014algorithmic}.

\begin{definition}
A randomized algorithm $M$ is $ (\varepsilon, \delta) $\textit{-differentially private} (DP) if for any pair of datasets $ S, S^{\prime} $ that differ in a single sample, and for any event $ E $,
\begin{align}
 \mathbb{P}[M(S) \in E] \leqslant \mathrm{e}^{\varepsilon} \mathbb{P}\left[M\left(S^{\prime}\right) \in E\right]+\delta.
\end{align}
\end{definition}

A common approach to learn a DP neural network (NN) is to use DP gradient methods, such as DP-SGD (see \Cref{alg:DPSGD}; possibly with the momentum and weight decay) and DP-Adam \cite{bu2020deep}, to update the neural network parameters, i.e. weights and biases. In order to guarantee the privacy, DP gradient methods differ from its non-private counterparts in two steps. For one, the gradients are \textit{clipped} on a per-sample basis, by a pre-defined clipping norm $C$. This is to ensure the sum of gradients has a bounded \textit{sensitivity} to data points (this concept is to be defined in \Cref{app:DP background}). We note that in non-neural-network training, DP gradient methods may apply without the clipping, for instance, DP-SGLD in \cite{wang2015privacy} requires no clipping and is thus different from our DP-SGLD in \Cref{alg:DPSGLD} (also our DP-SGLD need not to modify the noise scale). For the other, some level of random Gaussian noises are added to the clipped gradient at each iteration. This is known as the \textit{Gaussian mechanism} which has been rigorously shown to be DP by \cite[Theorem 3.22]{dwork2014algorithmic}.

In the training of neural networks, the Gaussian mechanism is applied multiple times and the privacy loss $\epsilon$ accumulates, indicating the model becomes increasingly vulnerable to privacy risk though more accurate. To compute the total privacy loss, we leverage the recent privacy accounting methods: Gaussian differential privacy (GDP) \cite{dong2019gaussian,bu2020deep} and Moments accountant \cite{abadi2016deep,canonne2020discrete}. Both methods give valid though different upper bounds of $\epsilon$ as a consequence of using different composition theories. Notably, the rate at which the privacy compromises depends on the certain hyperparameters, such as the number of iterations $T$, the learning rate $\eta$, the noise scale $\sigma$, the batch size $|B|$, the clipping norm $C$. In the following sections, we exploit how these training hyperparameters influences DP and the convergence, and subsequently the uncertainty quantification.

\begin{algorithm}[H]
 \caption{Differentially private SGD (DP-SGD) with regularization}
 \label{alg:DPSGD}
 \begin{algorithmic}
   \State {\bfseries Input:} Examples $\{x_1,x_2,\dots,x_n;y_1,\dots,y_n\}$, loss function $\ell(\cdot;\w)$, initial weights $\w_0$.

   \For{$t=1$ to $T$}
   \State Sample $S_t\subset \{1,2,\dots,N\}$ uniformly at random.
\For{$i\in S_t$}
   \State Compute $g_i =\nabla_{\w} \ell(x_i,y_i;\w_{t-1})$ \Comment $g_i$ is the per-sample gradient
  \State Define $\widetilde{g}_i=\min\{1,\frac{C_t}{\|g_i\|_2}\}\cdot g_i.$
  \Comment{Clip the per-sample gradients}
   \EndFor
   \State Define $\hat{g} = \frac{1}{B}\sum_{i=1}^B\widetilde{g}_i + \frac{\sigma \cdot C_t}{B}\cdot \mathcal{N}(0,I_d).$
    \Comment{Add noise}
   \State Update $\w_t \leftarrow \w_{t-1}-\eta_t\left(\hat{g}+\nabla_{\w}r(\w_{t-1})\right)$
   \Comment{Descend}
	\EndFor
\State {\bfseries Output:} $\w_1,\w_2,\cdots,\w_T$
\end{algorithmic}
\end{algorithm}

\section{Bayesian Neural Networks} \label{sec3: BNN framework}

BNNs have achieved significant success recently, by incorporating expert knowledge and making statistical inference through uncertainty quantification. On the high level, BNNs share the same architecture as regular NNs $f(x;\w)$ but are different in that BNNs treat weights as a probability distribution instead of a single deterministic value. Learned properly, these weight distributions can characterize the uncertainty in prediction and improve the generalization behavior. For example, suppose we have obtained the weight distribution $W$, then the prediction distribution of BNNs is $f(x;W)$, which is unavailable by regular NNs. We now describe three popular yet distinct approaches to learn BNNs, leaving the algorithms in \Cref{sec4: dp and bnn}, which has the DP-BNNs but reduces to non-DP BNNs when $\sigma=0$ (no noise) and $C_t=\infty$ (no clipping). We highlight that all three approaches are heavily based on SGD (though other optimizers can also be used): the difference lies in how SGD is applied. The implementation is available in Pytorch at a public repository \url{github.com/JavierAntoran/Bayesian-Neural-Networks}.

\begin{figure*}[!htb]
\centering
\begin{minipage}{.22\textwidth}
\includegraphics[width=0.95\linewidth]{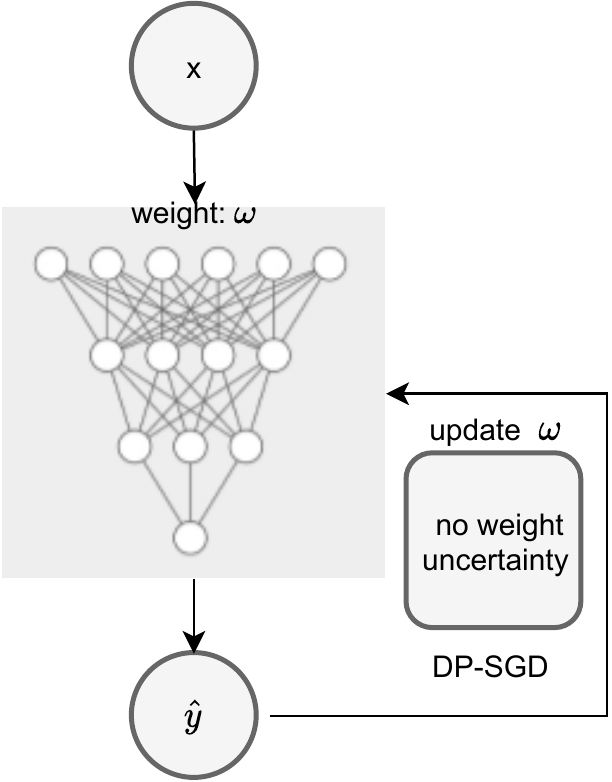}
\end{minipage}%
\hspace{0.5cm}
\begin{minipage}{.21\textwidth}
\includegraphics[width=\linewidth]{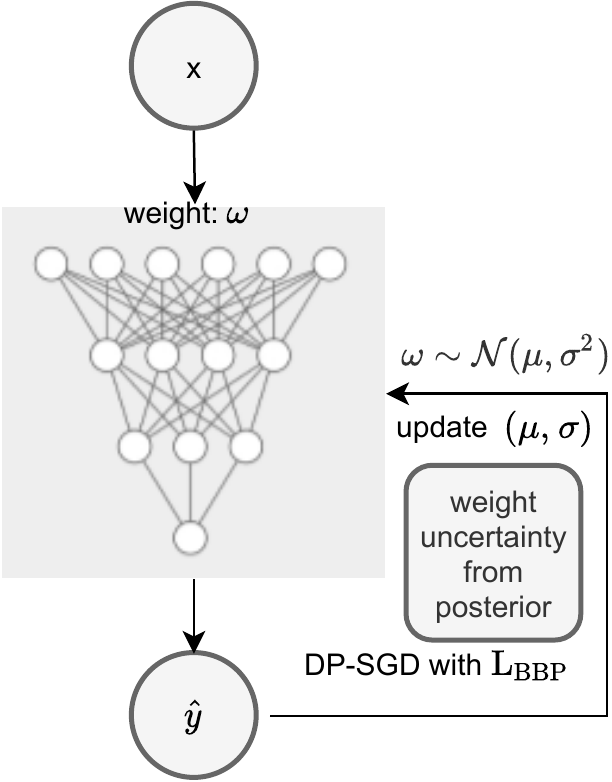}
\end{minipage}%
\hspace{0.5cm}
\begin{minipage}{.21\textwidth}
  \includegraphics[width=\linewidth]{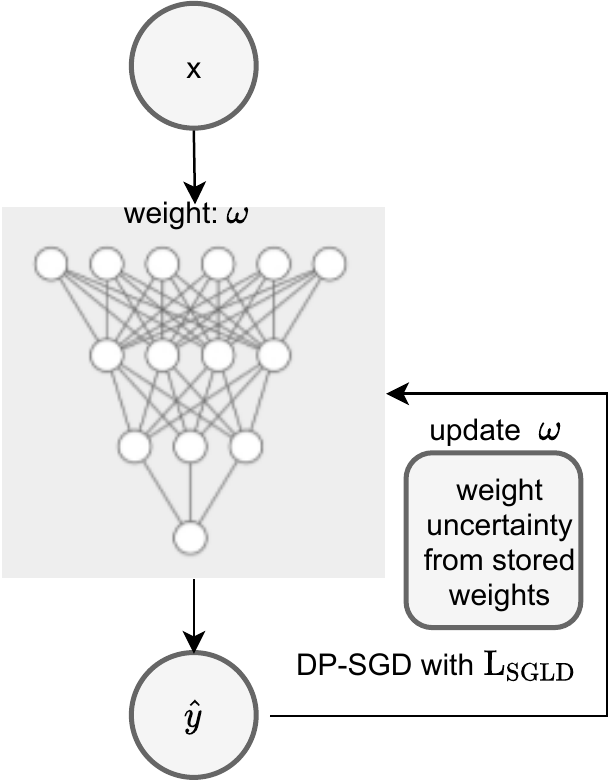}  
\end{minipage}%
\hspace{0.5cm}
\begin{minipage}{.21\textwidth}
\includegraphics[width=\linewidth]{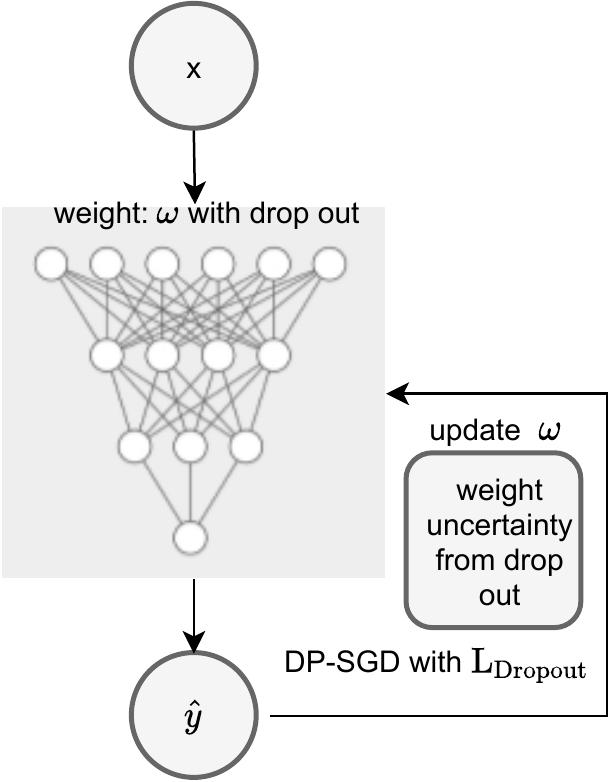}
\end{minipage}%
\vspace{-0.2cm}
\caption{Training procedure of private SGD, BBP, SGLD, and MC Dropout (left to right). Applying non-DP optimizers instead results in the regular training.}
\label{fig: BNN training procedure}
\end{figure*}

\begin{figure*}[!htb]
\centering
\begin{minipage}{.25\textwidth}
\vspace{1.2cm}
\includegraphics[width=0.4\linewidth]{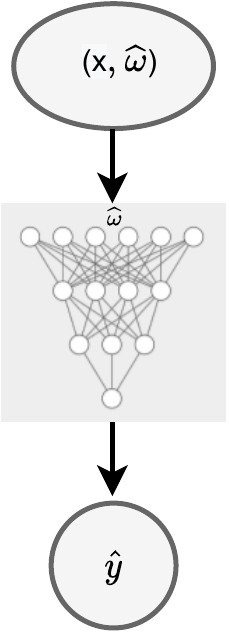}
\end{minipage}%
\hspace{-1cm}
\begin{minipage}{.24\textwidth}
\includegraphics[width=\linewidth]{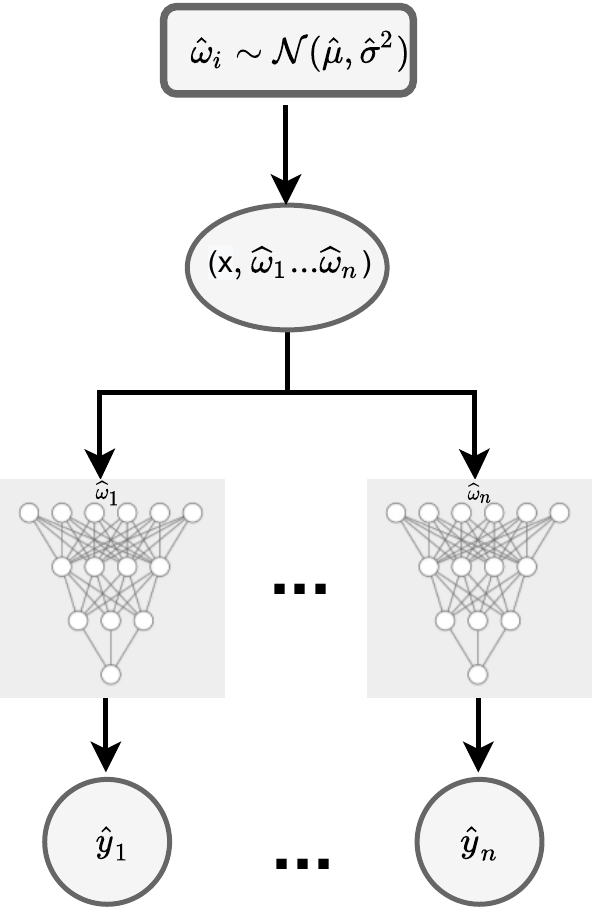}
\end{minipage}%
\hspace{0.5cm}
\begin{minipage}{.24\textwidth}
  \includegraphics[width=\linewidth]{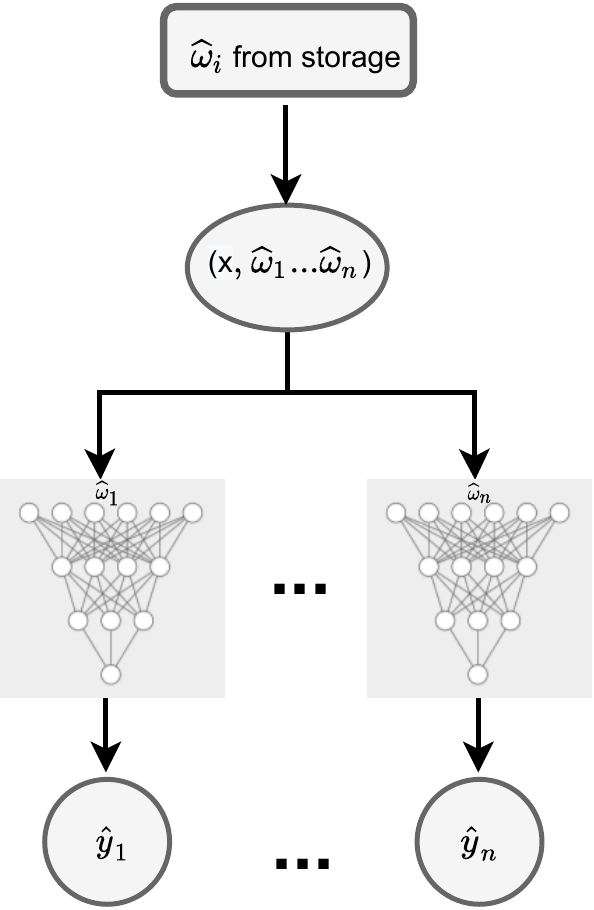}  
\end{minipage}%
\hspace{0.5cm}
\begin{minipage}{.24\textwidth}
\includegraphics[width=\linewidth]{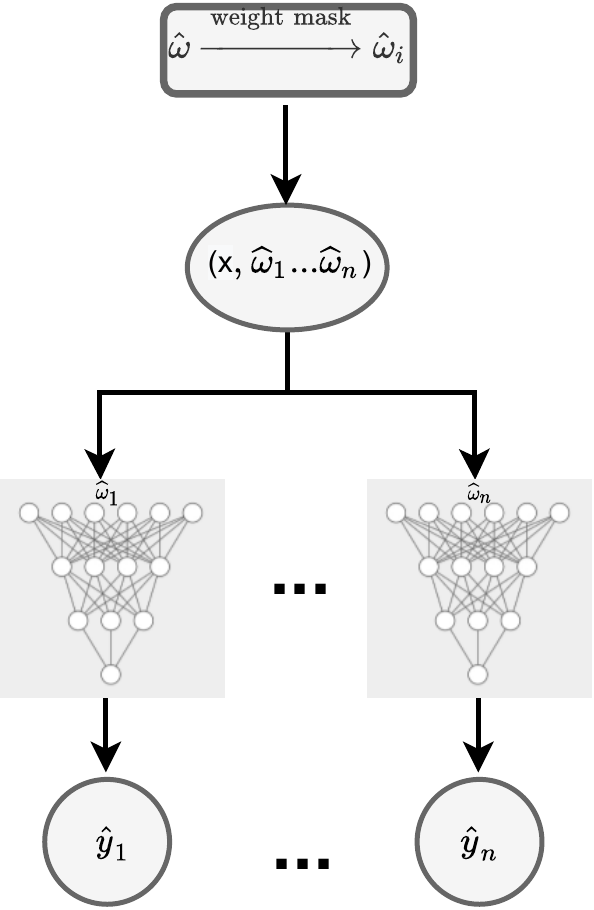}
\end{minipage}%
\vspace{-0.2cm}
\caption{Inference (or prediction) procedure of non-Bayesian NN and BNNs: BBP, SGLD, and MC Dropout (left to right). Note that DP-BNNs have the same procedure as regular BNNs, as DP is enforced during the training procedure.}
\label{fig: BNN inference procedure}
\vspace{-0.4cm}
\end{figure*}
\subsection{Bayesian NN via Sampling:  Stochastic Gradient Langevin Dynamics (SGLD)}


SGLD \cite{welling2011bayesian,li2016preconditioned} is a gradient method that applies on the weights $\w$ of NN, and the weight uncertainty arises from the random noises injected into the dynamics. Unlike SGD, SGLD makes $\w$ to converge to a posterior distribution rather than to a point estimate, from which SGLD can sample and characterize the uncertainty of $\w$. In details, SGLD takes the following form
$$\w_t = \w_{t-1}+\eta_t\left(\nabla\log p(\w_{t-1})+\frac{n}{|B_t|}\sum_{i\in B_t} \nabla \log p(\x_i,y_i|\w_{t-1})\right)+ \mathcal{N}(0,\eta_t)$$
where $p(\w)$ is the pre-defined prior distribution of weights and $p(\x,y|\w)$ is the likelihood of data. In the literature of empirical risk minimization, SGLD can be viewed as a special case of SGD with noise in the updates, even though it does not really correspond to any minimization problem.
$$\w_t = \w_{t-1}-\eta_t\left(\nabla r(\w_{t-1})+\frac{n}{|B_t|}\sum_{i\in B_t} \nabla \ell(\x_i,y_i;\w_{t-1})\right)+ \mathcal{N}(0,\eta_t),$$
where $r$ is the regularization and $\ell$ is loss, depending on the prior and the likelihood. See \Cref{footnote:regularization_prior} for details. Writing the penalized loss as 
$\mathcal{L}_\text{SGLD}(\x_i,y_i;\w):=n\cdot\ell(\x_i,y_i;\w)+r(\w),$
we obtain
$$\w_t = \w_{t-1}-\frac{\eta_t}{|B_t|}\sum_{i\in B_t}\frac{\partial L_\text{SGLD}(\x_i,y_i)}{\partial \w_{t-1}}+ \mathcal{N}(0,\eta_t).$$
\vspace{-0.3cm}

Interestingly, although SGLD adds noise to the gradient, it is not guaranteed as DP\footnote{SGLD is not DP in deep learning as the sensitivity is possibly unbounded. However, if the sensitivity is bounded, SGLD is automatically DP. This is known as `privacy for free'\cite{wang2015privacy}.}. While SGLD is different from SGD, we show in \Cref{thm:dpsgld=dpsgd} that DP-SGLD is a sub-class of DP-SGD.

\subsection{Bayesian NN via Optimization}
\subsubsection{Bayes By Backprop (BBP)}
BBP \cite{blundell2015weight} uses the standard SGD except it is applied on the hyperparameters of pre-defined weight distributions, rather than on weights $\w$ directly. This approach is known as the `variational inference' or the `variational Bayes', where a \textit{variational distribution} $q(\w|\theta)$ is learned through its governing hyperparameters $\theta$. Consequently, the weight uncertainty is included in such variational distribution from which we can sample. 

Given data $D=\{(\x_i,y_i)\}$, the likelihood is $p(D|w)=\Pi_i p(y_i|\x_i,\w)$ under some probabilistic model $p(y|\x,\w)$. By the Bayes theorem, the posterior distribution $p(w|D)$ is proportional to the likelihood and the prior distribution $p(w)$,
\begin{align*}
p(w|D)\propto p(D|w)p(w)= \Pi_i p(y_i|x_i,w)p(w).
\end{align*}
Within a pre-specified variational distribution $q(w|\theta)$, we seek the distributional parameter $\theta$ such that $q(w|\theta)\approx p(w|D)$. Conventionally, the variational distribution is restricted to be Gaussian and we learn its mean and standard deviation $\theta=(\mu,\sigma)$ through minimizing the KL divergence:
\begin{align}\label{problem:min KL}
\min_\theta KL\left(q(w|\theta)\big\|p(w|D)\right)
\equiv \mathbb{E}_{q(w|\theta)}\log q(w|\theta)-\mathbb{E}_{q(w|\theta)}\log p(w)-\mathbb{E}_{q(w|\theta)}\log p(D|w).
\end{align}
This objective function is analytically intractable but can be approximated by drawing $w^{(j)}$ from $q(w|\theta)$ for $N$ times. The optimization objective denoted by $\mathcal{L}_\text{BBP}$ is defined in \Cref{app:detail BBP}. It follows that the SGD updating rule for $\theta=(\mu,\rho)$ with $\rho:=\ln\sigma$ is
$$\mu_t = \mu_{t-1}-\frac{\eta_t}{|B_t|}\sum_{i\in B_t} \frac{d \mathcal{L}_\text{BBP}(\x_i,y_i)}{d\mu},
\quad\quad
\rho_t=\rho_{t-1}-\frac{\eta_t}{|B_t|}\sum_{i\in B_t}\frac{d \mathcal{L}_\text{BBP}(\x_i,y_i)}{d\rho}$$

\subsubsection{Monte Carlo Dropout (MC Dropout)}\label{mc_dropout}

MC Dropout is proposed by \cite{gal2016dropout} that establishes an interesting connection: optimizing the loss with $L_2$ penalty in regular NNs with dropout layers is equivalent to learning Bayesian inference approximately. From this perspective, the weight uncertainty is described by the randomness of the dropout operation. We refer to \Cref{app:more MC} for an in-depth review of MC dropout.

In more detail, given
$\mathcal{L}_\text{Dropout}(\x_i,y_i;\w):=\ell(\x_i,y_i;\w) + r(\w)$, such connection equalizes the problem $\min_{\w} \frac{1}{n} \sum_{i=1}^{n} \mathcal{L}_\text{Dropout}(\x_i,y_i;\w)$  
with the variational inference problem \eqref{problem:min KL} in \Cref{app:detail BBP}, when the prior distribution is a zero mean Gaussian one. 
This equivalence makes MC Dropout similar to BBP in the sense of minimizing the same KL divergence. Nevertheless, while BBP directly minimizes the KL divergence, MC Dropout in practice operates under the empirical risk minimization. Hence MC Dropout also shares similarity with SGD or SGLD. From the algorithmic perspective, suppose $\w_t$ is the remaining weights after the $t$-th dropout, then the updating rule with SGD is
\begin{align*}
\w_t = \w_{t-1}-\frac{\eta_t}{|B_t|}\sum_{i\in B_t}\frac{\partial L_\text{Dropout}(\x_i,y_i;\w_{t-1})}{\partial \w_{t-1}}.
\end{align*}

\section{Differentially Private Bayesian Neural Networks} \label{sec4: dp and bnn}
To prepare the development of DP-BNNs, we summarize how to transform a regular NN to be Bayesian and to be DP, respectively. To learn a BNN, we need to establish the relationship between the Bayesian quantities (likelihood and prior) and the optimization loss and regularization. Under the Bayesian regime, $\ell$ is the negative log-likelihood $-\log p(x,y|\theta)$ and $\log p(\theta)$ is the log-prior. Under the empirical risk minimization regime, $\ell$ is the loss function and we view $-\log p(\theta)$ as the regularization or penalty\footnote{For example, if the prior is $\mathcal{N}(0,\sigma^2)$, then $-\log p(\theta)\propto \frac{\|\theta\|^2}{2\sigma^2}$ is the $L_2$ penalty; if the prior is Laplacian, then $-\log p(\theta)$ is the $L_1$ penalty; additionally, the likelihood of a Gaussian model corresponds to the MSE loss.\label{footnote:regularization_prior}}. To learn a DP network, we simply apply DP gradient methods that guarantee DP via the Gaussian mechanism (see \Cref{app:DP background}). Therefore, we can privatize each BNN to gain DP guarantee by applying DP gradient methods to update the parameters, as shown in \Cref{fig: BNN training procedure} and \Cref{fig: BNN inference procedure}.

In what follows, we introduce DP-BNNs with a preview of their algorithmic properties in \Cref{table:DPBNN comparison}.

\begin{table}[!h]
\vspace{-0.2cm}
\centering
\begin{tabular}{ |c| c |c|c| }
    \hline
     & DP-SGLD& DP-BBP  & DP-MC Dropout\\ \hline
    General weight prior & Yes & Yes& No\\ \hline
    General network architecture& Yes& No  &Yes\\ \hline
    General optimizers & --- & Yes& Yes\\ \hline    Computational complexity &Low&High&Low\\\hline
    Storage memory cost &High&Low&Low\\\hline
    Accelerable by outer product &Yes&No&Yes\\\hline
    Analytic posterior distribution &No&Yes&No\\\hline
  \end{tabular}
  \caption{Algorithmic comparison of DP-BNNs .}
  \label{table:DPBNN comparison}
\end{table}

\subsection{Differentially Private Stochastic Gradient Langevin Dynamics}
\begin{algorithm}[H]
 \caption{Differentially private SGLD (DP-SGLD)}
 \label{alg:DPSGLD}
 \begin{algorithmic}
   \State {\bfseries Input:} Examples $\{x_1,x_2,\dots,x_n;y_1,\dots,y_n\}$, loss function $\ell(\cdot;\w)$, initial weights $\w_0$.
   
   \For{$t=1$ to $T$}
   \State Sample a batch $B\subset \{1,2,\dots,n\}$ uniformly at random.
   \For{$i\in B$}
   \State Compute $g_i =\nabla_{\w}  \ell(x_i,y_i;\w_{t-1})$ \Comment $g_i$ is the per-sample gradient
  \State Define $\widetilde{g}_i=\min\{1,\frac{C_t}{\|g_i\|_2}\}\cdot g_i.$
    \Comment{Clip the per-sample gradients}
   \EndFor
   \State Update $\w_t \leftarrow \w_{t-1}-\eta_t\left(\frac{n}{|B|}\sum_{i\in B} \widetilde{g}_i+\nabla_{\w} r(\w_{t-1})\right)+ \mathcal{N}(0,\eta_t)$
   \Comment{Add noise and descend}
	\EndFor
\State {\bfseries Output:} $\w_1,\w_2,\dots,\w_T$
\end{algorithmic}
\end{algorithm}

DP-SGLD was proposed by \cite{wang2015privacy} for non-deep learning with DP and then equipped with the per-sample clipping to work in deep learning \cite{li2019connecting}, though the uncertainty quantification has not been investigated nor compared to other DP-BNNs until this work. Furthermore, our analysis is different from both existing works\footnote{
Remarkably, our DP-SGLD is different from \cite{wang2015privacy} on the algorithmic level: (1) we need per-sample clipping as the gradient norm can be unbounded in deep learning; (2) we need not to adjust the noise scale to guarantee DP. Furthermore, our DP-SGLD privacy analysis is different than \cite{li2019connecting} as we work with any learning rate while \cite[Theorem 3]{li2019connecting} has multiple constraints on the learning rate, which renders their DP guarantee invalid if violated.}.

One can view SGLD as noisy SGD plus some regularization, and consequently view DP-SGLD as DP-SGD with regularization: e.g. DP-SGLD with non-informative prior is a special case of vanilla DP-SGD; DP-SGLD with Gaussian prior is equivalent to some DP-SGD with weight decay (i.e. with $L_2$ penalty). This equivalence is made clear in the following theorem.

\begin{wrapfigure}{l}{0.4\textwidth}
      \vspace{-0.1cm}
    \centering
  \includegraphics[width=5.5cm]{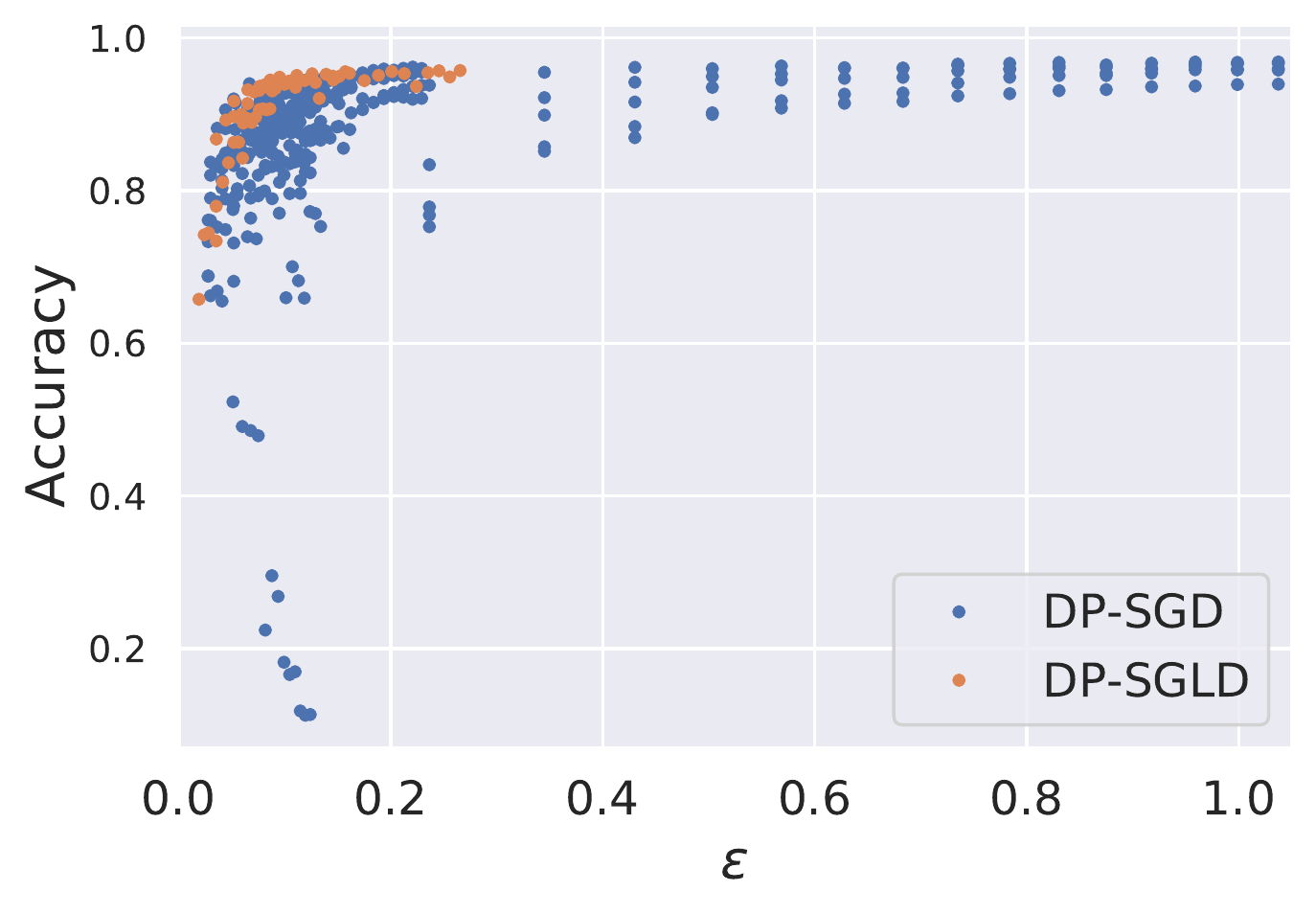}
      \vspace{-0.4cm}
    \caption{Illustration that DP-SGLD performs superiorly within the DP-SGD family, on MNIST with CNN. Here $\delta=10^{-5}, |B|=256, \eta_\textnormal{SGD}=0.25, \eta_\textnormal{SGLD}=10^{-5}$, $C \in \{0.5,1,1.5,2,5 \}, \sigma_{\textup{SGD}} \in \{0.5,0.9,1.3,2,3 \}$, epoch $\in[15]$.}
    \label{fig:SGD vs SGLD}
    \vspace{-0.6cm}
\end{wrapfigure}

\begin{theorem}\label{thm:dpsgld=dpsgd}
For DP-SGLD with some prior assumption and DP-SGD with the corresponding regularization, 
\begin{align*}
&\textnormal{DP-SGLD}\left(\eta_\textnormal{SGLD}=\eta,C_\textnormal{SGLD}=C\right)=\textnormal{DP-SGD}\Big(\eta_\textnormal{SGD} = \eta n, \sigma_\textnormal{SGD} = \frac{\sqrt{\eta}|B|}{nC}, C_\textnormal{SGD} = C \Big),
\\
&\textnormal{DP-SGD}\left(\eta_\textnormal{SGD} = \eta, \sigma_\textnormal{SGD} = \sigma, C_\textnormal{SGD} = C \right)=\textnormal{DP-SGLD} \Big(\eta_\textnormal{SGLD}= \frac{\eta}{n}, C_\textnormal{SGLD} = C = \frac{|B|}{\sqrt{n\eta}\sigma}\Big).
\end{align*}
\end{theorem}
\Cref{thm:dpsgld=dpsgd}, proven in \Cref{appendix: proof}, suggests that DP-SGLD is a sub-class of DP-SGD: every DP-SGLD is equivalent to some DP-SGD; however, only DP-SGD with $\sigma = \frac{|B|}{\sqrt{n\eta}C}$ is equivalent to DP-SGLD. In \Cref{fig:SGD vs SGLD}, we empirically observe that DP-SGLD is indeed a sub-class in the family of DP-SGD and is superior to other members of this family as it occupies the top left corner of the graph. In fact, it has been suggested by \cite{wang2015privacy} in the non-deep learning that, training a Bayesian model using SGLD automatically guarantees DP. In contrast, \Cref{thm:dpsgld=dpsgd} is established in the deep learning regime and brings in a new perspective: training a regular NN using DP-SGD may automatically allow Bayesian uncertainty quantification.

Furthermore, DP-SGLD is generalizable to any network architecture (whenever DP-SGD works) and to any weight prior distribution (via different regularization terms). Particularly, DP-SGLD enjoys fast computation speed since it can be significantly accelerated using the outer product method \cite{goodfellow2015efficient,rochette2019efficient}, the fastest acceleration implemented in \texttt{Opacus} library. For example, on MNIST in \Cref{sec5: experiments}, DP-SGLD requires only 10 sec/epoch, while DP-BBP takes 480 sec/epoch since it is incompatible with outer product.

However, DP-SGLD only offers empirical weight distribution $\{\w_t\}$ which requires large memory for storage in order to give sufficiently accurate uncertainty quantification (e.g. we record 100 iterations of $\w_t$ in \Cref{label_3} and 1000 iterations in \Cref{regression}). The memory burden can be too large to scale to large models that have billions of parameters.

\subsection{Differentially Private Bayes by BackPropagation}

\begin{algorithm}[!h]
 \caption{Differentially private Bayes by BackPropagation (DP-BBP)}
 \label{alg:DPBBP}
 
 \begin{algorithmic}
   \State {\bfseries Input:} Examples $\{x_1,x_2,\dots,x_n;y_1,\dots,y_n\}$, loss function $\ell(\cdot;\theta)$, initial parameters $\theta_0$.
   
   \For{$t=1$ to $T$}
   \State Sample a batch $B\subset \{1,2,\dots,n\}$ uniformly at random.
   \For{$i\in B$}
   \For{$j=1$ to $N$}
   \State Sample  $w^{(j)}$ from $q(w|\theta_{t-1})$.
   \State Compute $g_i^{(j)} =\nabla_\theta \mathcal{L}_\text{BBP}(x_i,y_i;w^{(j)})$ 
   \Comment \small{$g_i^{(j)}$ is the per-sample gradient in the $j$-th sampling}\normalsize
   \EndFor
  \State Define $\bar{g_i}=\frac{1}{N}\sum_j g_i^{(j)}$ \Comment $\bar{g_i}$ is the averaged per-sample gradient
  \State Define $\widetilde{g}_i=\min\{1,\frac{C_t}{\|\bar{g_i}\|_2}\}\cdot \bar{g_i}.$
  \Comment{Clip the per-sample gradients}
   \EndFor
   \State Define $\hat{g} = \frac{1}{B}\sum_{i\in B}\widetilde{g}_i + \frac{\sigma \cdot C_t}{|B|}\cdot \mathcal{N}(0,I_d).$
    \Comment{Add noise}
  \State Update $\theta_t \leftarrow \theta_{t-1}-\eta_t\widehat{g}$
      \Comment{Descend}
 \EndFor
\State {\bfseries Output:} $\theta_T$
\end{algorithmic}
\end{algorithm}

Our DP-BBP can be viewed as DP-SGD (or any other DP optimizers, e.g. DP-Adam) working on the distributional hyperparameters (in fact, it is the only method that does not works on weights directly). Similar to DP-SGLD, the DP-BBP can flexibly work under various priors by using different regularization terms. In sharp contrast to DP-SGLD and DP-MC Dropout, which only describe the weight distribution empirically, DP-BBP directly characterizes an analytic weight distribution.

However, DP-BBP suffers from high computation complexity and incapability of acceleration. Under Gaussian variational distributions, DP-BBP needs to compute two hyperparameters (mean and standard deviation) for a single parameter (weight), which doubles the complexity of DP-SGLD, DP-MC Dropout and DP-SGD. The computational issue is further exacerbated due to the $N$ samplings of $\w^{(j)}$ from $q(\w|\theta_t)$, which means the number of back-propagation is $N$ times that of DP-SGLD and DP-MC Dropout. This introduces an inevitable tradeoff: when $N$ is larger, DP-BBP tends to be more accurate but its computational complexity is also higher, leading to the overall inefficiency of DP-BBP. Moreover, DP-BBP cannot be accelerated by the outer product method as it violates the supported network layers\footnote{Since DP-BBP does not optimize the weights, the back-propagation is much different from using $\frac{\partial \ell}{\partial \w}$ (see \Cref{app:detail BBP MC}) and thus requires new design that is currently not available. See \url{https://github.com/pytorch/opacus/blob/master/opacus/supported_layers_grad_samplers.py}.}. Since the per-sample gradient clipping is the computational bottleneck for acceleration, DP-BBP can be too slow to be practically useful if the computation consideration overweighs its utility (see \Cref{acc_table}).

\subsection{Differentially Private Monte Carlo Dropout}

We can view our DP-MC Dropout as applying DP-SGD (or any other DP optimizers) on any NN with dropout layers, and thus DP-MC Dropout enjoys the acceleration provided by the outer product method in \texttt{Opacus}. Regarding the uncertainty quantification, DP-MC Dropout offers the empirical weight distribution at low computation costs and low storage costs since only $\w_T$ is stored. A limitation to the theory of MC Dropout \cite{gal2016dropout} is that the equivalence between the empirical risk minimization of $\mathcal{L}_\text{Dropout}$ and the KL divergence minimization \eqref{problem:min KL} no longer holds beyond the Gaussian weight prior. Nevertheless, algorithmically speaking, DP-MC Dropout also works with other priors by using different regularization terms. 
\begin{algorithm}[!htb]
 \caption{Differentially private MC Dropout (DP-MC Dropout)}
 \label{alg:DPMCDROPOUT}
 \begin{algorithmic}
   \State {\bfseries Input:} Examples $\{x_1,x_2,\dots,x_n;y_1,\dots,y_n\}$, loss function $\mathcal{L}_{dropout}(\cdot;\w)$, initial weights $\w_0$.
   
   \For{$t=1$ to $T$}
   \State Sample a batch $B\subset \{1,2,\dots,n\}$ uniformly at random.
   \State Randomly drop out some weights and denote the remaining ones as $\w_{t-1}$.
\For{$i\in B$}
   \State Compute $g_i =\nabla_{\w} \ell(\x_i,y_i;\w_{t-1})$ \Comment $g_i$ is gradient
   \State Define $\widetilde{g}_i=\min\{1,\frac{C_t}{\|g_i\|_2}\}\cdot g_i.$
  \Comment{Clip the per-sample gradients}
   \EndFor
   \State Define $\hat{g} = \frac{1}{B}\sum_{i\in B} \widetilde{g}_i + \frac{\sigma \cdot C_t}{|B|}\cdot \mathcal{N}(0,I_d)$
      \Comment{Add noise}
   \State Update $\w_t \leftarrow \w_{t-1}-\eta_t(\hat{g}+\nabla_{\w} r(\w_{t-1})) $
   \Comment{Descend}
	\EndFor
\State {\bfseries Output:} $\w_T$
\end{algorithmic}
\end{algorithm}

\subsection{Analysis of Privacy}
The following theorem gives the analytic privacy loss $\epsilon$, computed by the GDP accountant \cite{dong2019gaussian,bu2020deep}.
\begin{theorem}[Theorem 5 in \cite{bu2020deep}]\label{thm:dpmc&dpbbp privacy}
For both DP-MC Dropout and DP-BBP, under any DP-optimizers (e.g. DP-SGD, DP-Adam, DP-HeavyBall) with the number of iterations $T$, noise scale $\sigma$ and batch size $|B|$, the resulting neural network is $\sqrt{T(e^{1/\sigma^2}-1)}|B|/n$-GDP.
\end{theorem}
We remark that, from \cite[Corollary 2.13]{dong2019gaussian}, $\mu$-GDP can be mapped to $(\epsilon,\delta)$-DP via
$$
  \delta(\varepsilon ; \mu)=\Phi\left(-\varepsilon/\mu+\mu/2\right)-\mathrm{e}^{\varepsilon} \Phi\left(-\varepsilon/\mu-\mu/2\right).
$$
As alternatives to GDP, other privacy accountants such as the Moments Accountant (MA) \cite{abadi2016deep,mironov2019r,canonne2020discrete,asoodeh2020better}, Fourier accountant \cite{koskela2020computing, zhu2022optimal}, and Privacy Random Variable Accountant \cite{gopi2021numerical} can be applied to characterize $\epsilon$, though implicitly as they take a numerical integration approach (see \Cref{app:DP background}). Since DP-MC Dropout and DP-BBP do not quantify the uncertainty via optimizers, all privacy accountants give the same $\epsilon$ as training DP-SGD on regular NNs. We next give the privacy of DP-SGLD by writing it as a special case of DP-SGD.
\begin{theorem}\label{thm:dpsgld privacy}
For DP-SGLD with the number of iterations $T$, learning rate $\eta$, batch size $|B|$ and clipping norm $C$, the resulting neural network is $\sqrt{T(e^{n^2\eta C^2/|B|^2}-1)}|B|/n$-GDP.
\end{theorem}

\begin{figure}[!htb]
\minipage{0.33\textwidth}
  \includegraphics[width=1.1\linewidth]{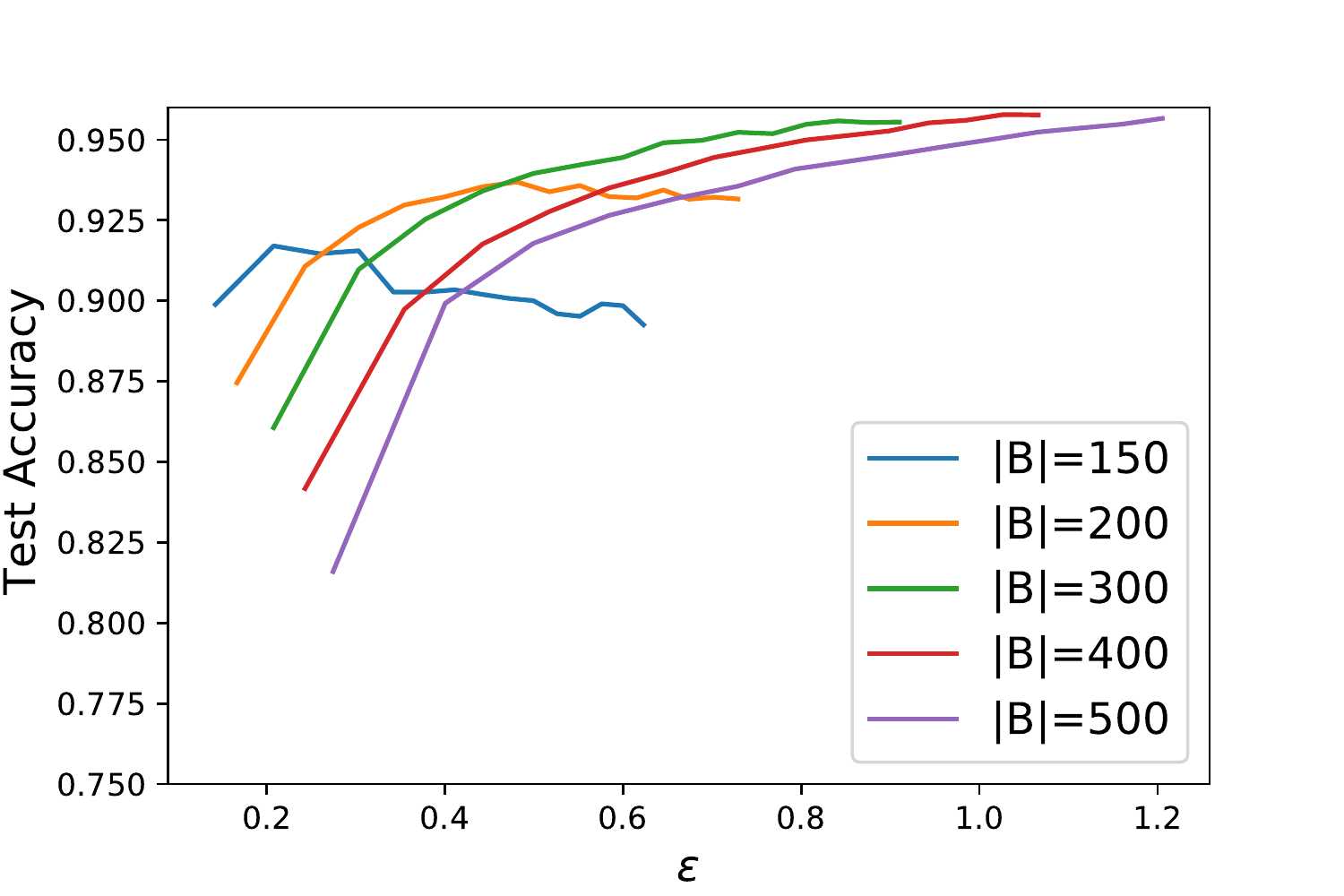}
\endminipage
\minipage{0.33\textwidth}%
  \includegraphics[width=1.1\linewidth]{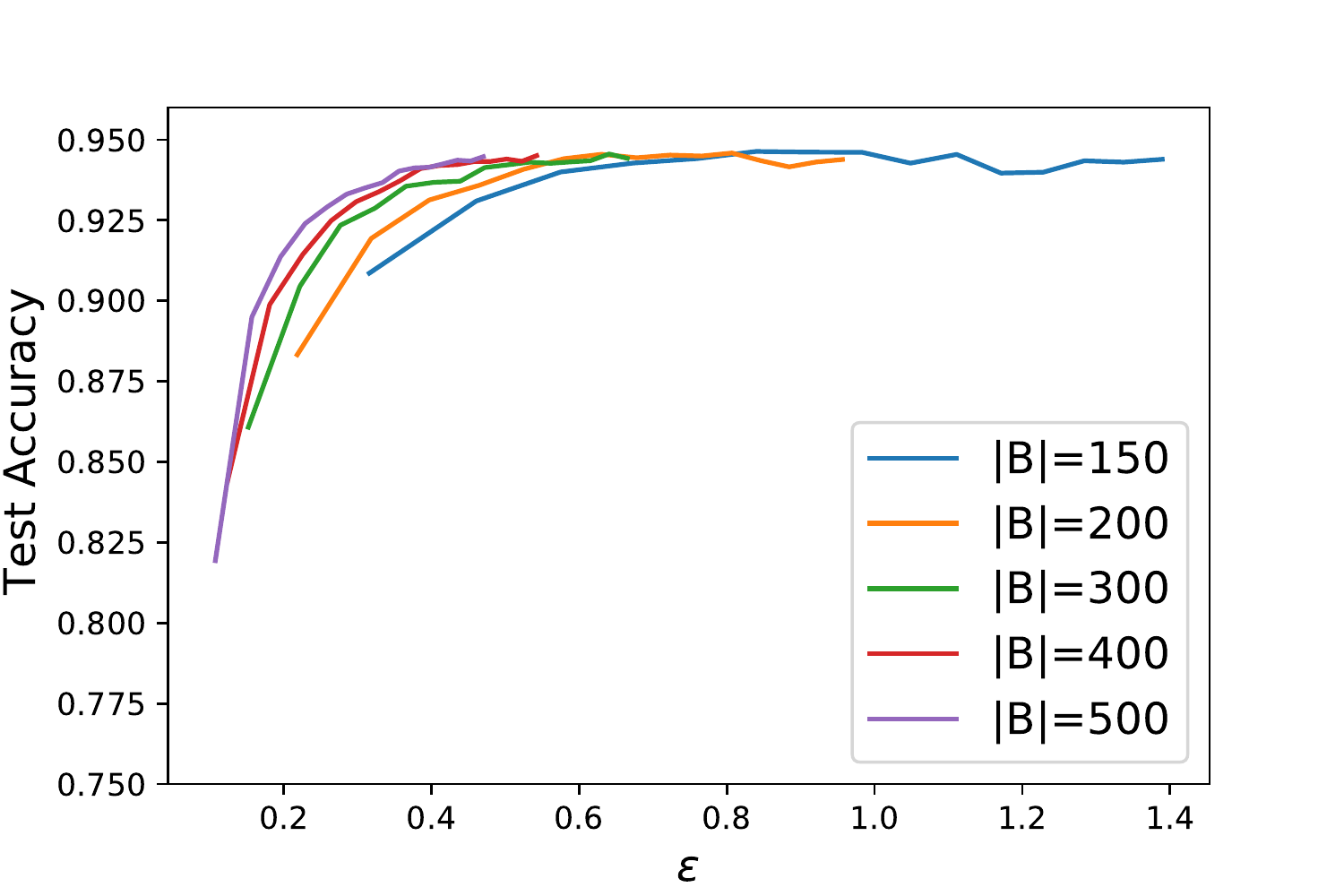}
\endminipage
\minipage{0.33\textwidth}
  \includegraphics[width=1.1\linewidth]{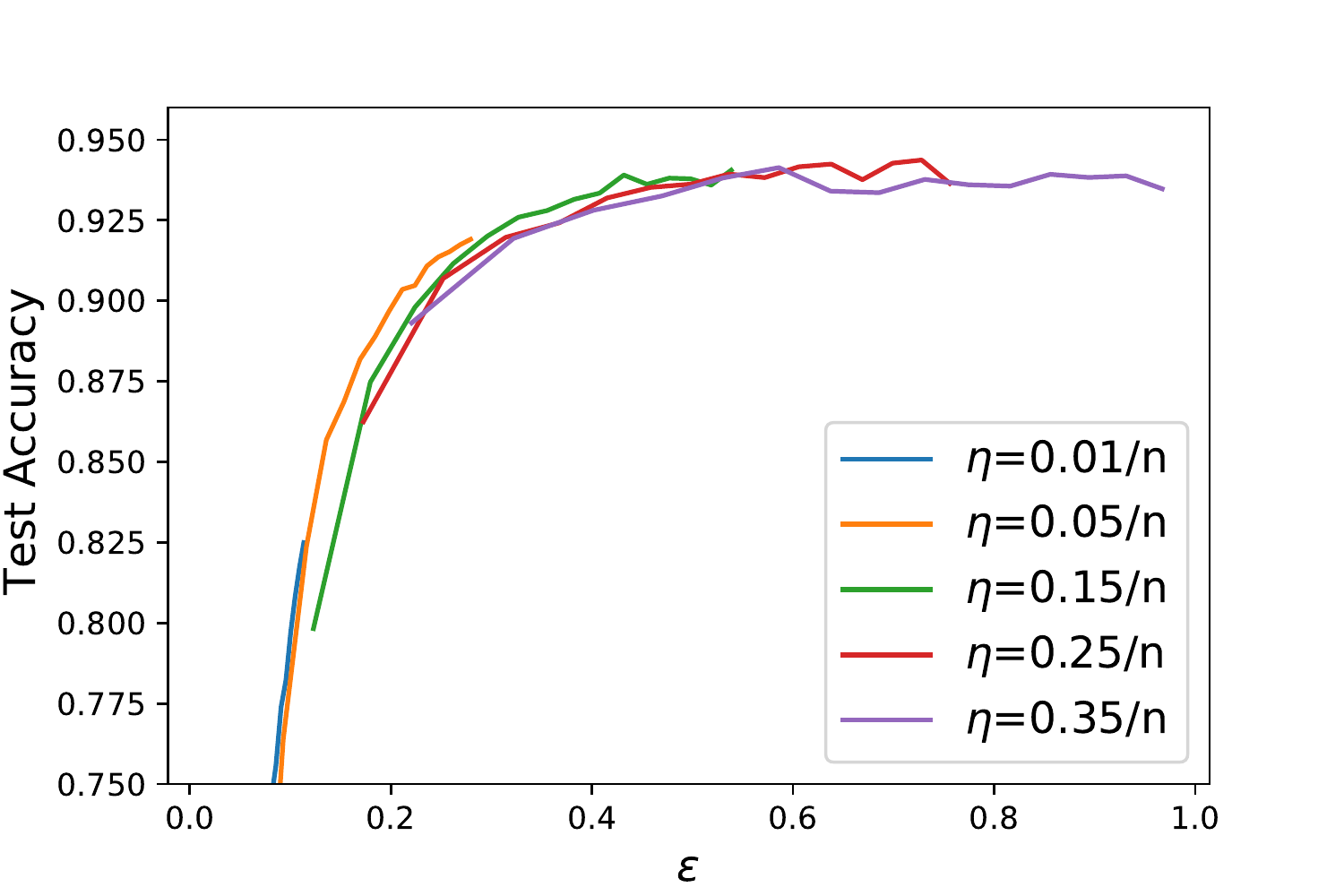}
\endminipage
\caption{Effects of batch size and learning rate on DP-SGD (left) and DP-SGLD (middle \& right) with CNNs under accuracy and privacy measures. Experiment setting can be found in \Cref{app:experiment on training hyperparameter}.}
\label{fig:dpsgld factors}
\end{figure}

The proof follows from \Cref{thm:dpsgld=dpsgd} and \cite[Theorem 5]{bu2020deep}, given in \Cref{appendix: proof}. We observe sharp contrast between \Cref{thm:dpmc&dpbbp privacy} and \Cref{thm:dpsgld privacy}: (1) while the clipping norm $C$ and learning rate $\eta$ have no effect on the privacy guarantee of DP-MC Dropout and DP-BBP, these hyperparameters play important roles in DP-SGLD. For instance, the learning rate triggers a tradeoff: larger $\eta$ converges faster but smaller $\eta$ is more private; see \Cref{fig:dpsgld factors}. (2) To get stronger privacy guarantee, DP-MC Dropout and DP-BBP need smaller $T$ and larger $\sigma$; however, DP-SGLD needs smaller $T, C$ and $\eta$. (3) Surprisingly, the batch size $|B|$ has opposite effects in DP-SGLD and in other methods: DP-SGLD with larger $|B|$ is more private, 
in sharp contrast with DP-SGD for which smaller subsampling probability (i.e. smaller $|B|$) can amplify the privacy \cite{balle2018privacy,wang2019subsampled,kasiviswanathan2011can,dong2019gaussian,beimel2010bounds}.
This observation is further visualized in \Cref{fig:dpsgld factors}, lending support to the striking difference between the optimization approach (DP-MC Dropout and DP-BBP) and the sampling approach (DP-SGLD). 

\section{Convergence analysis of DP-SGLD}
In this section, we give the convergence analysis of DP-SGLD with the per-sample gradient clipping. Our analysis works with non-convex loss or likelihood $\mathcal{L}$ and shows that DP-SGLD can converge to stationary points at a rate $O(T^{-1/4})$.

We make the following assumptions which are standard in the non-private SGD literature \cite{allen2018natasha,bottou2018optimization,ghadimi2013stochastic} and in the recent study of DP-SGD \cite{bu2022automatic}.
\begin{assumption}[Lipschitz Smoothness]\label{aspt: smooth}
Let $g(\bm{w})$ denote the gradient of the objective $\mathcal{L}(\bm{w})$. Then $\forall \bm{w}$, $\bm{v}$, there is an non-negative constant $L$ such that:
\begin{equation}\label{smooth}
    \mathcal{L}(\bm{v})- [ \mathcal{L}(\bm{w})+ g(\bm{w})^T(\bm{v}-\bm{w}) ] \leq \frac{L}{2}\norm{\bm{w}-\bm{v}}^2
\end{equation}
\end{assumption}

\begin{assumption}[Gradient noise]\label{aspt: gn}
At each iteration, the per-sample ($i$-th sample) gradient noise $ \tilde{g}_{i} - g_i$ is i.i.d. from some distribution such that: 
\begin{equation}\label{gn}
    \mathbb{E}(\tilde{g}_{i} - g_i)=0, \mathbb{E}\norm{\tilde{g}_{i} - g_i}^2 \leq \xi^2,
\end{equation}
where $\tilde{g}_{i} $ is centrally symmetric about $g_i$ in distribution, which is defined by:
$$
\tilde{g}_{i}  - g_i  \overset{\mathcal{D}}{=} g_i-\tilde{g}_{i}.
$$
\end{assumption}

\begin{assumption}[Clipping always happens]\label{aspt: clipping happens}
At each iteration, we have $\|g_i\|_2>C$.
That is, the clipping indeed takes place on all per-sample gradients\footnote{This assumption has been empirically verified in \citep{bu2022automatic} that on several language tasks and GPT2 models, state-of-the-art accuracy is achieved with small clipping norms such that all per-sample gradients $g_i$ are clipped at all iterations.}.
\end{assumption}

With these assumptions in place, we formally analyze the convergence of DP-SGLD with the proof in \Cref{appendix: proof}.

\begin{theorem}\label{thm:dpsgld upperbound}
Let $\mathcal{L}_0$ be the initial loss and $d$ be the number of parameters in neural network. Under \Cref{aspt: smooth}, \ref{aspt: gn}, \ref{aspt: clipping happens}, running DP-SGLD for $T$ iterations with learning rate $\eta=O(T^{-1/3})$ and $C=C_0 T^{-1/6}$ gives:
\begin{equation}
    \min_{0\leq t \leq T} \mathbb{E}(\norm{g_t}) =O(T^{-1/4}).
    \label{eq:dpsgld grad norm}
\end{equation}
\end{theorem}

Interestingly, DP-SGLD converges to a stationary point at the same asymptotic rate as the DP-SGD, supporting our empirical observation in \Cref{fig:SGD vs SGLD} that DP-SGLD is a superior member within the DP-SGD family. In other words, DP-SGLD not only inherits the fast convergence rate of DP-SGD but additionally allows the uncertainty quantification as we will show in \Cref{sec5: experiments}.

\begin{remark}
We note from \eqref{eq:dpsgld grad norm} that DP-SGLD has the same asymptotic convergence rate $O(T^{-1/4})$ as the DP-SGD and the non-private SGD, which is shown in \citep[Theorem 4]{bu2022automatic}. However, the hyperparamters to achieve such convergence rate are different for DP-SGLD ($\eta_\text{SGLD}=O(T^{-1/3}), C_\text{SGLD}=O(T^{-1/6})$) and DP-SGD ($\eta_\text{SGD}=O(T^{-1/2}), C_\text{SGD}=O(1)$). \end{remark}
\begin{remark}
In fact, from an optimization only viewpoint, we can also use $\eta_\text{SGLD}=O(T^{-1/2}), C_\text{SGLD}=O(1)$, the same as DP-SGD, to achieve $\min_{0\leq t \leq T} \mathbb{E}(\norm{g_t}) \leq O(T^{-1/4})$. Nevertheless, if we further take the privacy into consideration through \Cref{thm:dpsgld privacy}, we observe an $O(T^{0.75})$ increase in GDP, which means faster growth of privacy risk along the training. This rate can only be reduced to $O(\sqrt{T})$, the same as DP-SGD, using
$\eta_\text{SGLD}=O(T^{-1/3}), C_\text{SGLD}=O(T^{-1/6})$.
\end{remark}

\section{Experiments} \label{sec5: experiments}
We further evaluate the proposed DP-BNNs on the classification (MNIST) and regression tasks, based on performance measures including uncertainty quantification, computational speed and privacy-accuracy tradeoff. In particular, we observe that DP-SGLD tends to outperform DP-MC Dropout, DP-BBP and DP-SGD, with little reduction in performance compared to non-DP models. All experiments (except BBP) are run with \texttt{Opacus} library under Apache License 2.0 and on Google Colab with a P100 GPU. A detailed description of the experiments can be found in \Cref{appendix:experiment}.

\subsection{Classification on MNIST} \label{sec:classification}  
We first evaluate three DP-BNNs on the MNIST dataset, which contains $n=60000$ training samples and 10000 test samples of $28\times28$ grayscale images of hand-written digits. 

\textbf{Accuracy and Privacy}
While all of non-DP methods have similar high test accuracy, in the DP regime in \Cref{acc_table}, DP-SGLD outperforms other Bayesian and non-Bayesian methods under almost identical privacy budgets (DP-SGLD has $\epsilon_\text{GDP}=0.861$ or $\epsilon_\text{MA}=0.989$; other DP models have $\epsilon_\text{GDP}=0.834$ or $\epsilon_\text{MA}=0.955$; for details of both accountants, see \Cref{app:DP background}). For the multilayer perceptron (MLP), all BNNs (DP or non-DP) do not lose much accuracy when gaining the ability to quantify uncertainty, compared to the non-Bayesian SGD. However, DP comes at high cost of accuracy, except for DP-SGLD which does not deteriorate comparing to its non-DP version, while other methods experience an accuracy drop $\approx 20\%$. Furthermore, DP-SGLD enjoys clear advantage in accuracy when the more complicated convolutional neural network (CNN) is used. 

\begin{table}[!htb]  
\centering 
\begin{tabular}{|c|c|c|c|c|} 
\hline  
 Methods & Weight Prior & DP Time/Epoch &DP accuracy & Non-DP accuracy   
\\ [0.2ex]  
\hline   
& Gaussian& 10s & 0.90 (0.95) & 0.95 (0.96)\\[-1ex]
\raisebox{1ex}{SGLD} & Laplacian&  10s 
&0.89 (0.89) & 0.90 (0.89)  \\  \hline  
 & Gaussian & 480s& 0.80 (-----)  & 0.97 (-----)  \\[-1ex]  
\raisebox{1.5ex}{BBP} & Laplacian &  480s &0.81 (-----)  & 0.98 (-----) \\
\hline  
MC Dropout &  Gaussian  & 9s
&0.78 (0.77) & 0.98 (0.97)   \\
\hline  
SGD (non-Bayesian) &  ------  & 10s
& 0.77 (0.95) & 0.97 (0.99)   \\
\hline 
\end{tabular}  
\caption{Test accuracy and running time of DP-BBP, DP-SGLD, DP-MC Dropout, DP-SGD, and their non-DP counterparts. We use a default two-layer MLP and additionally the benchmark four-layer CNN in parentheses, which is adopted in \texttt{Opacus} library. Note that DP-BBP is not compatible to the outer product and it is not trivial to extend BBP to convolutional layers (see \cite{gal2015bayesian,shridhar2019comprehensive}).}
\label{acc_table}
\end{table}

\begin{figure}[!htb]
\minipage{0.33\textwidth}%
  \includegraphics[width=\linewidth]{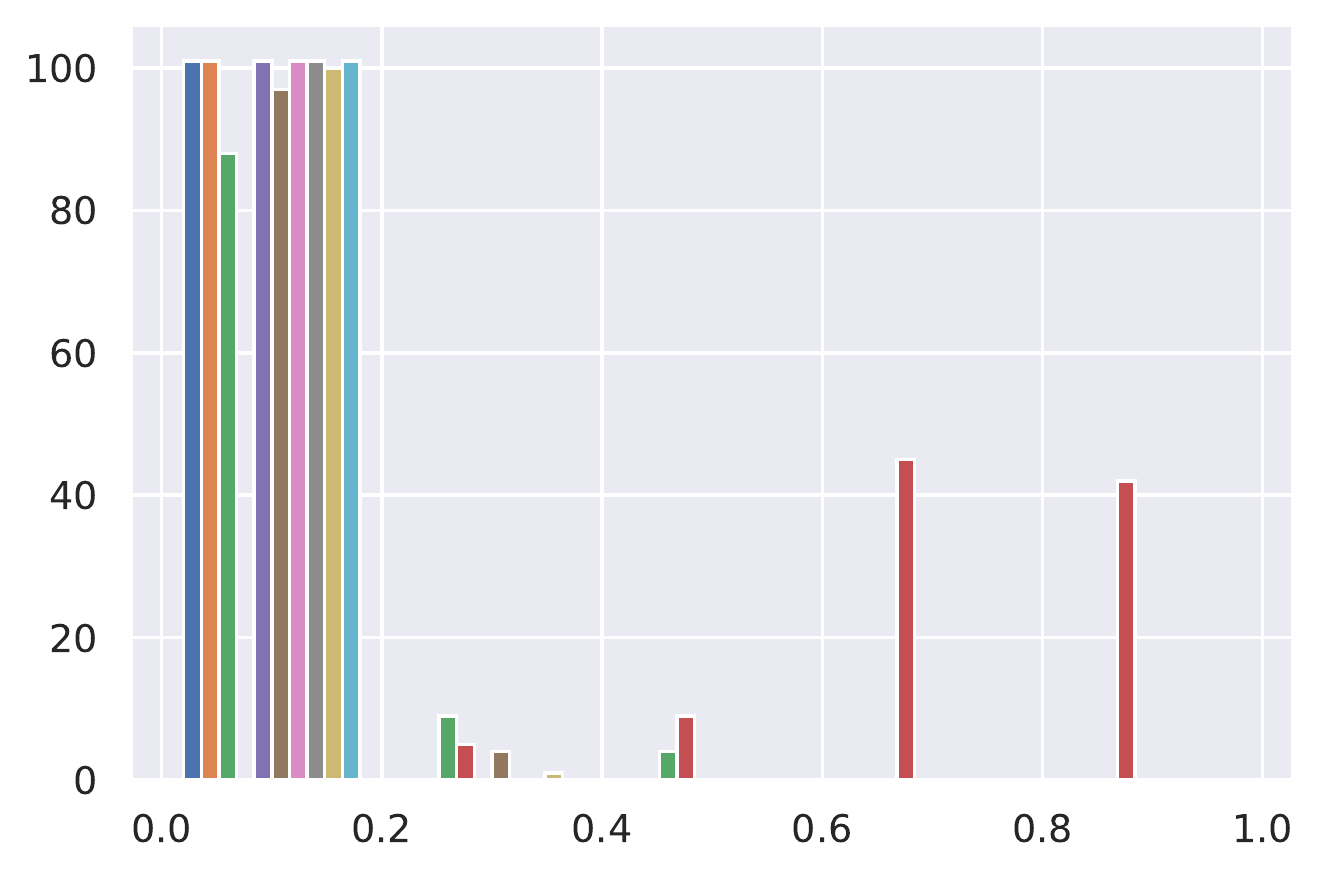}
\endminipage
\minipage{0.33\textwidth}%
  \includegraphics[width=\linewidth]{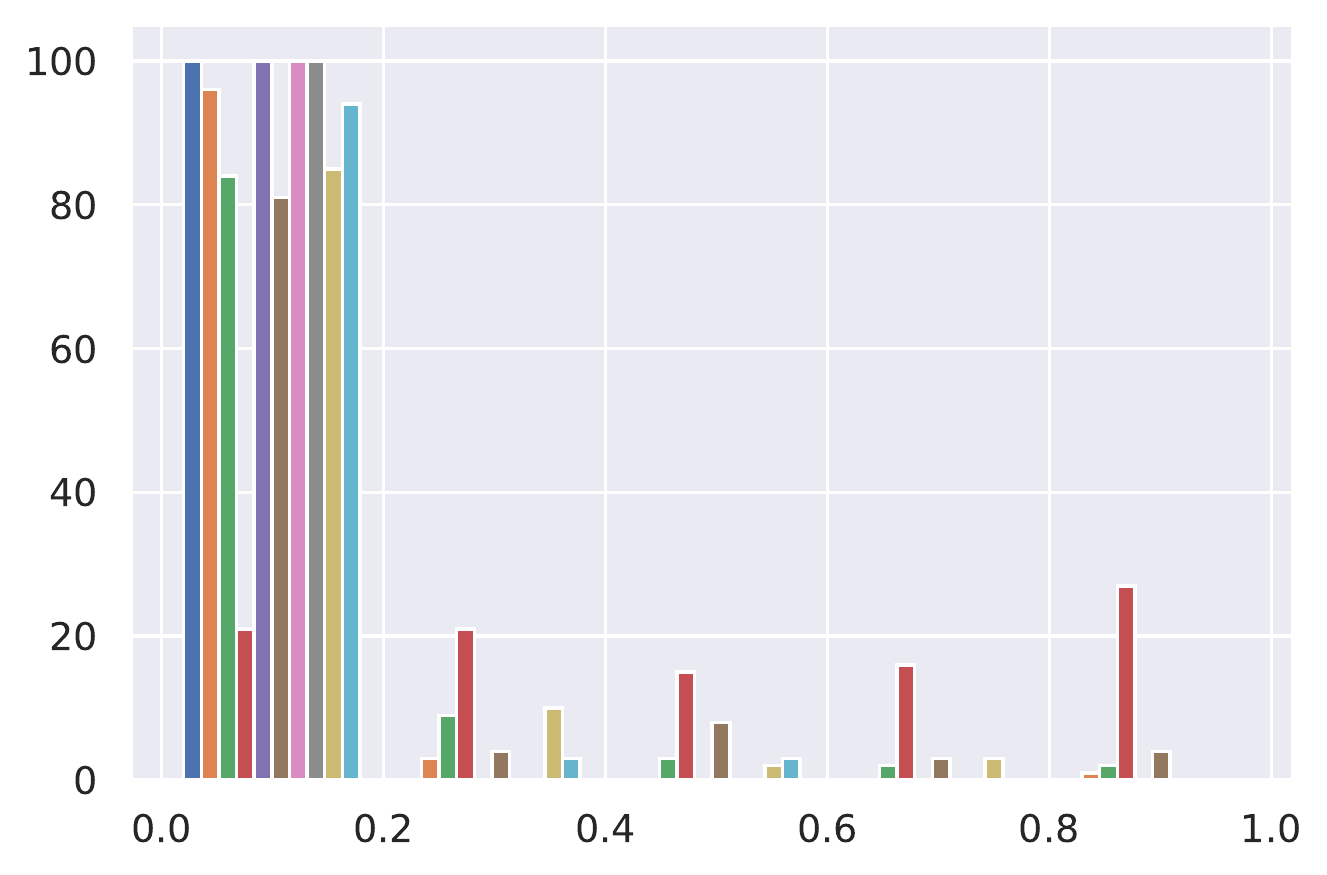}
\endminipage
\minipage{0.33\textwidth}
  \includegraphics[width=\linewidth]{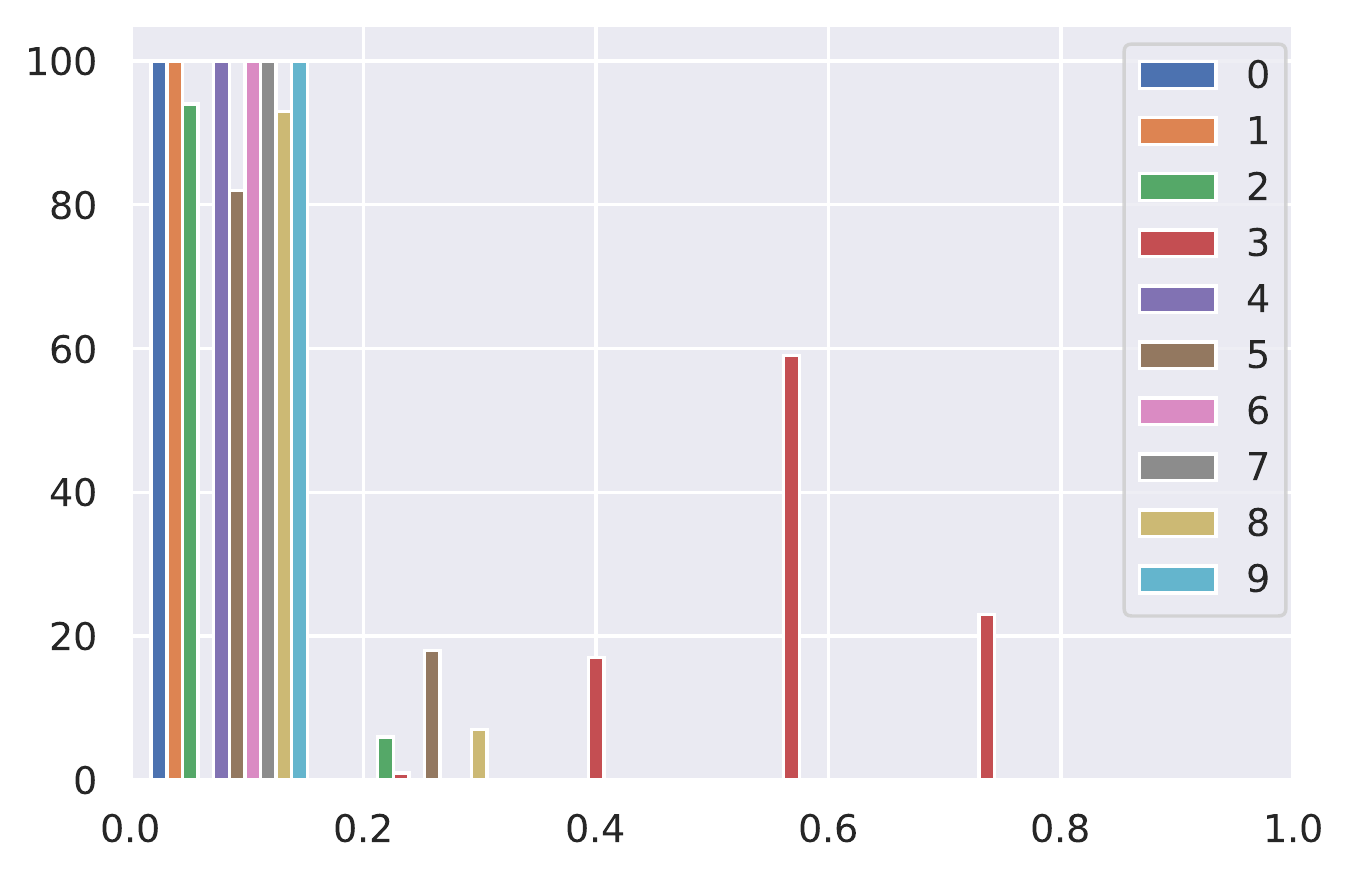}
\endminipage
\vfill
\minipage{0.33\textwidth}%
  \includegraphics[width=\linewidth]{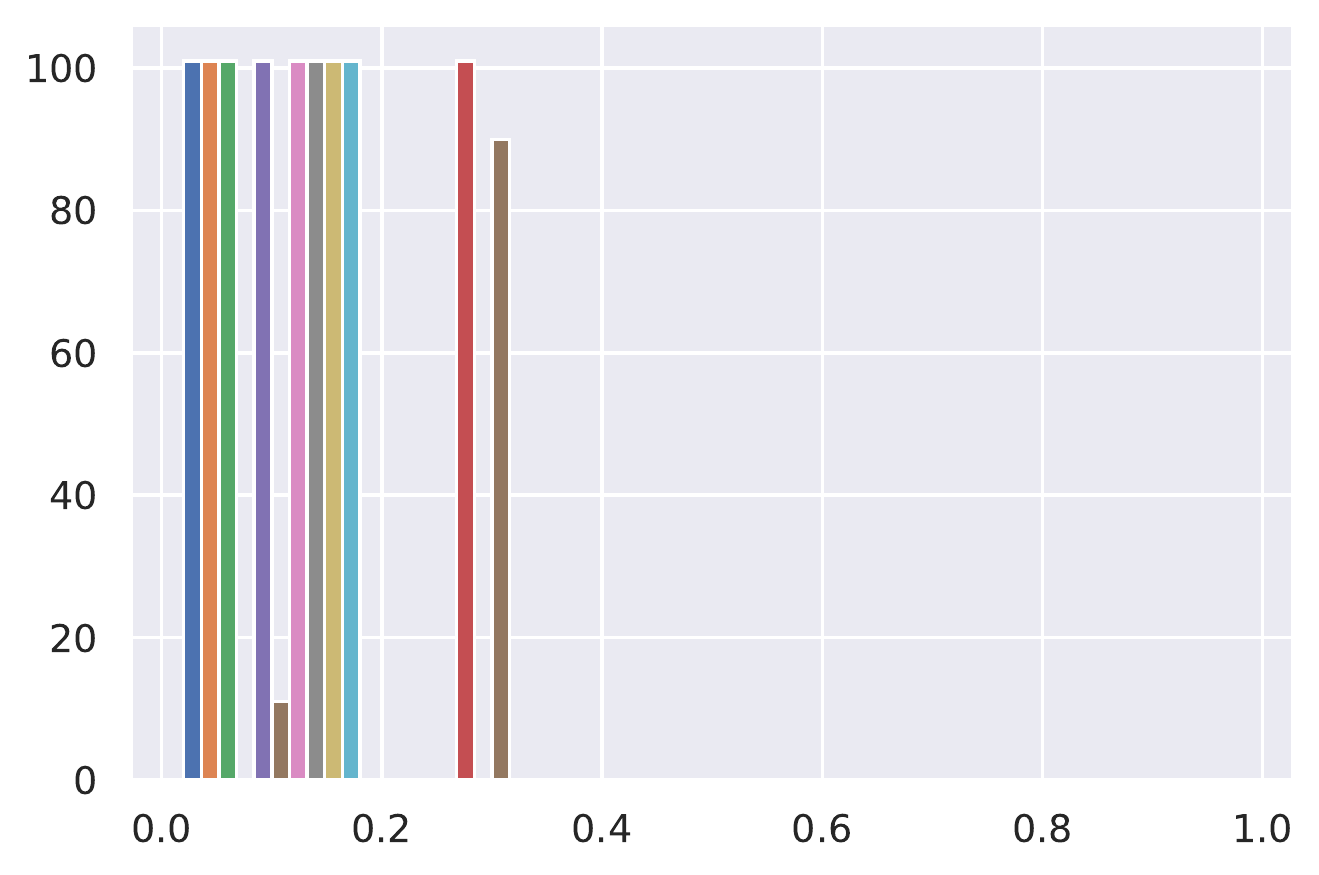}
\endminipage
\minipage{0.33\textwidth}%
  \includegraphics[width=\linewidth]{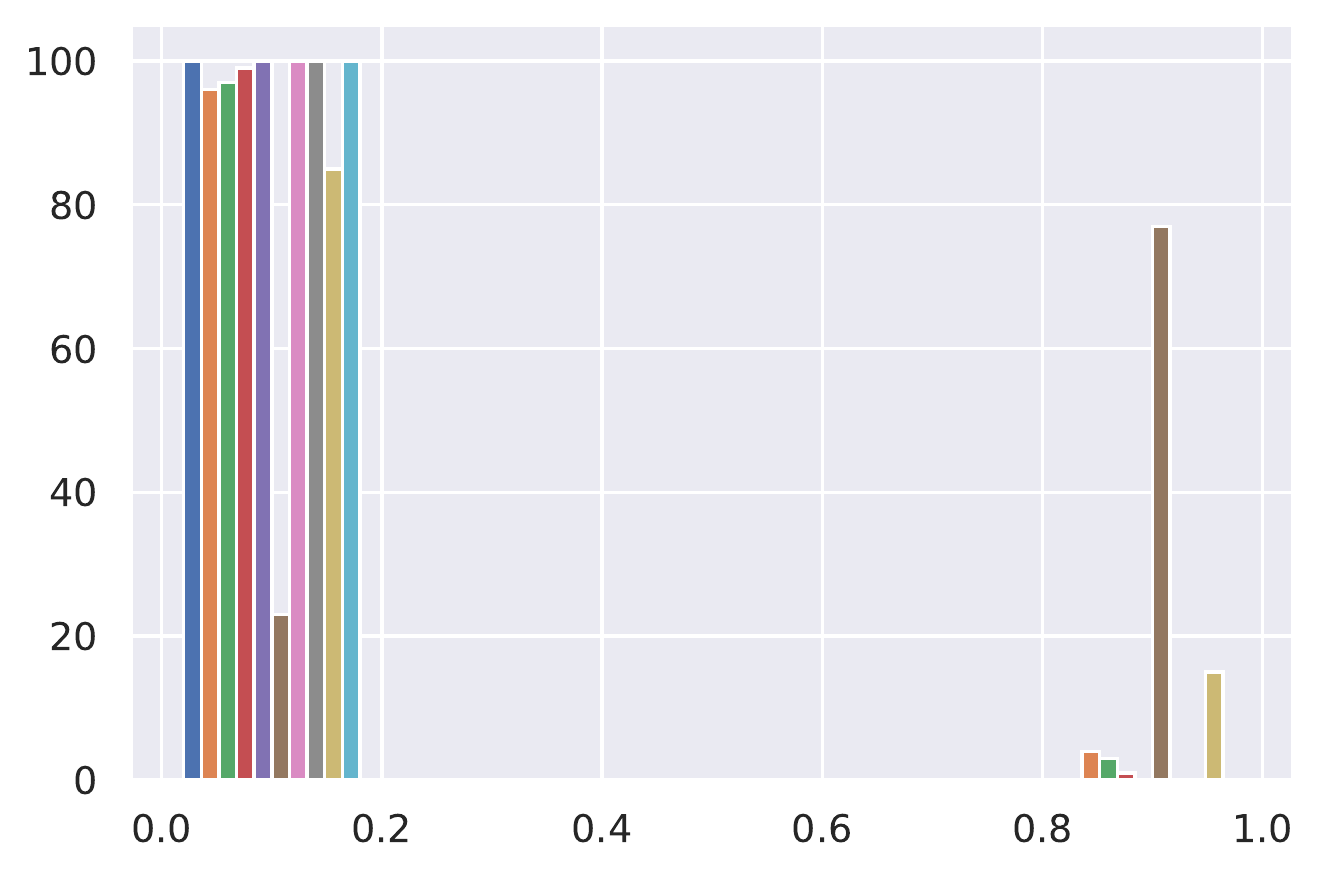}
\endminipage
\minipage{0.33\textwidth}
  \includegraphics[width=\linewidth]{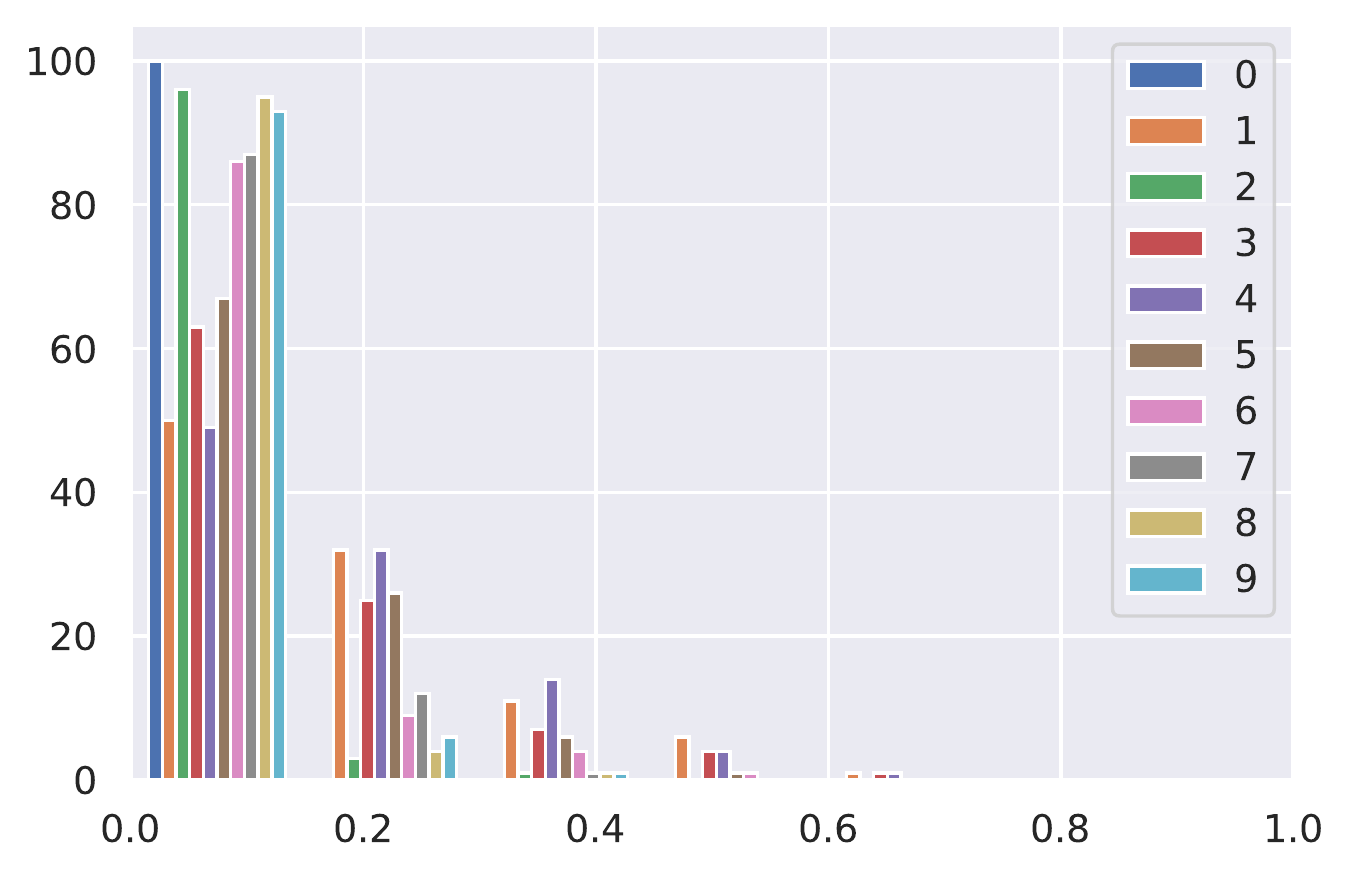}
\endminipage
\caption{Prediction distribution on MNIST with two-layer MLP over 100 repeated samplings (input image is $3$, see \Cref{Predicted_Labels}). Y-axis refers to the frequency and x-axis refers to the prediction probability. Left to right: SGLD, BBP, MC Dropout. Upper: non-DP BNNs. Lower: DP-BNNs.}
\label{label_3}
\end{figure}

\textbf{Uncertainty Quantification} 
Regarding uncertainty quantification, we visualize the empirical prediction posterior of Bayesian MLPs in \Cref{label_3} over 100 predictions on a single image. Note that at each probability (x-axis), we plot a cluster of bins each of which represents a class\footnote{Within each cluster, the bins can interchange the ordering. Thus the bin's x-coordinate is not meaningful and only the cluster's x-coordinate represents the prediction probability.}. For example, the left-most cluster represents \textit{not predicting} a class. Concretely, in the left-bottom plot, DP-SGLD has low red (class 3) and brown (class 5) bins on the left-most cluster, meaning it will predict 3 or 5. We see that non-DP BNNs usually predict correctly (with a low red bin in the left-most cluster), though the posterior probabilities of the correct class are different across three BNNs. Obviously, DP changes the empirical posterior probabilities significantly in distinct ways. First, all DP-BNNs are prone to make mistakes in prediction, e.g. both DP-SGLD and DP-BBP tend to predict class 5. In fact, DP-SGLD are equally likely to predict class 3 and 5 yet DP-BBP seldom predicts class 3 anymore, when DP is enforced. Additionally, DP-SGLD is less confident about its mistake compared to DP-BBP. This is indicated by the small x-coordinate of the right-most bins, and implies that DP-SGLD can be more calibrated, as discussed in the next paragraph. For MC Dropout, DP also reduces the confidence in predicting class 3 but the mistaken prediction spreads over several classes. Hence the quality of uncertainty quantification provided by DP-MC Dropout lies between that by DP-SGLD and DP-BBP.

\textbf{Calibration}
As a measure of the reliability, the calibration \cite{niculescu2005predicting,guo2017calibration} measures the distance between a classification model's accuracy and its prediction probability, i.e. confidence. Formally, denoting the vector of prediction probability for the $i$-th sample as $\bm\pi_i$, the \textit{confidence} for this sample is $\text{conf}_i=\max_k[\pi_i]_k$ and the prediction is $\text{pred}_i=\text{argmax}_k [\pi_i]_k$. Two commonly applied calibration errors are the expected calibration error (ECE) and the maximum calibration error (MCE). By splitting the predictions into $M$ equally-spaced bins $\{B_m\}$, we have
\begin{align*}
\text{ECE}&=\sum_{m\in[M]} \frac{|B_m|}{n}\Big|\text{acc}(B_m)-\text{conf}(B_m)\Big|,
\\
\text{MCE}&=\max_{m\in[M]} \Big|\text{acc}(B_m)-\text{conf}(B_m)\Big|,
\end{align*}
where acc is the average accuracy and conf is the average confidence within a bin.

Ideally, a reliable classifier should be calibrated in the sense that the accuracy matches the confidence. When a model is highly confident in its prediction yet it is not accurate, such classifier is over-confident; otherwise it is under-confident. It is well-known that the regular NNs are over-confident \cite{guo2017calibration,minderer2021revisiting} and (non-DP) BNNs are more calibrated \cite{maronas2020calibration}.
Recently, \cite{bu2021convergence} observe that DP non-Bayesian NNs can be even more over-confident than its non-DP counterparts. To mitigate the mis-calibration, the authors propose to apply a different per-sample gradient norm clipping, known as the global clipping, which is amazingly effective in learning calibrated models. This is orthogonal to our weight uncertainty approach as we use the classic clipping on BNNs. We believe it may be of independent interest to study the calibration of DP-BNNs with global clipping as a future direction.

In \Cref{table:cali MLP} and \Cref{table:cali CNN}, we again test the two-layer MLP and four-layer CNN on MNIST, with or without Gaussian prior under DP-BNNs regime. Notice that in the BNN regime, training with weight decay is equivalent to adopting a Gaussian prior, while training without weight decay is equivalent to using a non-informative prior.

\begin{table}[!htb]  
\centering 
\begin{tabular}{|c|c|c|c|c|} 
\hline  
 Methods  & DP-ECE & DP-MCE & Non-DP ECE &  Non-DP MCE 
\\ [0.2ex]  
\hline   
BBP (w/ prior) & 0.204& 0.641  & 0.024  & 0.052  \\ \hline  
BBP (w/o prior) & 0.167 & 0.141 & 0.166 & 0.166 \\ \hline
SGLD (w/ prior) & 0.007  & 0.175  & 0.035   & 0.175   \\
\hline  
SGLD (w/o prior) &  0.126 & 0.465  & 0.008  & 0.289  \\
\hline  
MC Dropout (w/ prior) &  0.008  & 0.080 
&0.030 & 0.041   \\
\hline 
MC Dropout (w/o prior) & 0.078    & 0.225 
&  0.002 & 0.725     \\
\hline  
SGD (w/ prior) &  0.013    & 0.089 
 & 0.016  &  0.139    \\
 \hline
 SGD (w/o prior) & 0.106   & 0.625 
 &0.005  & 0.299   \\
\hline 
\end{tabular}  
\caption{Calibration errors of DP-SGLD, DP-BBP, DP-MC Dropout, DP-SGD, and their non-DP counterparts on two-layer MLP. `prior' means Gaussian prior.}
\label{table:cali MLP}
\end{table} 
\begin{table}[!htb]  
\centering 
\begin{tabular}{|c|c|c|c|c|} 
\hline  
 Methods  & DP-ECE & DP-MCE & Non-DP ECE &  Non-DP MCE 
\\ [0.2ex]  
\hline  
SGLD (w/ prior) & 0.003  & 0.775 & 0.001   & 0.011   \\
\hline  
SGLD (w/o prior) &  0.043 & 0.371 & 0.006  & 0.219  \\
\hline  
MC Dropout (w/ prior) &  0.001  & 0.275
&0.030 & 0.325  \\
\hline 
MC Dropout (w/o prior) &0.033   & 0.230
&  0.003& 0.225  \\
\hline  
SGD (w/ prior) &  0.005   & 0.391
 & 0.002 &  0.059  \\
 \hline
 SGD (w/o prior) & 0.037  &0.365
 &0.014 & 0.325   \\
\hline 
\end{tabular}  
\caption{Calibration errors of DP-SGLD, DP-MC Dropout, DP-SGD, and their non-DP counterparts on four-layer CNN. `prior' means Gaussian prior.}
\label{table:cali CNN}
\end{table} 

\begin{figure}[!htb]
\minipage{0.33\textwidth}%
  \includegraphics[width=\linewidth]{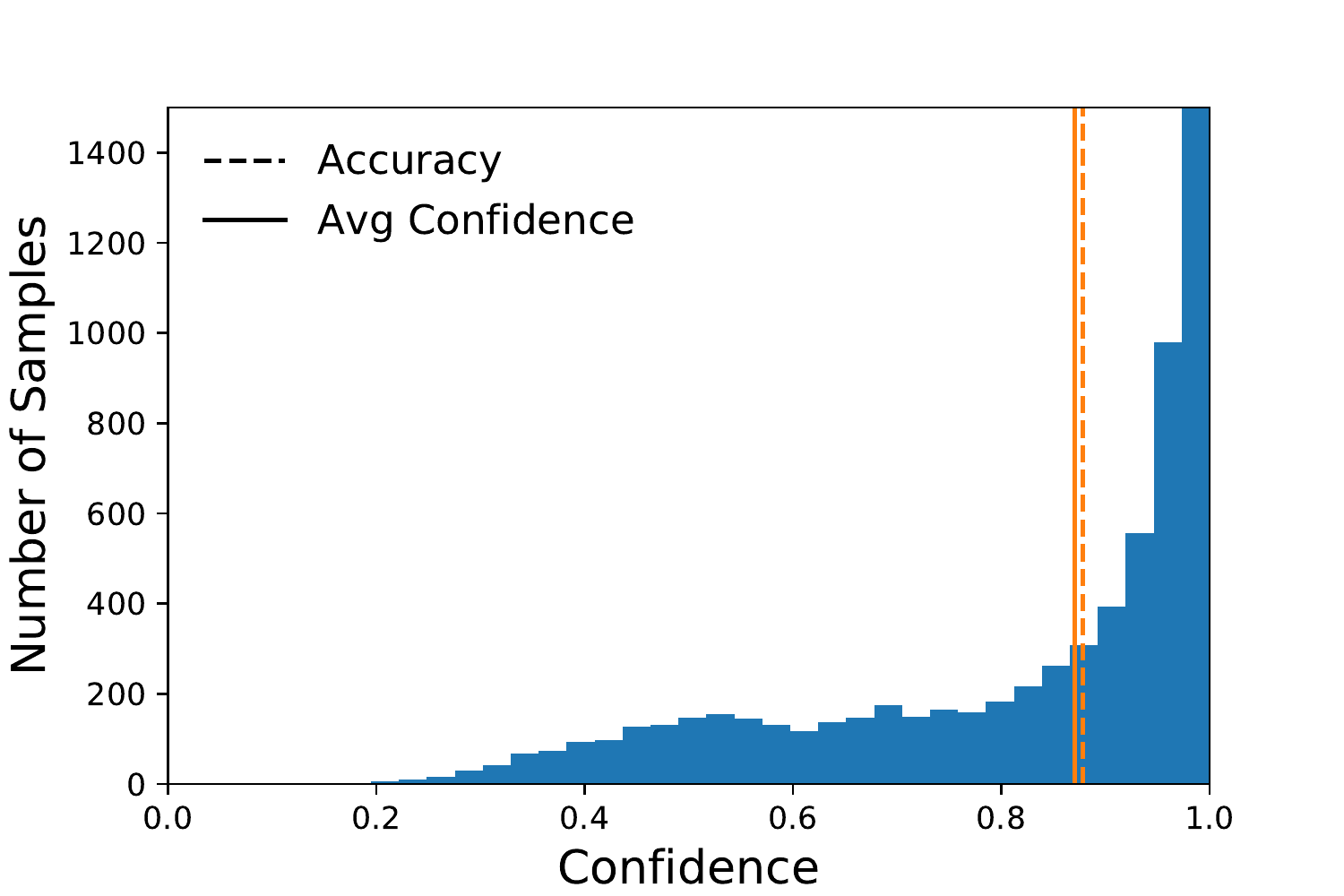}
\endminipage
\minipage{0.33\textwidth}
  \includegraphics[width=\linewidth]{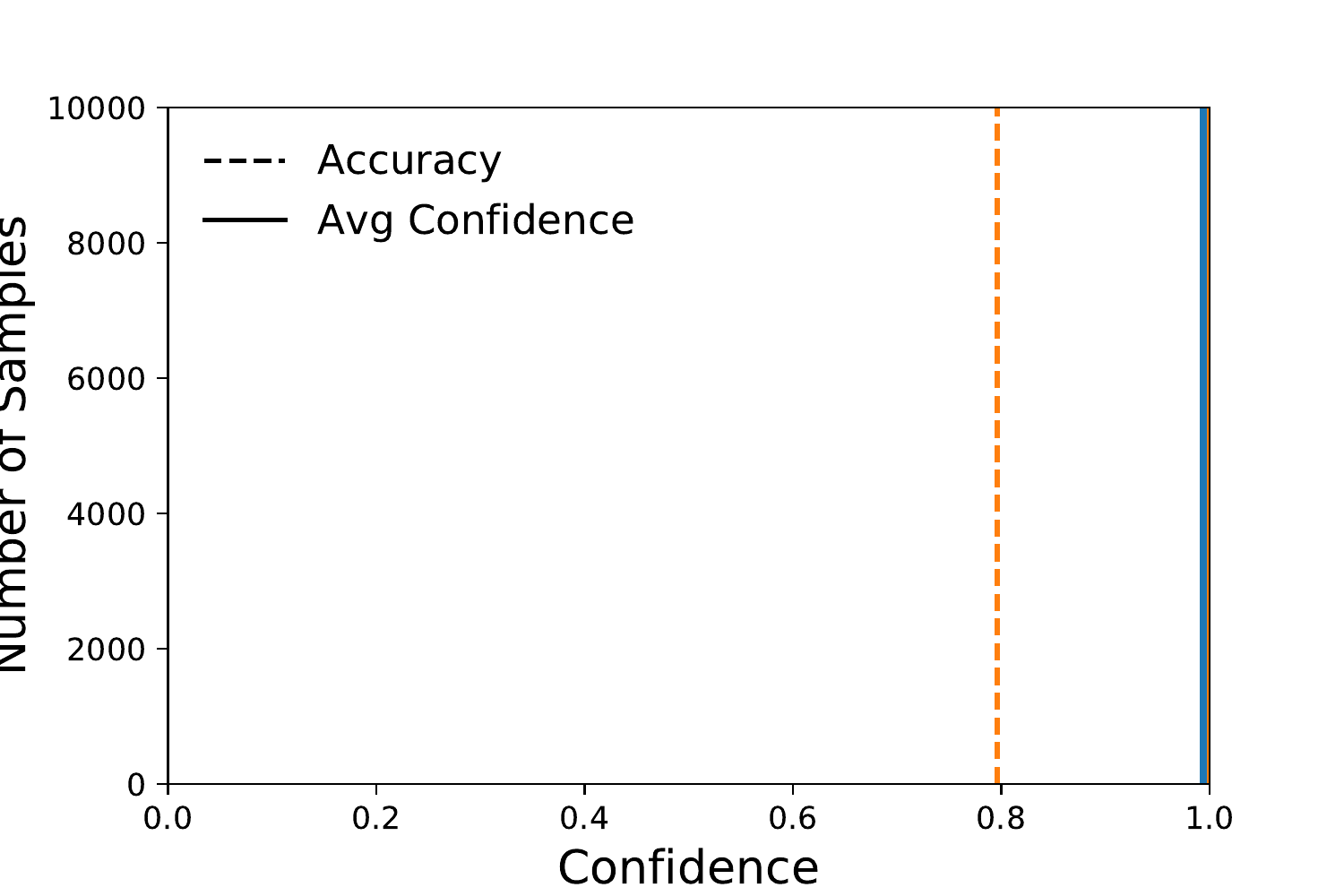}
\endminipage
\minipage{0.33\textwidth}%
  \includegraphics[width=\linewidth]{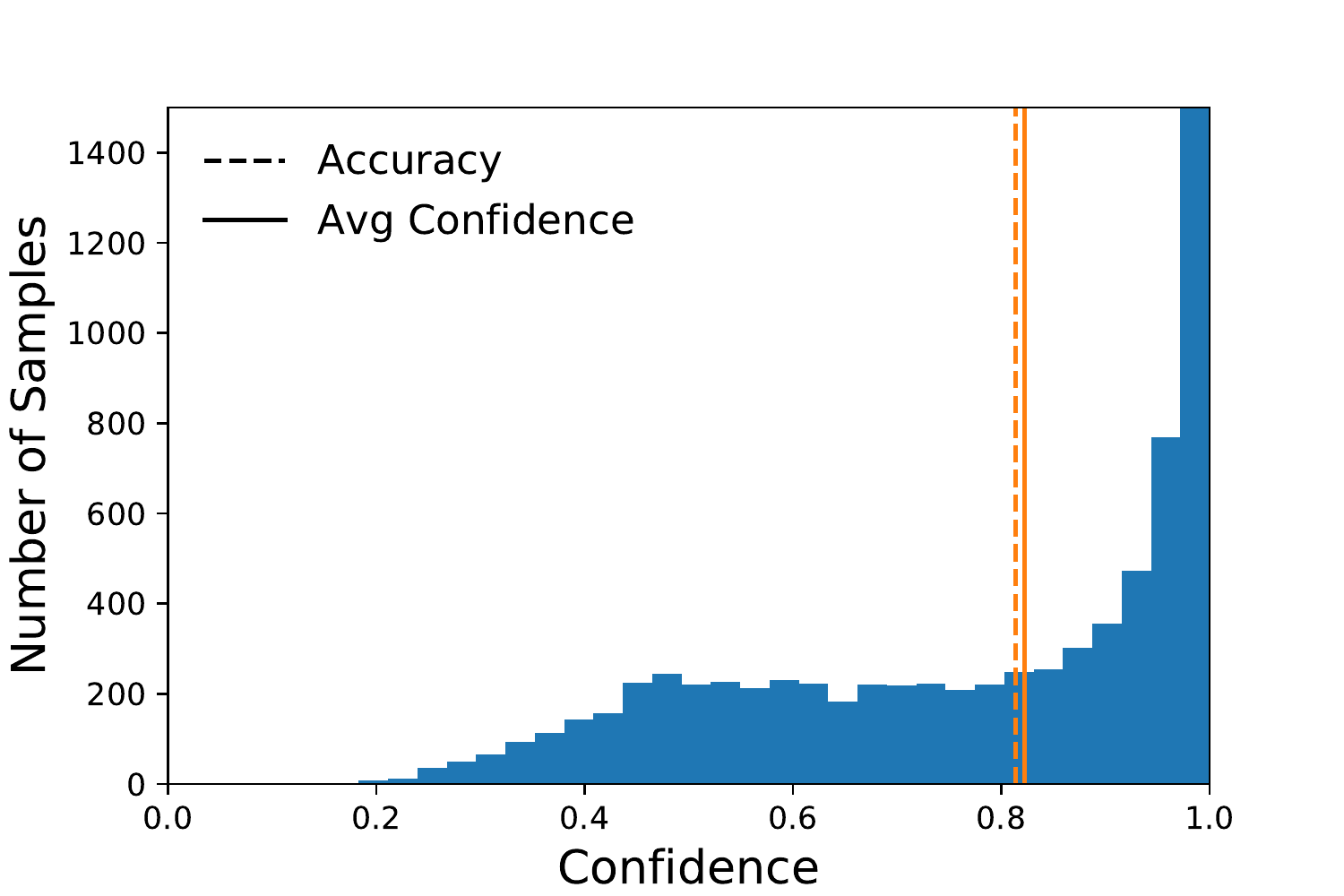}
\endminipage
\vfill
\minipage{0.33\textwidth}%
  \includegraphics[width=\linewidth]{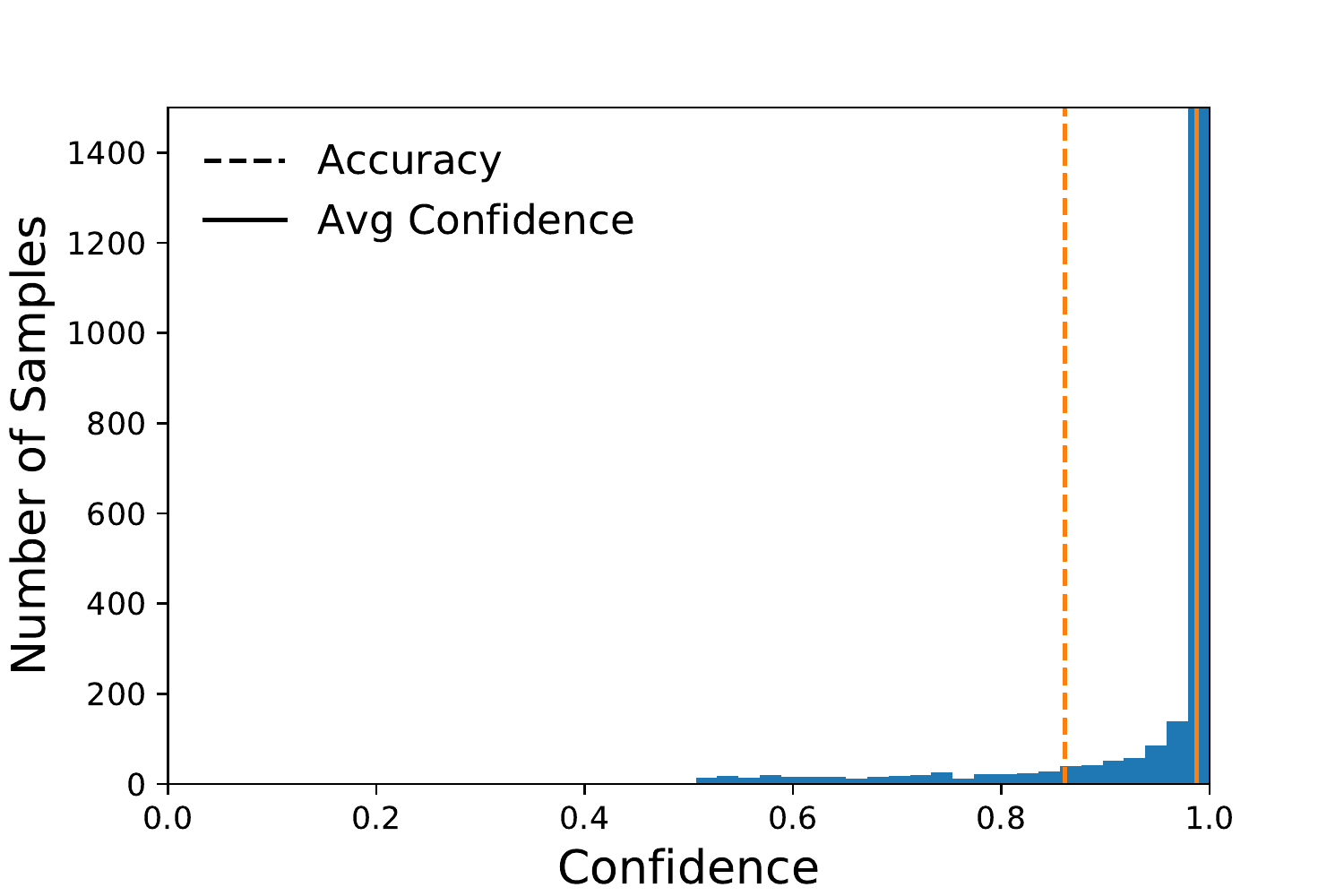}
\endminipage
\minipage{0.33\textwidth}
  \includegraphics[width=\linewidth]{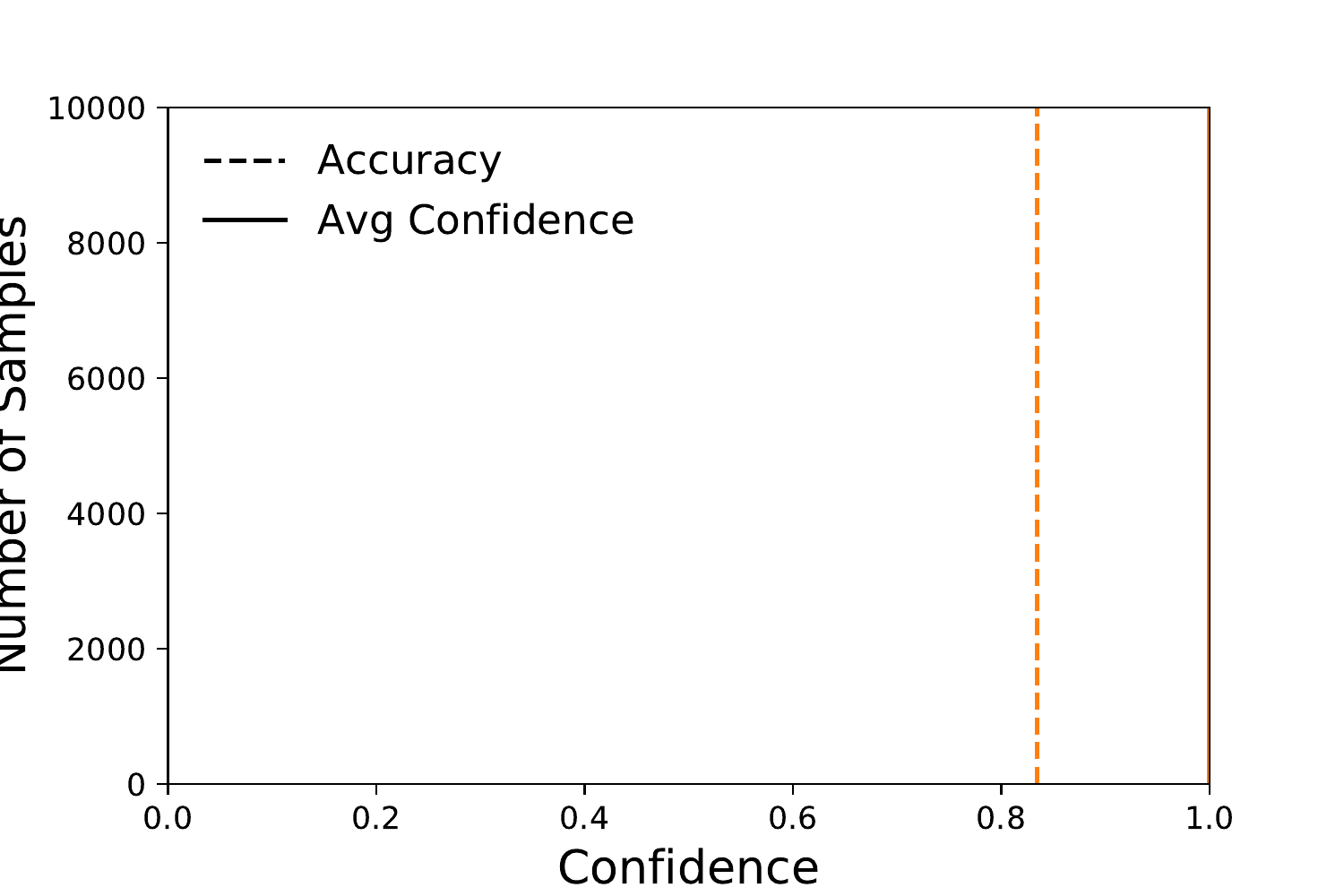}
\endminipage
\minipage{0.33\textwidth}%
  \includegraphics[width=\linewidth]{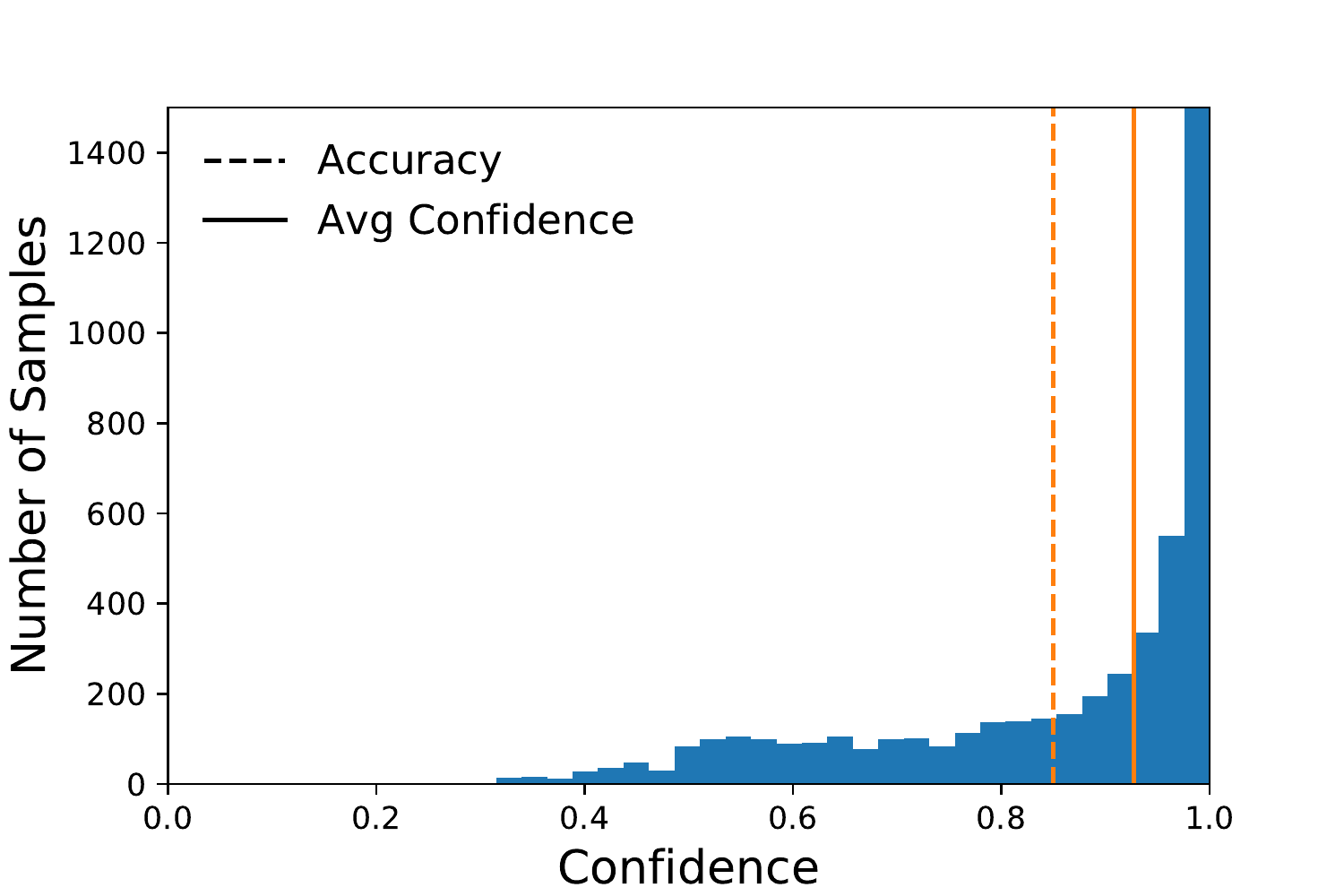}
\endminipage
\vspace{-0.2cm}
\caption{Confidence histogram on MNIST with two-layer MLP under DP regime. Left to right: SGLD, BBP, MC Dropout. Upper: with Gaussian prior. Lower: without Gaussian prior.}
\label{fig: w/o wd histogram}
\end{figure}

\begin{figure}[!htb]
\vspace{-0.1cm}
\minipage{0.33\textwidth}%
  \includegraphics[width=\linewidth]{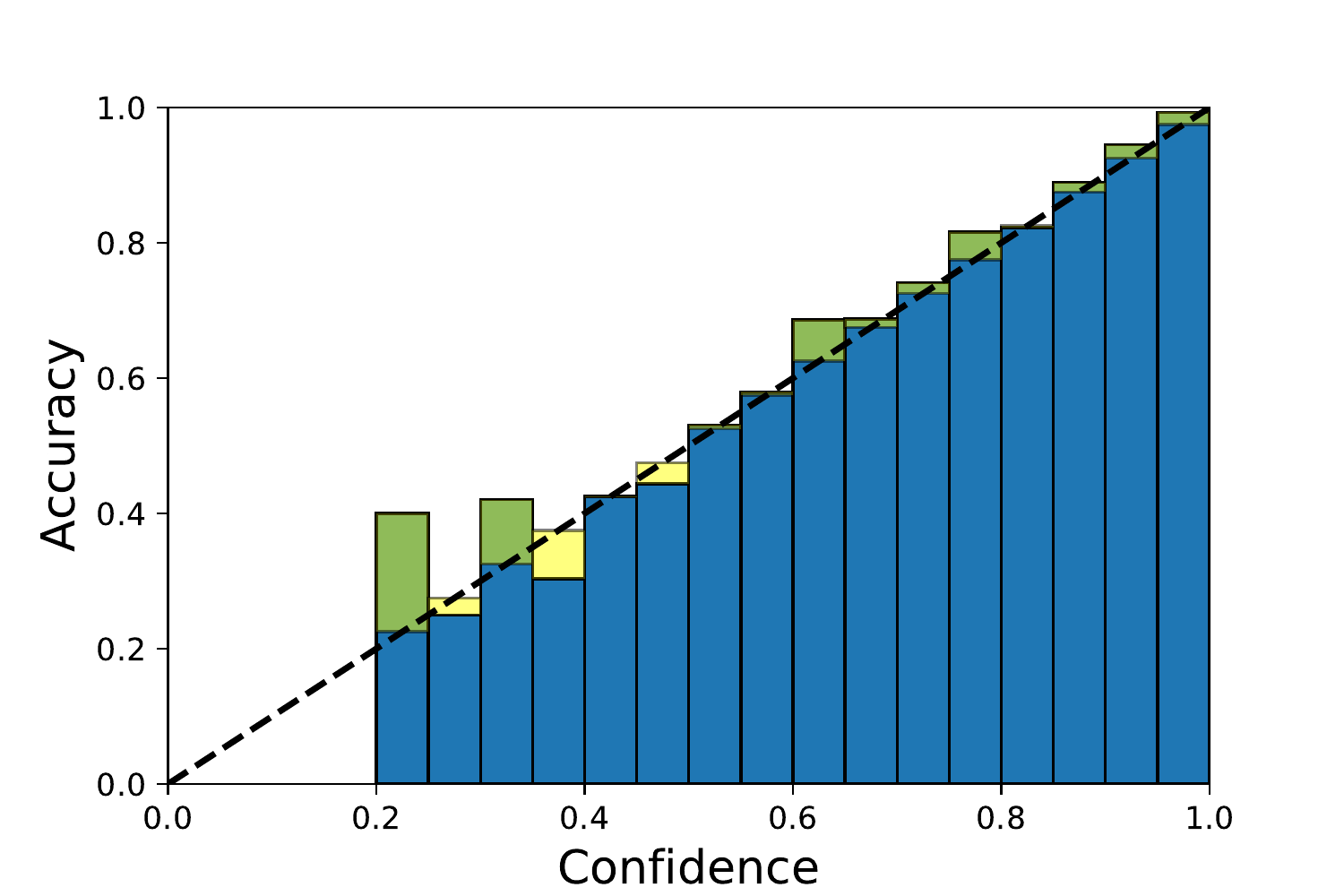}
\endminipage
\minipage{0.33\textwidth}
  \includegraphics[width=\linewidth]{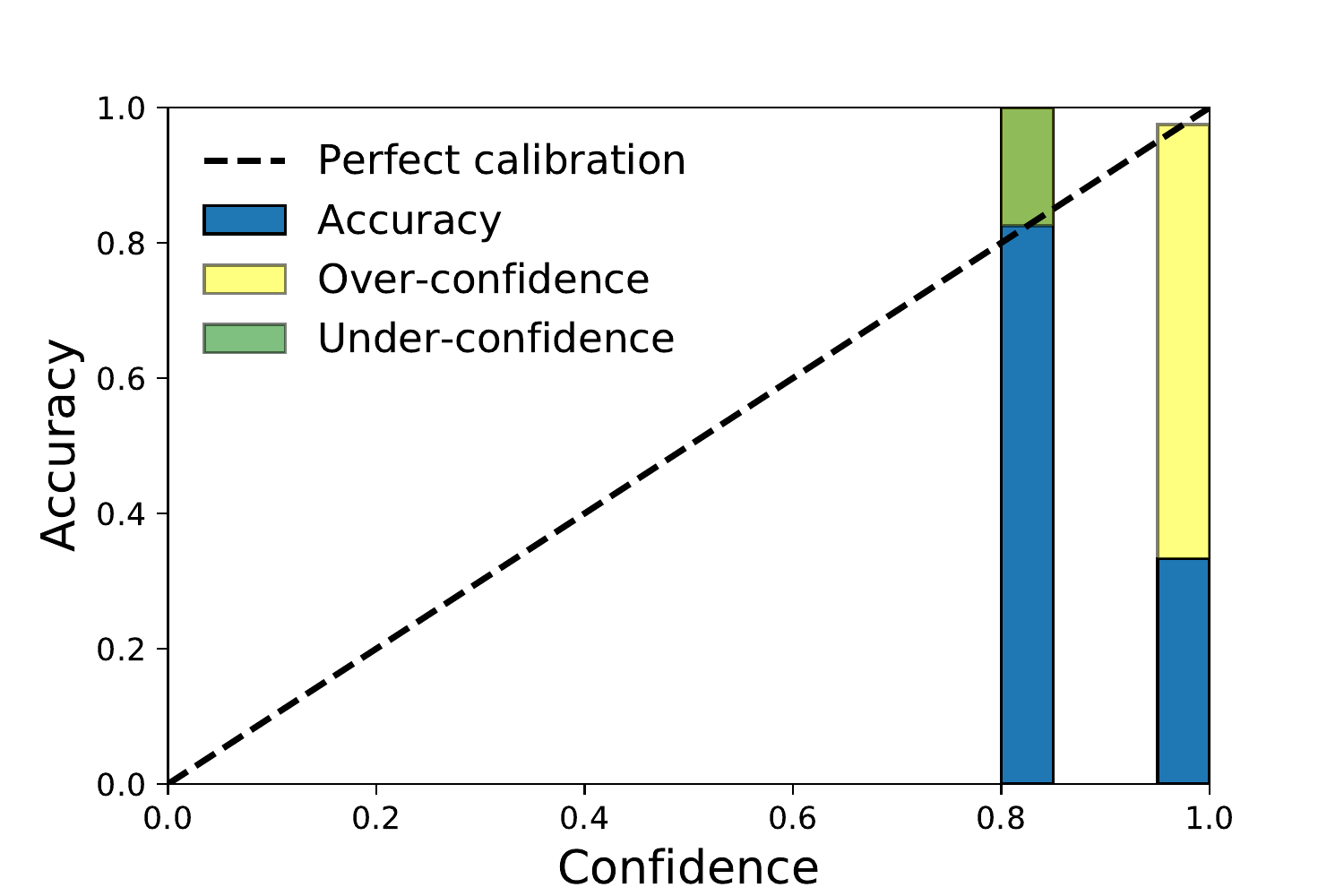}
\endminipage
\minipage{0.33\textwidth}%
  \includegraphics[width=\linewidth]{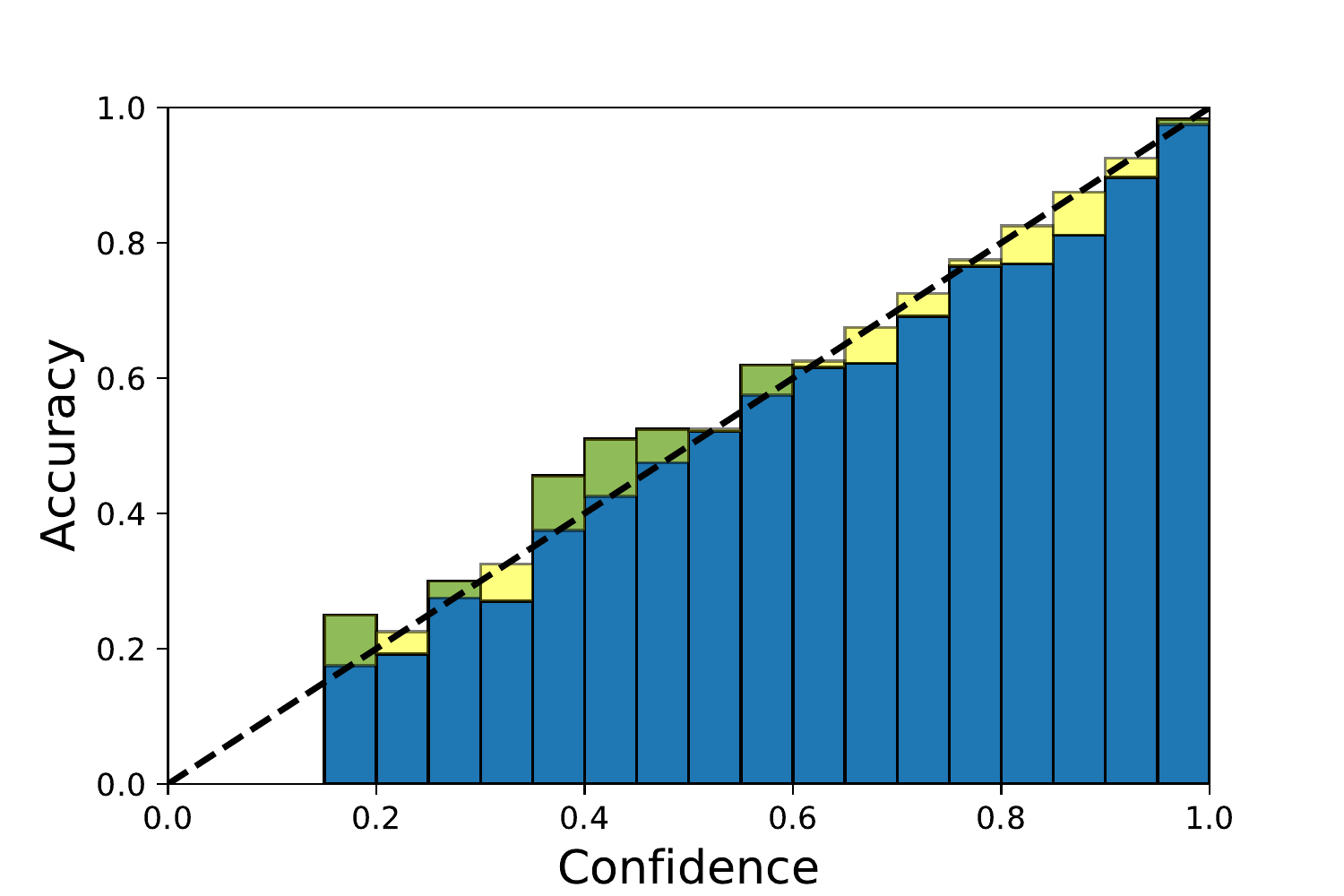}
\endminipage
\vfill
\minipage{0.33\textwidth}%
  \includegraphics[width=\linewidth]{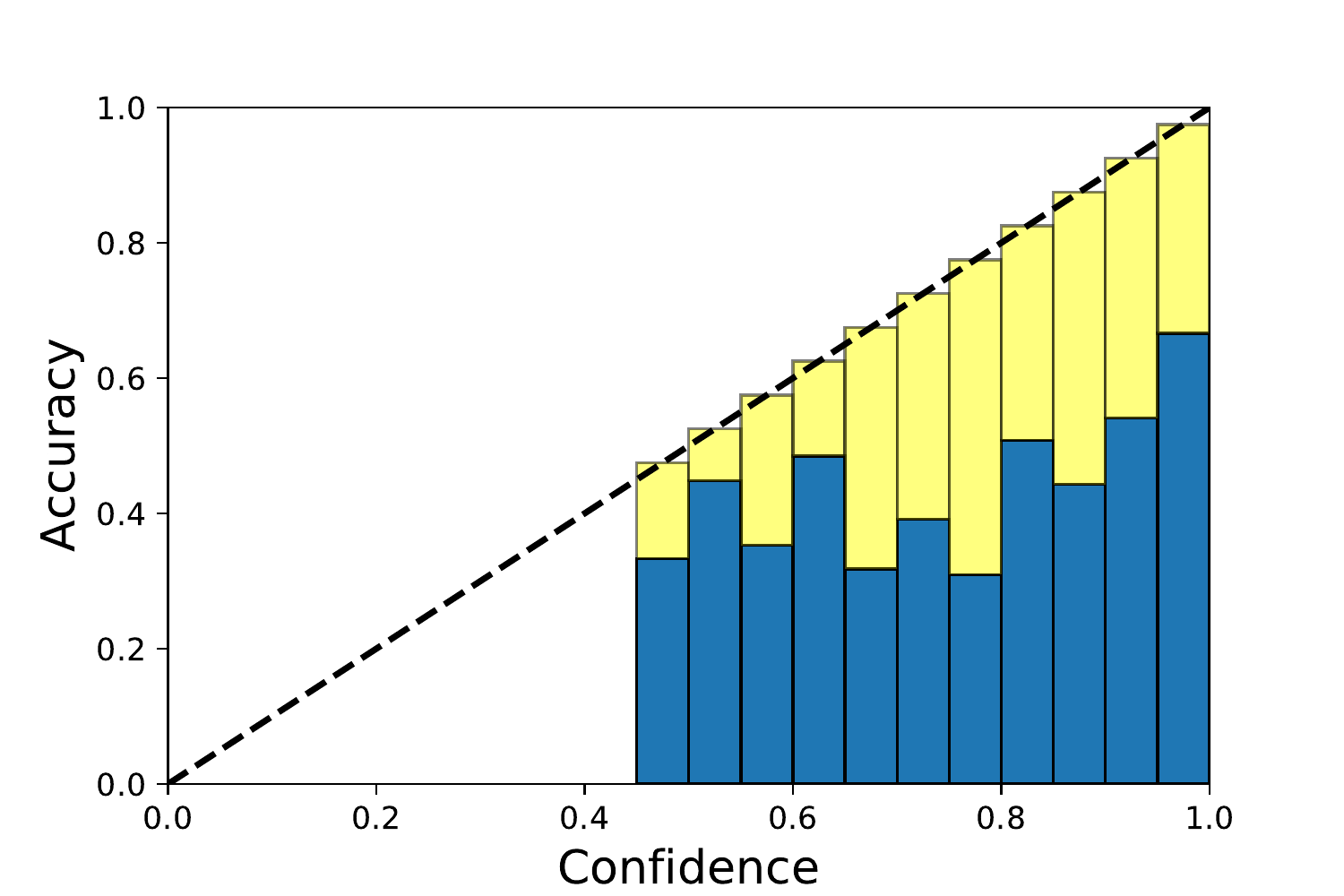}
\endminipage
\minipage{0.33\textwidth}
  \includegraphics[width=\linewidth]{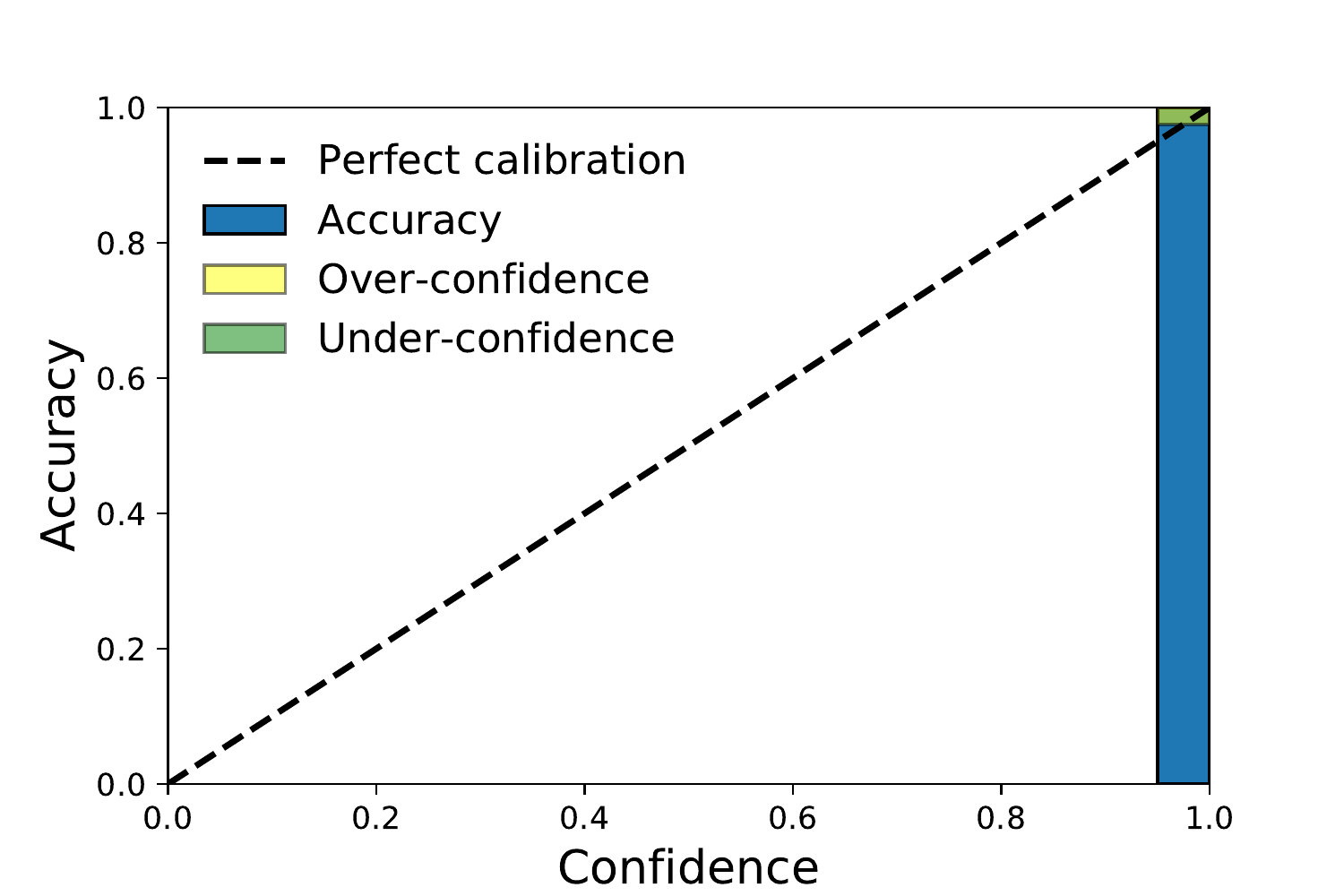}
\endminipage
\minipage{0.33\textwidth}%
  \includegraphics[width=\linewidth]{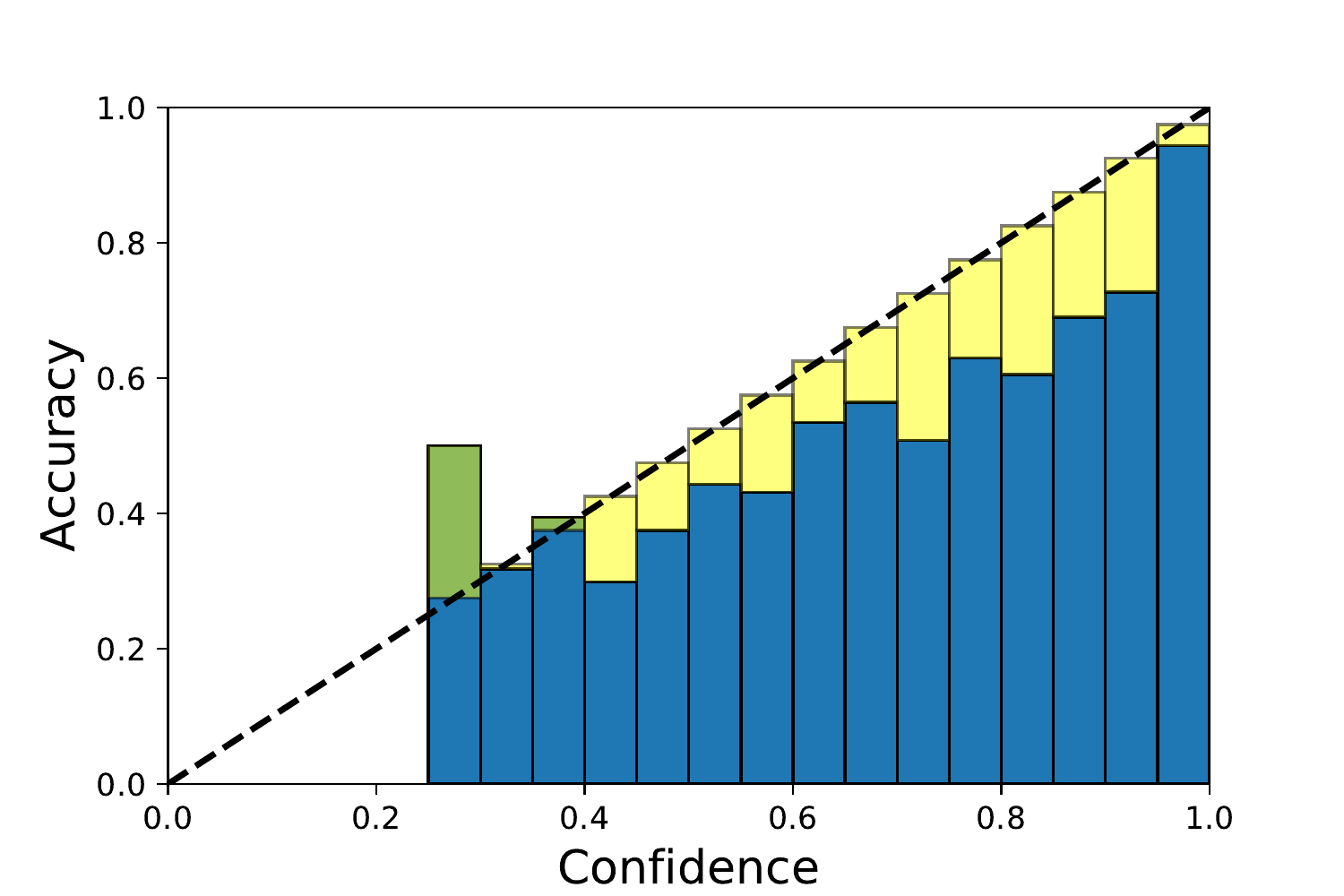}
\endminipage
\vspace{-0.2cm}
\caption{Reliability diagram on MNIST with two-layer MLP under DP regime. Left to right: SGLD, BBP, MC Dropout. Upper: with Gaussian prior. Lower: without Gaussian prior.}
\label{fig: w/o wd diagram}
\end{figure}

On MLP, the Gaussian prior (or weight decay) significantly improves the MCE, in the non-DP regime and furthermore in the DP regime. See \Cref{fig: w/o wd histogram} and \Cref{fig: w/o wd diagram}. However, on CNN, while the Gaussian prior helps in the non-DP regime, this may not hold true in the DP regime. 

\begin{figure}[!htb]
\vspace{-0.1cm}
\minipage{0.33\textwidth}%
  \includegraphics[width=\linewidth]{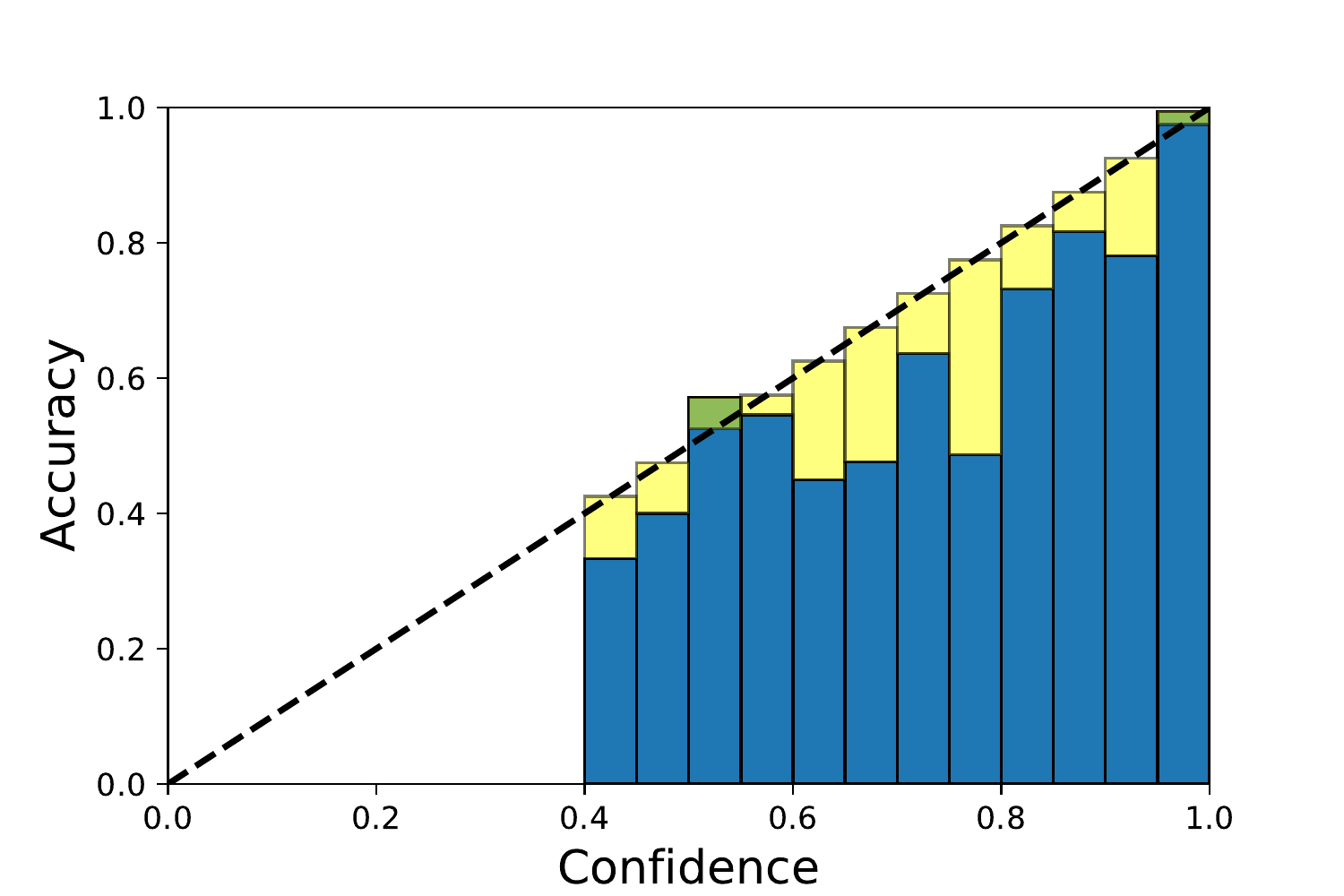}
\endminipage
\minipage{0.33\textwidth}
  \includegraphics[width=\linewidth]{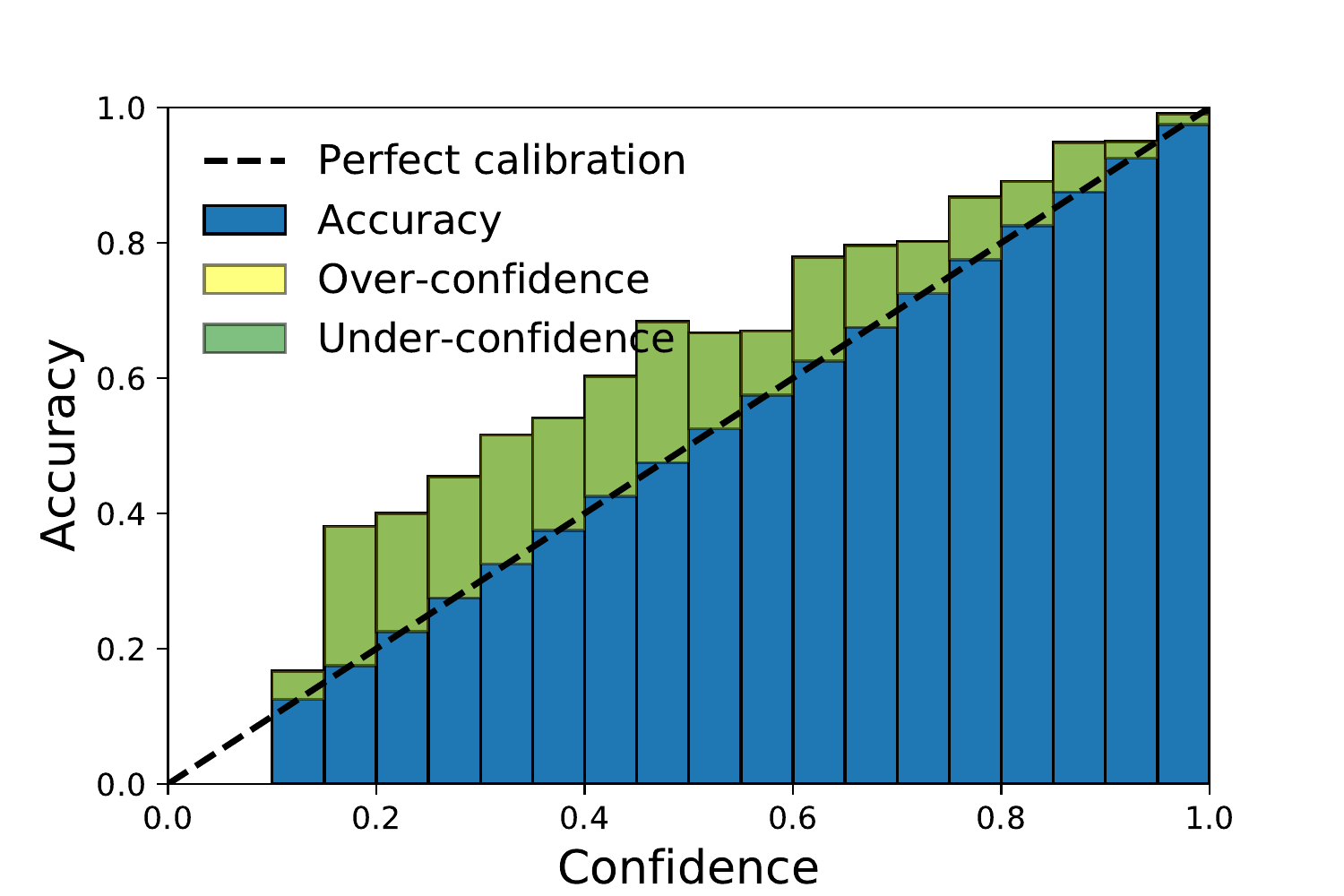}
\endminipage
\minipage{0.33\textwidth}%
  \includegraphics[width=\linewidth]{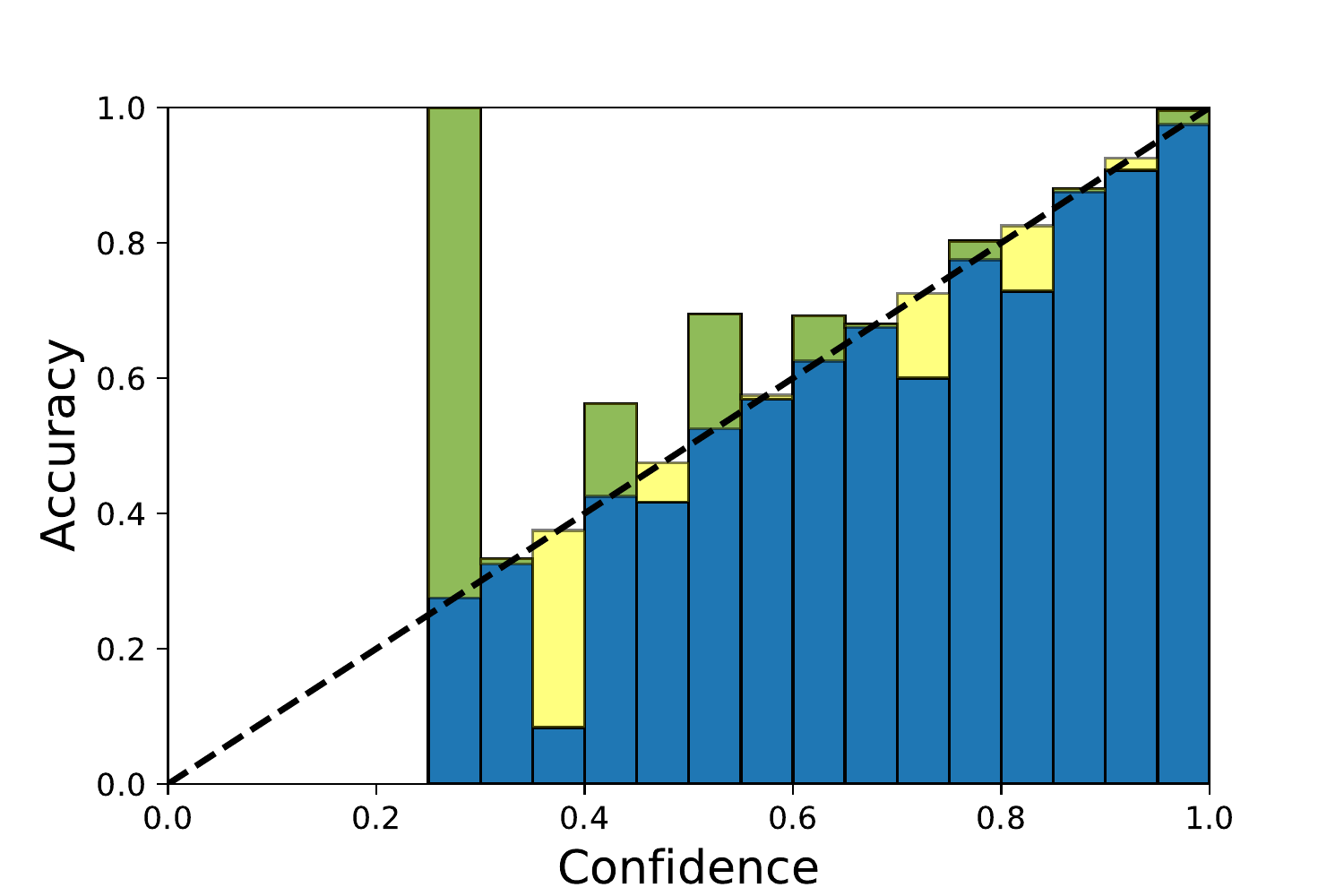}
\endminipage
\vspace{-0.2cm}
\caption{Reliability diagram on MNIST with two-layer MLP under non-DP regime without prior. Left to right: SGLD, BBP, MC Dropout.}
\label{fig: w/o DP diagram}
\end{figure}

For both neural network structures, DP exacerbates the calibration: leading to worse MCE when the non-informative prior is used. See lower panel of \Cref{fig: w/o wd diagram} and \Cref{fig: w/o DP diagram}. However, this is usually not the case when DP is guaranteed under the Gaussian prior.

\begin{figure}[!htb]
\centering
\minipage{0.33\textwidth}%
  \includegraphics[width=\linewidth]{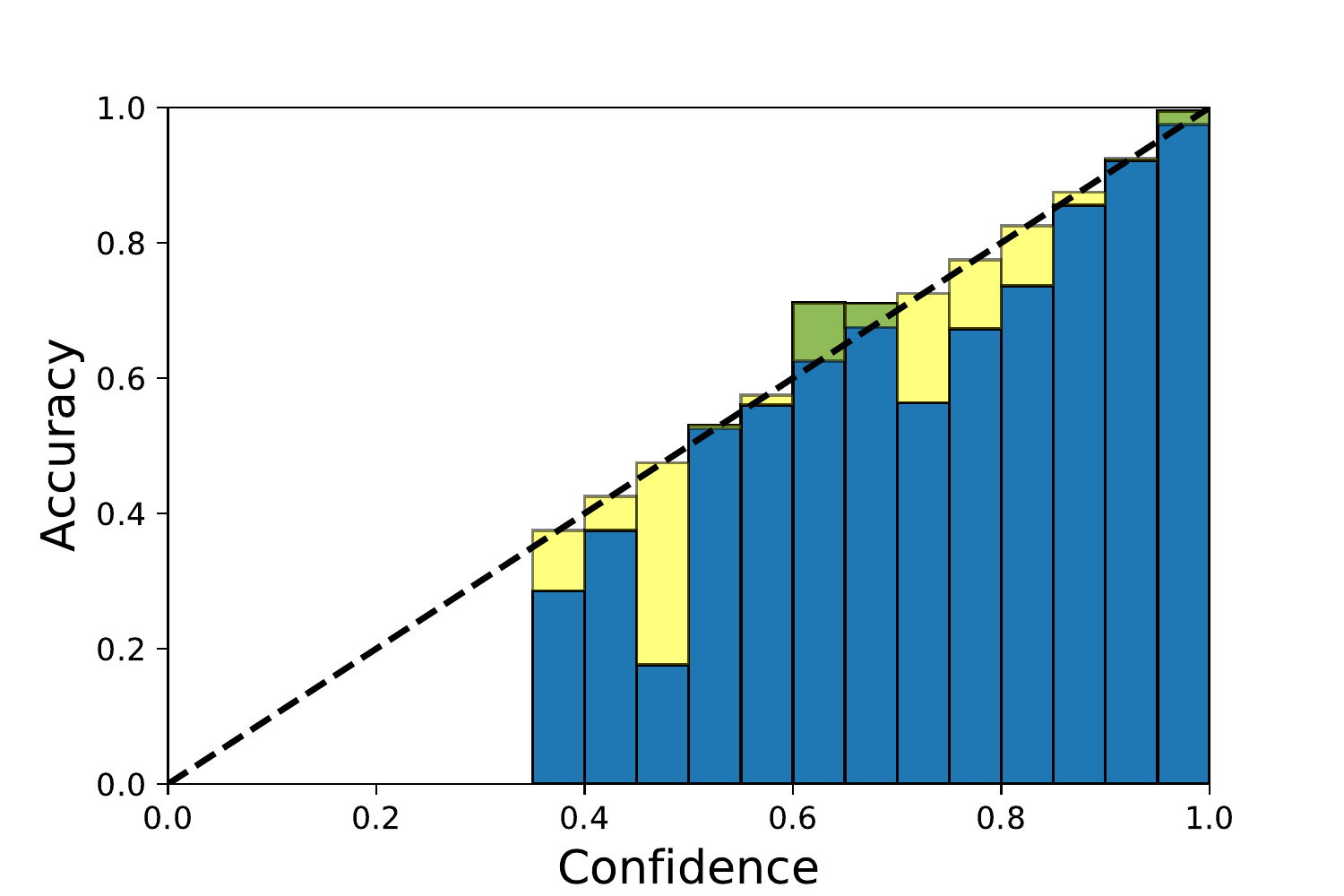}
\endminipage
$\overset{\text{ DP }}{\longrightarrow}$
\minipage{0.33\textwidth}%
  \includegraphics[width=\linewidth]{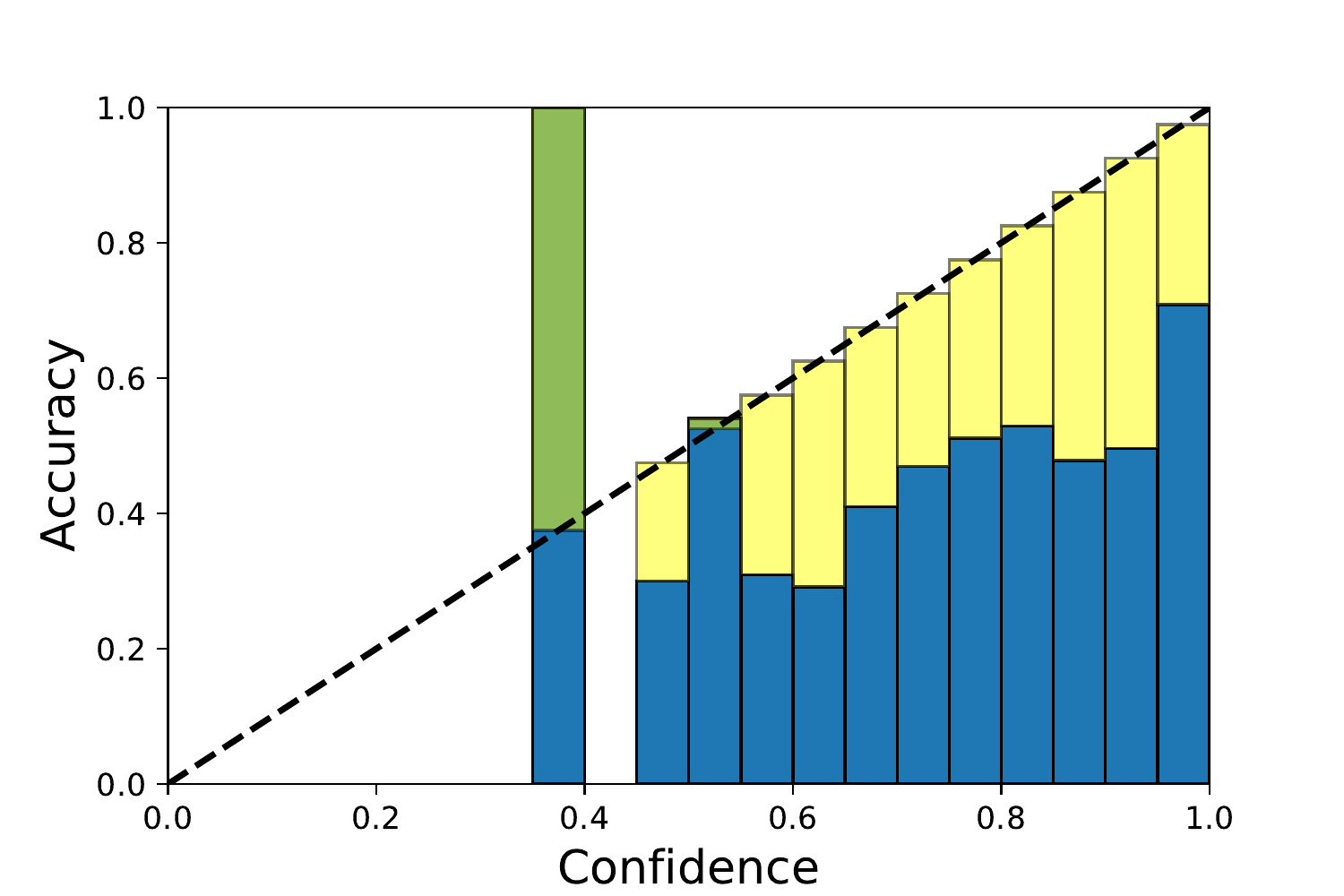}
\endminipage
\vfill
\quad\quad\quad\quad\quad\quad\quad\quad\quad\quad\quad\quad\quad\quad\quad
{\scriptsize BNN} $\big\downarrow$
\vfill
\minipage{0.33\textwidth}%
  \includegraphics[width=\linewidth]{dpsgld2lwd_local.pdf}
\endminipage
$\overset{\text{prior}}{\longleftarrow}$
\minipage{0.33\textwidth}%
  \includegraphics[width=\linewidth]{dpsgld2lnwd_local.pdf}
\endminipage
\vspace{-0.2cm}
\caption{Reliability diagram on MNIST with two-layer MLP. Upper left: SGD without prior (non-Bayesian, non-DP). Upper right: DP-SGD without prior (non-Bayesian). Lower right: DP-SGLD without prior. Lower left: DP-SGLD with Gaussian prior.}
\label{fig: w/o BNN diagram}
\end{figure}

Additionally, BNNs often enjoy smaller MCE than the regular MLP but may have larger MCE than the regular CNN. In the case of SGLD, the effect of DP-BNN and prior distribution is visualized in \Cref{fig: w/o BNN diagram}. In the MNIST experiment, the ECE is relatively small and thus the effects of using DP and/or Bayesian methods are less clear.

\subsection{Regression on heteroscedastic synthetic data}
\label{sec:regression}

We compare the prediction uncertainty of BNNs on the heteroscedastic data generated from Gaussian process (see details in \Cref{appendix:experiment}). Here, the prediction uncertainty for each data point is estimated by the empirical posterior over $1000$ predictions. Specifically, the prediction uncertainty can be decomposed into the posterior uncertainty (also called epistemic uncertainty, the blue region) and the data uncertainty (also called aleatoric uncertainty, the orange region), whose mathematical formulation is delayed in \Cref{appendix:experiment}. In \Cref{regression}, all three non-DP BNNs (upper panel) characterize similar prediction uncertainty, regarded as the benchmark truth. 

In our experiments, we train all BNNs with DP-GD for 200 epochs and noise multiplier such that the DP is $\epsilon=4.21, \delta=1/250$. SGLD is surprisingly accurate in both DP and non-DP scenarios while BBP and MC Dropout suffer notably from DP, even though their non-DP versions are accurate.

Clearly, the prediction uncertainty of SGLD and BBP are barely affected by DP; additionally, given that DP-SGLD has much better MSE, this experiment confirms that DP-SGLD is more desirable for uncertainty quantification with DP guarantee. Unfortunately, for MC Dropout, DP leads to substantially greater posterior uncertainty and unstable mean prediction. The resulting wide out-of-sample predictive intervals provide little information.

\begin{table}[!htb]  
\centering 
\begin{tabular}{|c|c|c|c|} 
\hline  
 Methods  & SGLD & BBP  & MC Dropout  
\\ [0.2ex]  
\hline   
DP & 0.510& 1.276  & 0.682   \\ \hline  
Non-DP & 0.523& 0.562  & 0.591   \\  
\hline 
\end{tabular}  
\caption{Mean square error of heteroscedasticity regression with Gaussian prior. The reported error is the median over 20 independent simulations.}
\label{acc_table_reg}
\end{table}

\begin{figure}[!htb]
\vspace{-0.3cm}
\minipage{0.3\textwidth}%
\includegraphics[width=\linewidth,height=0.75\linewidth]{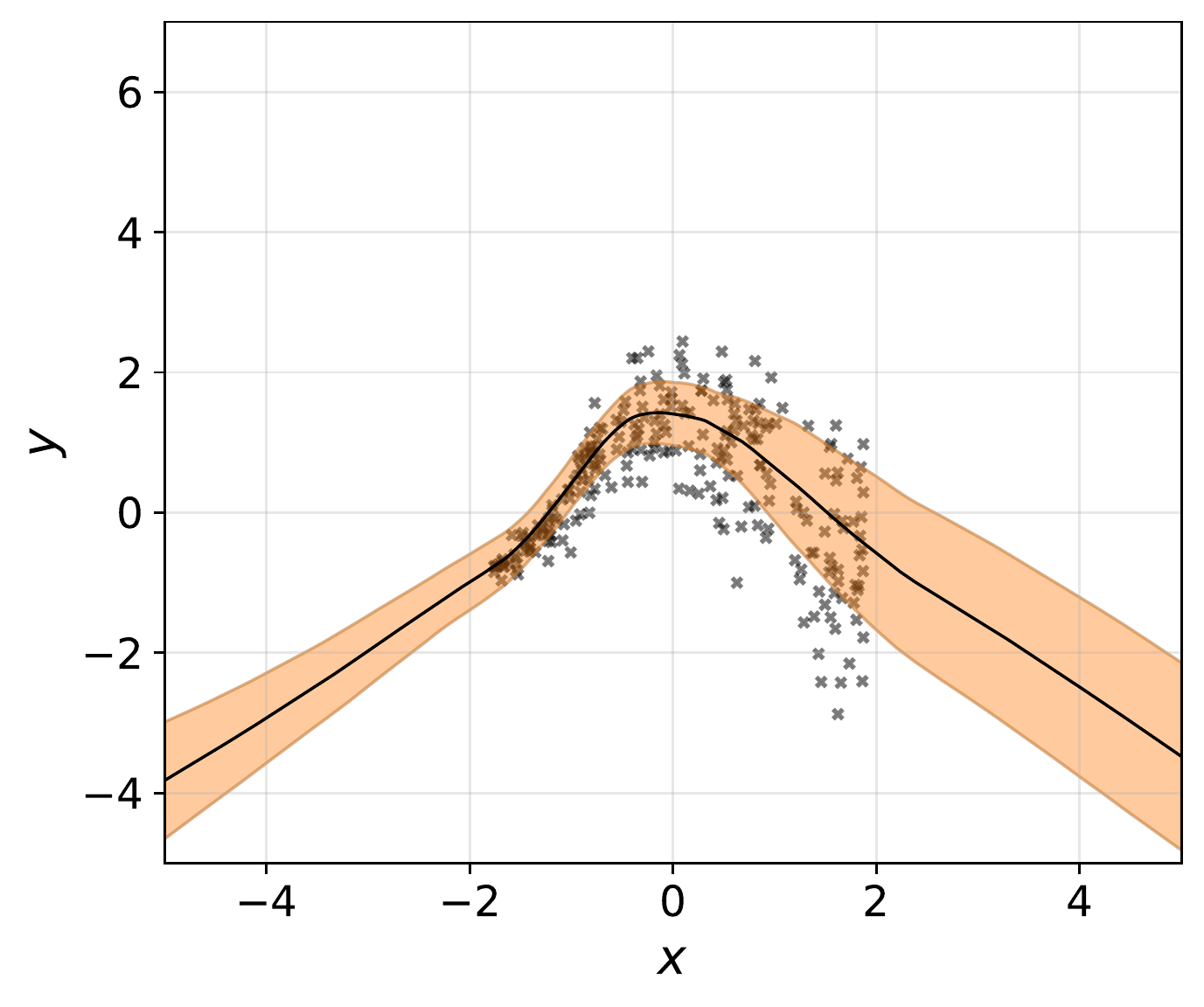}
\endminipage
\hfill
\minipage{0.3\textwidth}%
\includegraphics[width=\linewidth,height=0.75\linewidth]{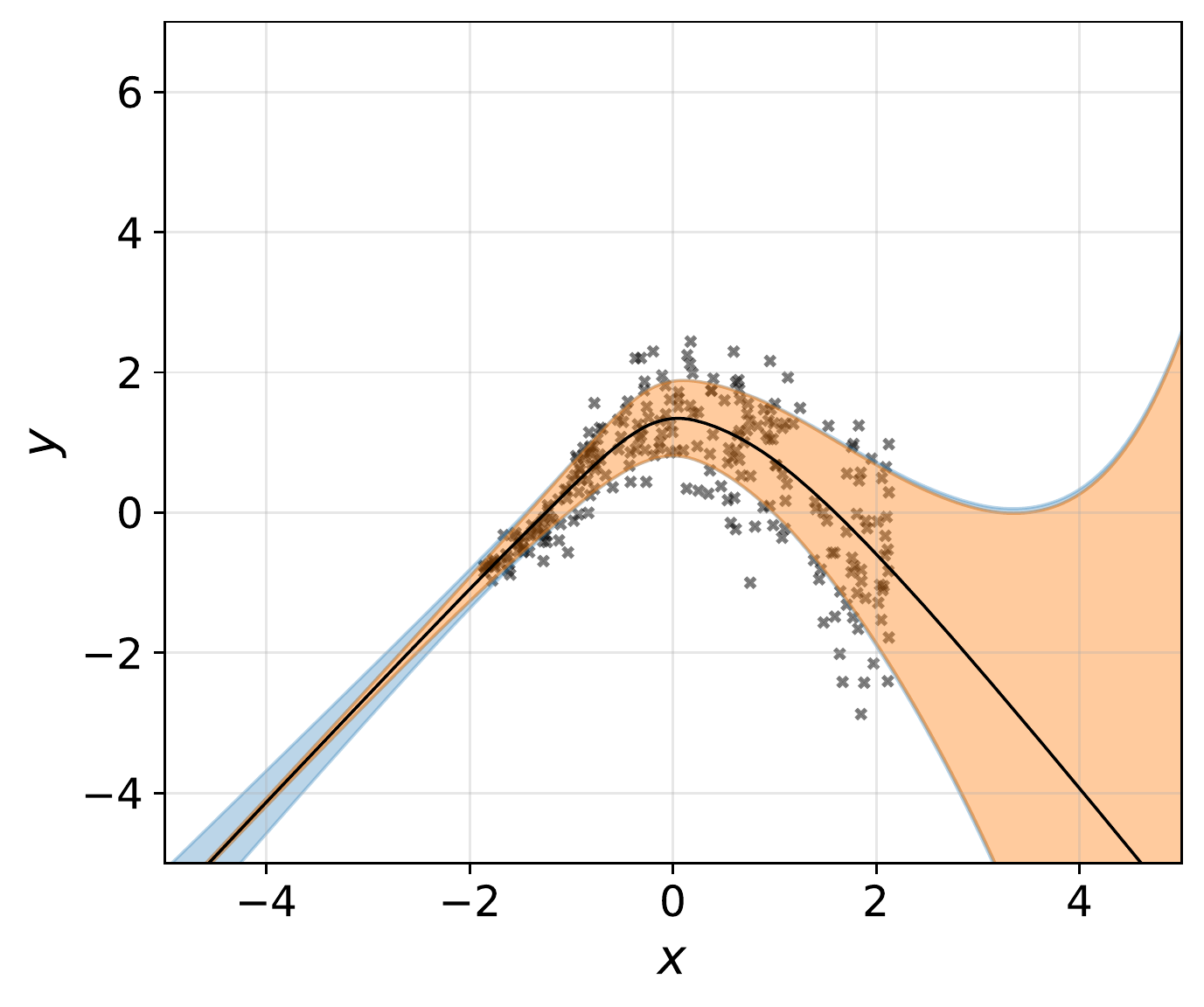}
\endminipage
\hfill
\minipage{0.3\textwidth}%
\includegraphics[width=\linewidth,height=0.75\linewidth]{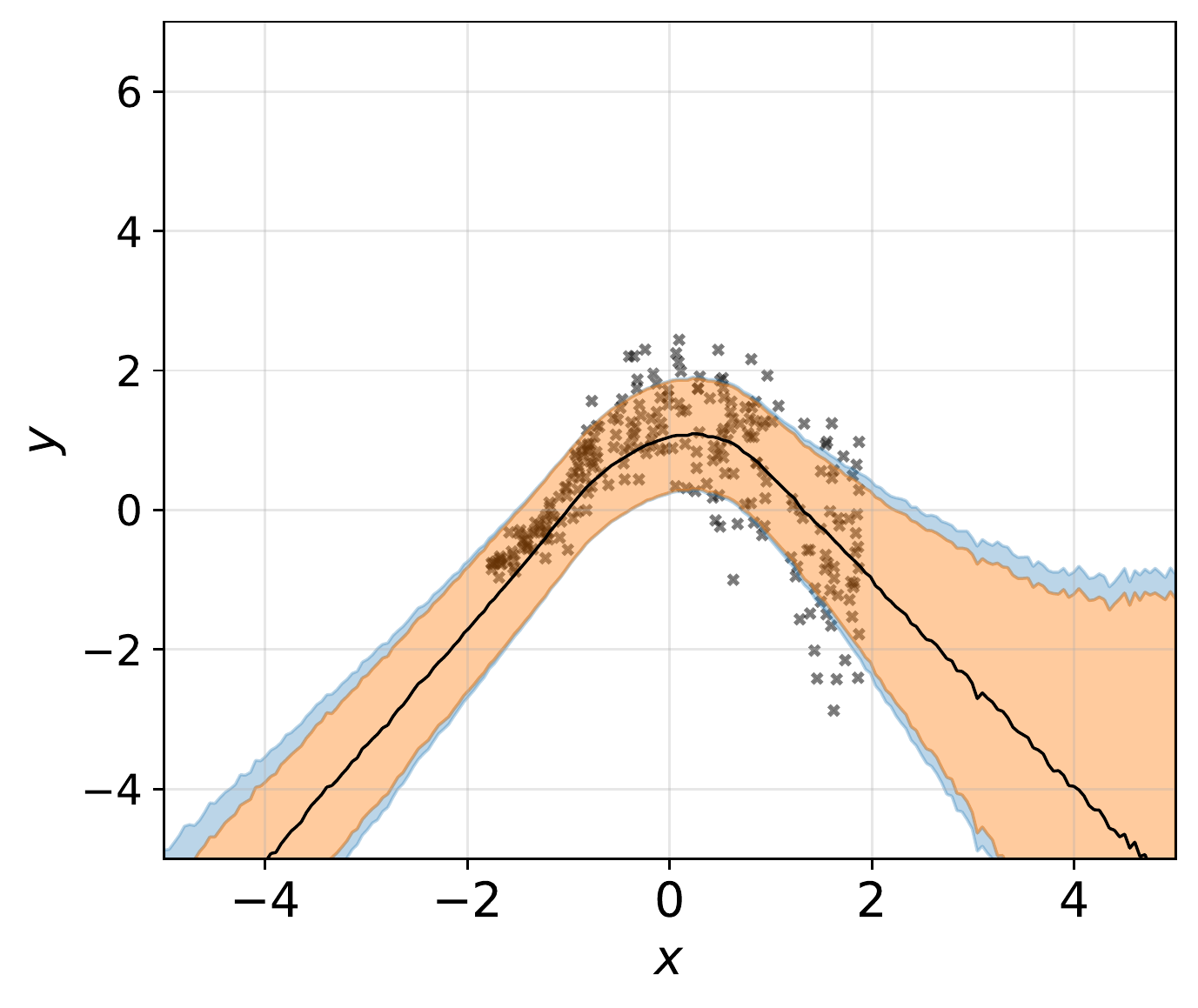}
\endminipage
\vfill
\minipage{0.3\textwidth}%
\includegraphics[width=\linewidth,height=0.75\linewidth]{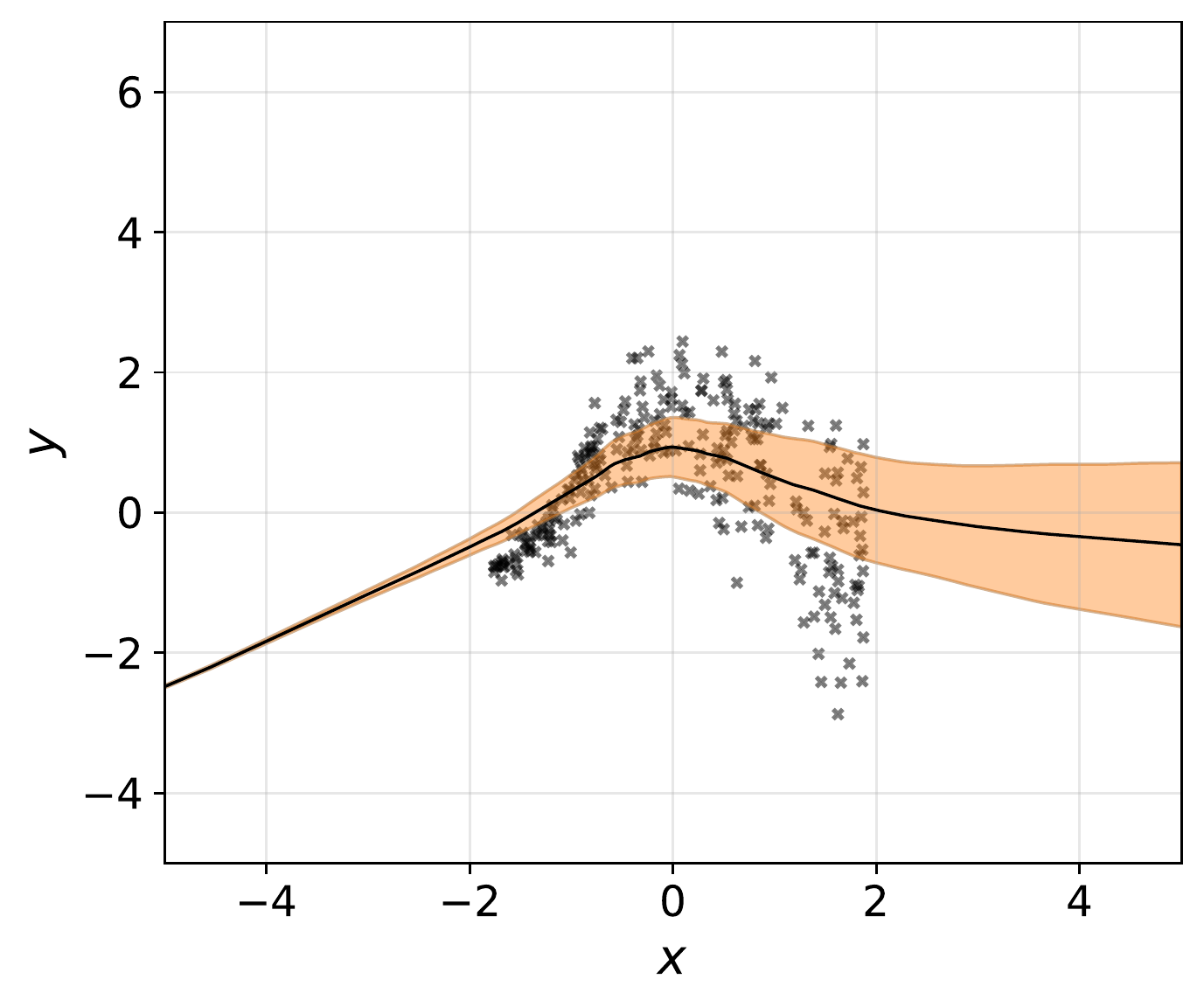}
\endminipage
\hfill
\minipage{0.3\textwidth}%
\includegraphics[width=\linewidth,height=0.75\linewidth]{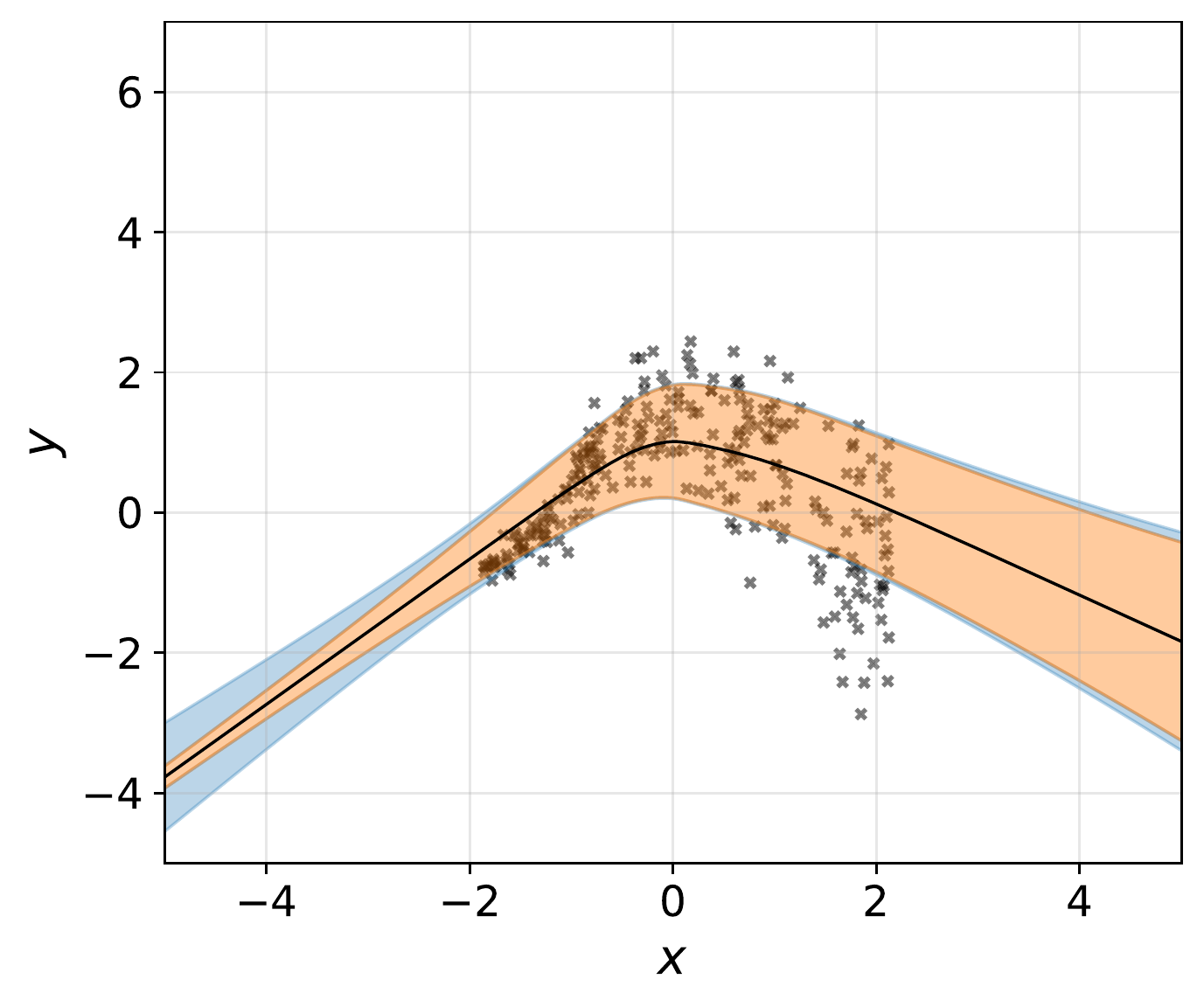}
\endminipage
\hfill
\minipage{0.3\textwidth}%
\includegraphics[width=\linewidth,height=0.75\linewidth]{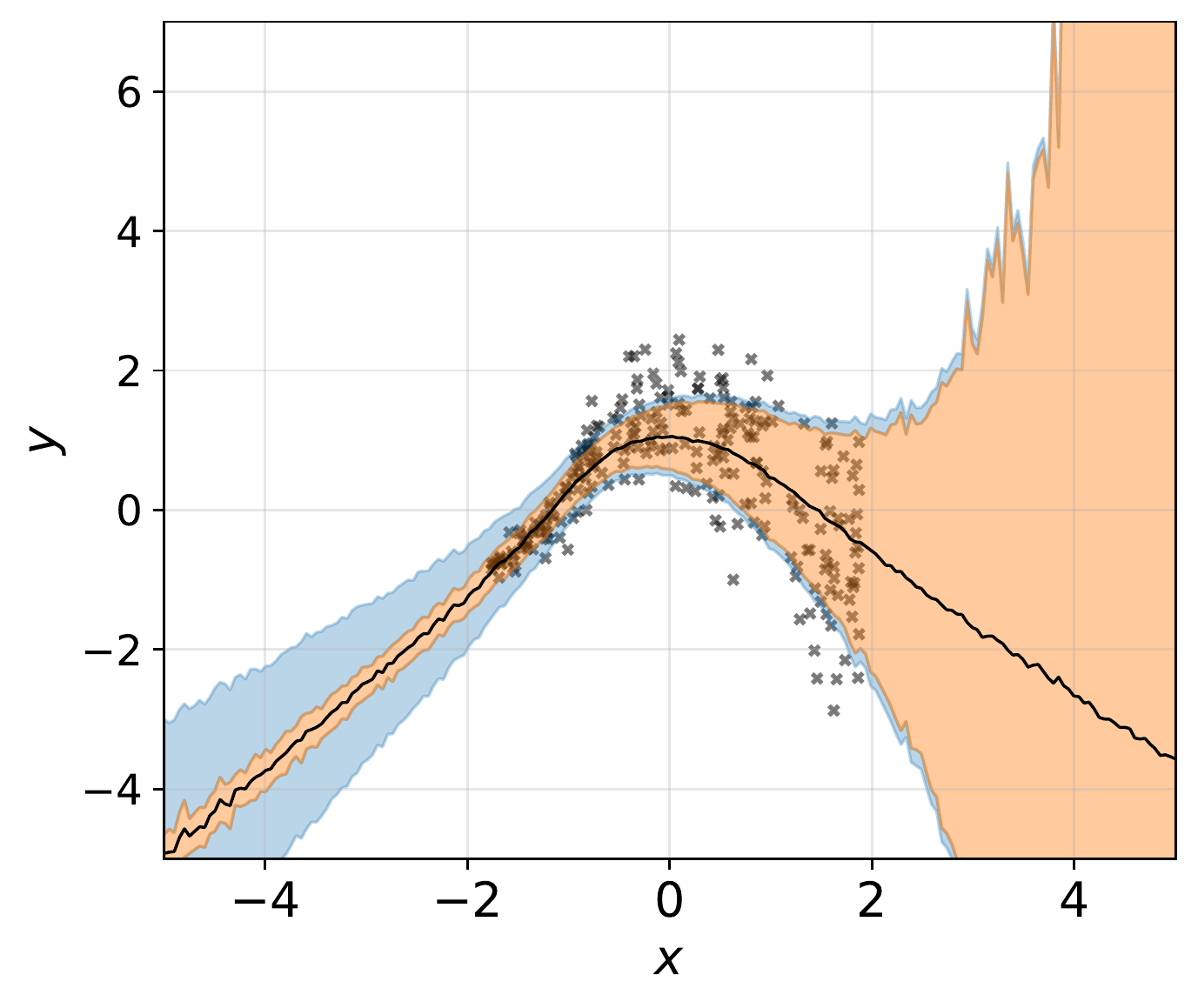}
\endminipage
\vspace{-0.3cm}
\caption{Prediction uncertainty on heteroscedasticity regression with Gaussian priors. Left to right: SGLD, BBP, MC Dropout. Upper: non-DP BNNs. Lower: DP-BNNs. Orange region refers to the posterior uncertainty. Blue region refers to the data uncertainty. Black line is the mean prediction.}
\label{regression}
\vspace{-0.3cm}
\end{figure}


\section{Discussion} \label{sec6:discussion}
This work proposes three DP-BNNs, namely DP-SGLD, DP-BBP and DP-MC Dropout, to both quantify the model uncertainty and guarantee the privacy in deep learning. All three DP-BNNs are evaluated through multiple metrics and demonstrate their advantages and limitations, supported by both theoretical and empirical analyses. For instance, as a sampling method, DP-SGLD can outperform the optimization methods, DP-BBP and DP-MC Dropout, on classification and regression tasks, at little expense of performance in comparison to the non-Bayesian or non-DP counterparts. However, DP-SGLD requires a possibly long period of burn-in to converge and its uncertainty quantification requires storing hundreds of weight iterates, making the method less scalable. 

We further empirically study the tradeoff between the privacy, the accuracy, and the reliability (via uncertainty quantification and calibration). In the regression experiment, DP-SGLD gives much more consistent uncertainty quantification, but BBP and MC Dropout can be largely affected by DP. In MNIST classification, DP tends to worsen the calibration yet the Bayesian methods (particularly the prior information) may reduce the calibration errors, as have been observed for the non-DP case \cite{maronas2018offline}.

Our work also provides valuable insights about the connection between the DP-SGLD, a method often applied in the Bayesian settings, and the DP-SGD, which is widely used without the consideration of Bayesian inference. This connection reveals novel findings about the impact of training hyperparameters on DP optimizers, e.g. larger batch size in fact enhances the privacy of DP-SGLD. Additionally, it bring attention to consider the DP-SGD as a Bayesian method to provide uncertainty quantification for free.

For future directions, it is of interest to extend the connection between DP-SGD and DP-SGLD to a more general class, DP-SG-MCMC (stochastic gradient Markov chain Monte Carlo), so as to accelerate the convergence of Bayesian gradient methods. Particularly, the convergence (especially the rate of convergence), the generalization, and the calibration behaviors of DP-BNNs needs more investigation from the theoretical viewpoint, similar to the analysis of DP linear regression \cite{wang2018revisiting} and DP deep learning \cite{bu2021convergence}.
\medskip
\nocite{*}
\bibliography{ref}

\clearpage
\appendix
\section{Background of Differential Privacy}
\label{app:DP background}

At the core of DP is the Gaussian mechanism which must work on functions with bounded $\ell_2$ sensitivity.
\begin{lemma}[Definition 3.8 \& Theorem 3.22 \cite{dwork2014algorithmic}]\label{lem:sensitivity and mechanism}
The \textbf{$ \ell_{2} $ sensitivity} of any function $ g $ is
$$\Delta g=\sup _{S, S'}\|g(S)-g(S')\|_{2}$$
where the supreme is over all pairs of neighboring datasets $(S,S')$. Consequently, the \textbf{Gaussian mechanism} which outputs 
$$\hat g(S) =g(S) +\sigma\Delta g\cdot\mathcal{N}(0,\mathbf{I})$$
is $(\epsilon,\delta)$-DP for some $\epsilon$ depending on $(\sigma,n,p,\delta)$, where the dependence is determined by the specific privacy accountant.
\end{lemma}

Notably, in the deep learning regime, the gradients may have unbounded sensitivity. Therefore, the per-sample clipping with a clipping norm $C$ defined \textit{ab initio} is applied to guarantee that the sensitivity of the sum of per-sample gradients is $C$.

As for the privacy accountant, we focus on two of the most popular privacy accountants, which may give different $\epsilon$'s for the same DP algorithm. In practice, we choose the smallest $\epsilon$ given by multiple privacy accountants as all are valid bounds of the true privacy loss. We remark that empirically GDP always give tighter $\epsilon$ than MA and GDP is known to be exact asymptotically. Furthermore, GDP's $\epsilon$ is explicit in terms of training parameters, yet MA, Fourier accountant \cite{koskela2020computing} and other methods \cite{gopi2021numerical,zhu2022optimal} require numerical integration to compute $\epsilon$. Thus the characterization is implicit and hard to analyze directly.

In this work, we apply the moments accountant (MA) \cite[Theorem 1 \& 2]{abadi2016deep} and GDP accountant \cite{bu2020deep}, both implemented efficiently in the \texttt{Tensorflow Privacy} library\footnote{See MA in \url{https://github.com/tensorflow/privacy/blob/master/tensorflow_privacy/privacy/analysis/rdp_accountant.py}; GDP in \url{https://github.com/tensorflow/privacy/blob/master/tensorflow_privacy/privacy/analysis/gdp_accountant.py}}. We remark that empirically GDP always give tighter $\epsilon$ than MA. Furthermore, GDP's $\epsilon$ is explicit in terms of training parameters, yet MA and Fourier accountant \cite{koskela2020computing} require numerical integral to compute $\epsilon$ and thus the characterization is implicit.

\section{Details of BBP and MC Dropout}
\label{app:detail BBP MC}
\subsection{BBP} \label{app:detail BBP}
To learn the hyperparameters, we minimize a KL divergence between the posterior distribution and the variational distribution, known as the `variational free energy' and its negative is the ELBO (see \url{http://krasserm.github.io/2019/03/14/bayesian-neural-networks/#appendix} for proof):
\begin{align}\label{problem:min KL append}
\min_\theta KL\left(q(w|\theta)\big\|p(w|D)\right).
\end{align}
We can rewrite this through the following optimization problem
\begin{align}\label{KL_div}
\min_{\theta} KL\left(q(w|\theta)\big\|p(w|D)\right)
&=KL\left(q(w|\theta)\big\|p(w)\right)-\mathbb{E}_{q(w|\theta)}\log p(D|w)
\\
&=\mathbb{E}_{q(w|\theta)}\log q(w|\theta)-\mathbb{E}_{q(w|\theta)}\log p(w)-\mathbb{E}_{q(w|\theta)}\log p(D|w)
\end{align}
$KL\left(q(w|\theta)\big\|p(w)\right)$ is called the `complexity cost', $\mathbb{E}_{q(w|\theta)}\log p(D|w)$ is called the `likelihood cost'. This objective function can hardly be calculated, though it can be approximated by drawing $w^{(j)}$ from $q(w|\theta)$. We thus define the optimization objective as
$$\frac{1}{N}\sum_{j=1}^N [\log q(w^{(j)}|\theta)-\log p(w^{(j)})-\log p(D|w^{(j)})]$$
where $N$ is the number of sampling. We further denote the summand, with $\ell$ being the loss, as $$\mathcal{L}_\text{BBP}(D;w^{(j)},\theta):=\log q(w^{(j)}|\theta)-\log p(w^{(j)})+\ell(D;\w^{(j)})$$

To learn $\theta$ from this objective, we borrow a transformation $\sigma= \log(1 + \exp(\rho))$ so that we can optimize on the unbounded $\rho$ without constraint, instead of the non-negative $\sigma$. Hence we sample from $q(w|\theta)$ by $w= \mu+\log(1 +\exp(\rho))\circ \epsilon$, where $\circ$ is Hadamard prodcut and $\epsilon\sim \mathcal{N}(0,I)$, and we optimize over $\theta=(\mu,\rho)$. The gradient of the objective with respect to the mean is 

$$
  \frac{d \mathcal{L}_\text{BBP}(D)}{d\mu}=\frac{1}{N}\sum_{j=1}^N\left(\frac{\partial \mathcal{L}_\text{BBP}(D;\w^{(j)}, \mu,\rho)}{\partial \w^{(j)}}+\frac{\partial \mathcal{L}_\text{BBP}(D;\w^{(j)}, \mu,\rho)}{\partial \mu}\right)
$$

The gradient with respect to the standard deviation term is 

$$
   \frac{d \mathcal{L}_\text{BBP}(D)}{d\rho}=\frac{1}{N}\sum_{j=1}^N\left(\frac{\partial \mathcal{L}_\text{BBP}(D;\w^{(j)}, \mu,\rho)}{\partial \w^{(j)}} \frac{\epsilon}{1+\exp (-\rho)}+\frac{\partial \mathcal{L}_\text{BBP}(D;\w^{(j)}, \mu,\rho)}{\partial \rho}\right)
$$

This objective leads to the SGD updating rule as
$$\mu_t = \mu_{t-1}-\frac{\eta_t}{|B_t|}\sum_{i\in B_t} \frac{d \mathcal{L}_\text{BBP}(\x_i,y_i)}{d\mu},
\quad\quad
\rho_t=\rho_{t-1}-\frac{\eta_t}{|B_t|}\sum_{i\in B_t}\frac{d \mathcal{L}_\text{BBP}(\x_i,y_i)}{d\rho}$$

\subsection{MC Dropout}
\label{app:more MC}
We review the MC Dropout \cite{gal2016dropout} for the case of a single hidden layer. Mathematically, any NN with dropout is approximated to a probabilistic Gaussian process model, which has a close relationship with BNN. We show this connection in the context of the regression problem.

Suppose the input $\mathbf{x_i}$ is an $1\times Q$ vector, the output $\mathbf{y_i}$ is an $1\times D$ vector and the hidden layer include $K$ units. We denote the two weight matrices by $\mathbf{W_1}$ and $\mathbf{W_2}$ which connect the first layer to the hidden layer and the hidden layer to the output layer respectively. $\sigma(\cdot)$ is some element-wise non-linear function such as RelU (rectified linear). $\mathbf{b}$ refers to the biases controlling the input location of each layer. Therefore, the output is $\hat{\mathbf{y}} = \sigma(\mathbf{x}\mathbf{W_1}+\mathbf{b})\mathbf{W_2}$. When applying dropout, we first sample two binary vectors $z_1=(z_{1,1},\dots,z_{1,Q})$ and $z_2=(z_{2,1},\dots,z_{2,K})$ with $z_{1,q}\sim$ Bernoulli($p_1$) for $q=1,\dots,Q$, $z_{2,k}\sim$ Bernoulli($p_2$) for $k=1,\dots,K$. Now, the output with dropout is given by $\hat{\mathbf{y}} = (\sigma((\mathbf{x}\circ z_1)\mathbf{W_1}+\mathbf{b})\circ z_2)\mathbf{W_2}$. It is mathematiclaly equivalent to $ \hat{\mathbf{y}} = \sigma(\mathbf{x}(z_1\mathbf{W_1})+\mathbf{b})(z_2\mathbf{W_2})$, which multiplies the weight matrices with the binary vector by row. 

For the regression problem, the NN model is often to optimize the following objective: 
\begin{equation}\label{ob-dropout}
     \mathcal{L}_\text{Dropout} = \frac{1}{2N} \sum_{i=1}^{N}\norm{y_i-\hat{y}_i}^2_2 + \lambda_1 \norm{\mathbf{W_1}}^2_2 + \lambda_2 \norm{\mathbf{W_2}}^2_2 + \lambda_3 \norm{\mathbf{b}}^2_2,
\end{equation}
with regularization parameters $\lambda_i$ for $i=1,2,3$. Now we would apply the Gaussian Process (GP) to the NN model described above to see why NN with dropout is equivalent to BNN. First, define the covariance function:
$$
K(\mathbf{x},\mathbf{y}) = \int p(w)p(b)\sigma(w^Tx+b)\sigma(w^Ty+b)dwdb
$$
with $1\times Q$ standard multivariate normal distribution $p(w)$  and some distribution $p(b)$. The Monte Carlo approximation to this covariance function is given by: 
$$
 \hat{K}(\mathbf{x},\mathbf{y}) = \frac{1}{K}\sum_{k=1}^{K} \sigma(w_k^Tx+b_k)\sigma(w_k^Ty+b_k)
$$
with $w_k\sim p(w)$ and $b_k\sim p(b)$. Therefore, our NN model with Gaussian process is equivalent to the following generative model: 
$$
\begin{aligned}
  \mathbf{W_1}=[w_k]_{k=1}^K , \mathbf{b} = [b_k]_{k=1}^K \\
  w_k\sim p(w), b_k \sim p(b) \\
 \mathbf{F}(\mathbf{X})| \mathbf{X}, \mathbf{W_1},\mathbf{b} \sim N(0,\hat{K}(\mathbf{X},\mathbf{X}))\\
 \mathbf{Y}|\mathbf{F}(\mathbf{X}) \sim N(\mathbf{F}(\mathbf{X}),\tau^{-1}\mathbf{I}_n),
\end{aligned}
$$
from which the predictive distribution is given by: 
$$
\begin{aligned}
   p(\mathbf{Y}|\mathbf{X}) &=&& \int p(\mathbf{Y}|\mathbf{F}(\mathbf{X}))p(\mathbf{F}(\mathbf{X})|\mathbf{W_1},\mathbf{b},\mathbf{X})p(\mathbf{W_1})p(\mathbf{b}) d\mathbf{W_1}d\mathbf{b}d\mathbf{F(\mathbf{X})} \\
   &=& & \int \mathcal{N}(\mathbf{Y};0,\Phi\Phi^T + \tau^{-1}\mathbf{I}_n)p(\mathbf{W_1})p(\mathbf{b}) d\mathbf{W_1}d\mathbf{b},
\end{aligned}
$$
where $\hat{K}(\mathbf{X},\mathbf{X})=\Phi\Phi^T $. The normal distribution of $\mathbf{Y}$ could be viewed as a joint normal distribution over the column of the $N\times D$ matrix $\mathbf{Y}$. In particular, we introduce a $K\times 1$ standard multivariate normal variable $w_d$ and each term in the joint distribution is:
$$
\mathcal{N}(y_d;0,\Phi\Phi^T + \tau^{-1}\mathbf{I}_n) = \int \mathcal{N}(y_d;\Phi w_d, \tau^{-1}\mathbf{I}_n) \mathcal{N}(w_d;0,\mathbf{I}_K) d w_d.
$$
Therefore, the predictive distribution could be written in the following way: 
\begin{equation}\label{pred_dist}
 p(\mathbf{Y}|\mathbf{X}) = \int p(\mathbf{Y}|\mathbf{X},\mathbf{W_1},\mathbf{W_2},\mathbf{b})p(\mathbf{W_1})p(\mathbf{W_2})p(\mathbf{b}) d\mathbf{W_1}d\mathbf{W_2}d\mathbf{b}, 
\end{equation}
where $\mathbf{W_2}=[w_d]_{d=1}^{D}$. Naturally, the expression \eqref{pred_dist} could be viewed as BNN with multivariate standard normal distributions on the weights. To connect with the idea of dropout, we proceed with the variational inference. Suppose we use the variational distribution $q(\mathbf{W_1},\mathbf{W_2},\mathbf{b})= q(\mathbf{W_1})q(\mathbf{W_2})q(\mathbf{b})$ to approximate the posterior distribution $p(\mathbf{W_1},\mathbf{W_2},\mathbf{b}|X,Y)$. In particular, the variational distribution on the weight matrix is factorized over the rows and each term is a mixture of normal distributions, one centered at $0$ and the other centered away from $0$: 
$$
\begin{aligned}
q(\mathbf{W_1}) & =&& \prod_{q=1}^{Q} q(\mathbf{w}_q),  \\
q(\mathbf{w}_q)  &= && p_1 \mathcal{N}(\mathbf{m}_{1,q},\sigma^2\mathbf{I}_k) + (1-p_1) \mathcal{N}(0,\sigma^2\mathbf{I}_k), 
\end{aligned}
$$
where $p_1$ refers to dropout rate of the first layer, $\sigma>0 $ and $\mathbf{m}_q \in \mathbb{R}^{K}$. A similar distribution is assigned to $\mathbf{W}_2$. And the variational distribution $q(\mathbf{b})$ of the bias  $\mathcal{N}(\mathbf{m}, \sigma^2\mathbf{I}_k)$. Given the definition above, $\mathbf{W_1}$ corresponds to a location matrix $\mathbf{M_i}= [\mathbf{m}_{1,1},\mathbf{m}_{1,2},\dots,\mathbf{m}_{1,Q}]^T$. Similar for $\mathbf{W_2}$. The variational Bayes is aimed to minimize the Kullback–Leibler (KL) divergence between the variational distribution and the posterior distriburion, which results in the following objective function:
\begin{equation}\label{dropout_bnn}
   \mathcal{L}_{bnn}(x_i;\hat{\w}^{(n)}_{t-1}) = \frac{1}{N}\sum_{n=1}^{N}\frac{-\log p(y_n|x_n,\hat{\w}_n)}{\tau} + \sum_{i=1}^{2}\frac{p_i l^2}{2\tau N}\norm{\mathbf{M_i}}^2_2 + \frac{1}{2\tau N}\norm{\mathbf{m}}^2_2,
\end{equation}
where $\hat{\w}_n$ is sampled from $q(\mathbf{W}_1)$. Compared to \Cref{pred_dist}, setting parameters appropriately will lead two optimization problems equivalent. More detailed explanation could be found in \cite{gal2016dropout}.

\section{Experiments}\label{appendix:experiment}
\subsection{Classifiction}
\begin{figure}[!htb]
    \centering
  \includegraphics[width=4cm]{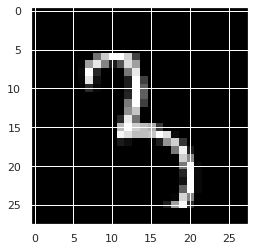}
  \hspace{1cm}
    \includegraphics[width=4cm]{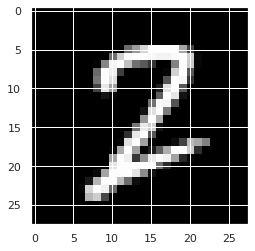}
    \caption{Left input image is 3, which is difficult to predict (low probability in the true class across three methods); Right input image is 2, which is easy to predict.}
    \label{Predicted_Labels}
\end{figure}
\begin{figure}[!htb]
\minipage{0.33\textwidth}
  \includegraphics[width=\linewidth]{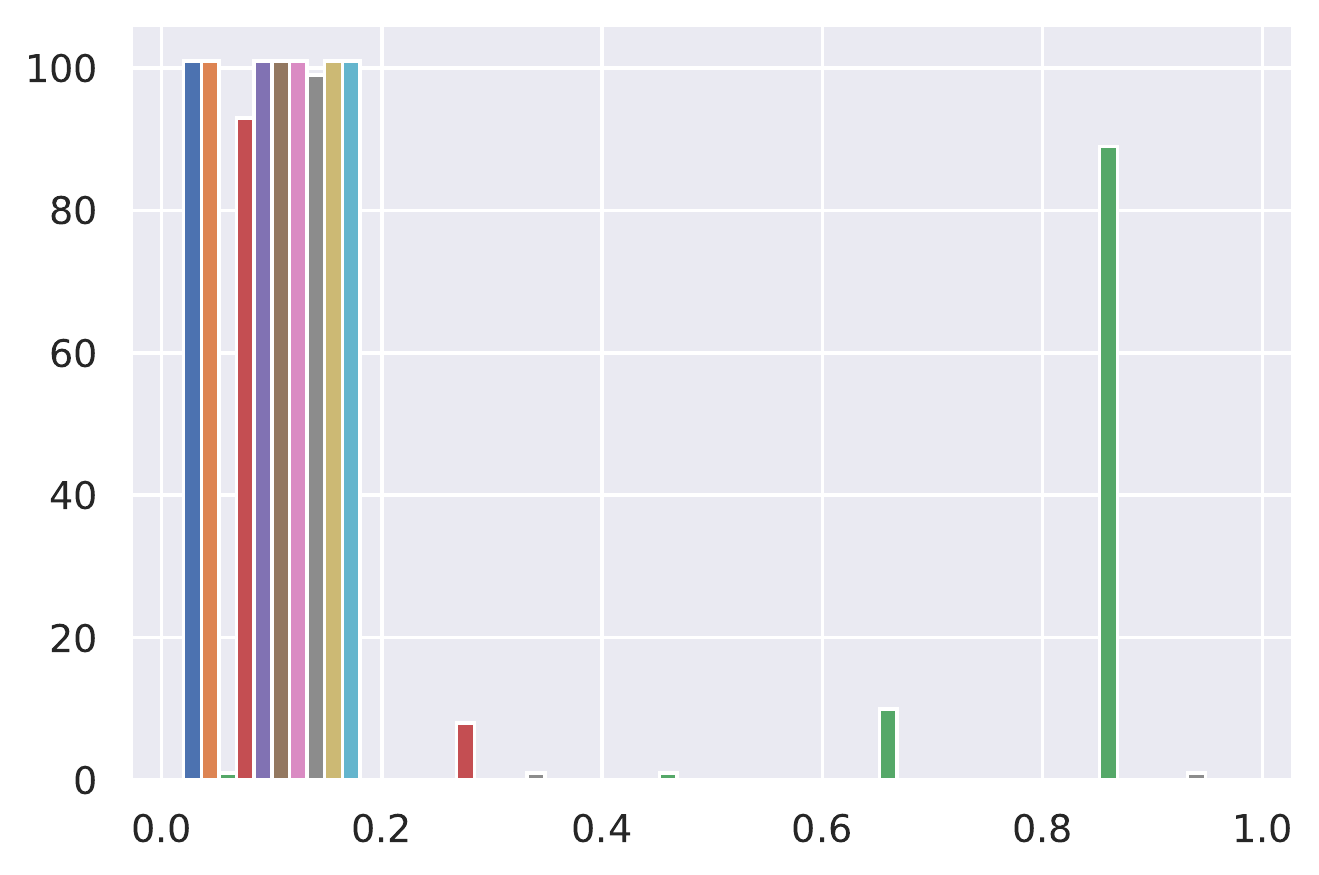}
\endminipage
\minipage{0.33\textwidth}%
  \includegraphics[width=\linewidth]{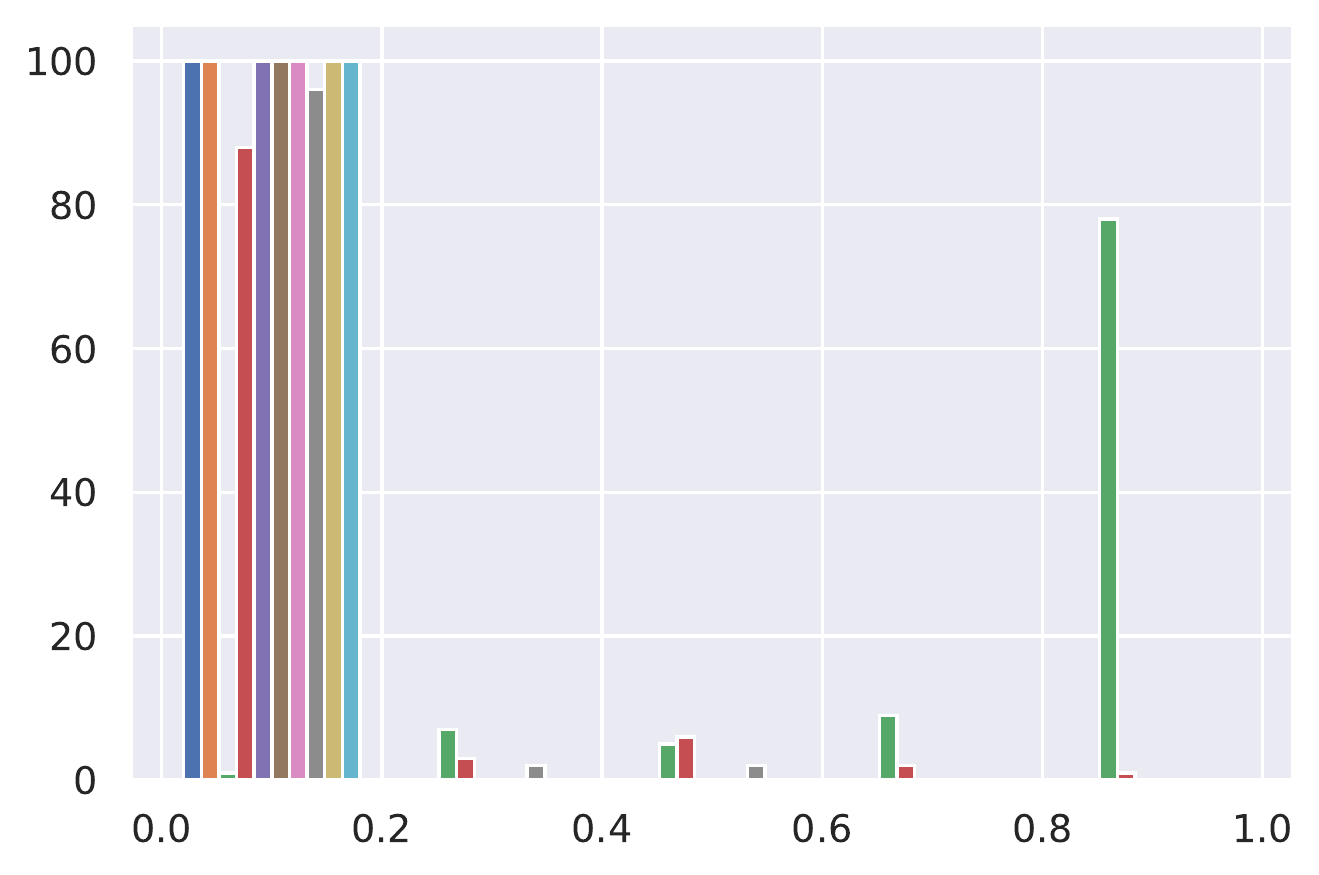}
\endminipage
\minipage{0.33\textwidth}%
  \includegraphics[width=\linewidth]{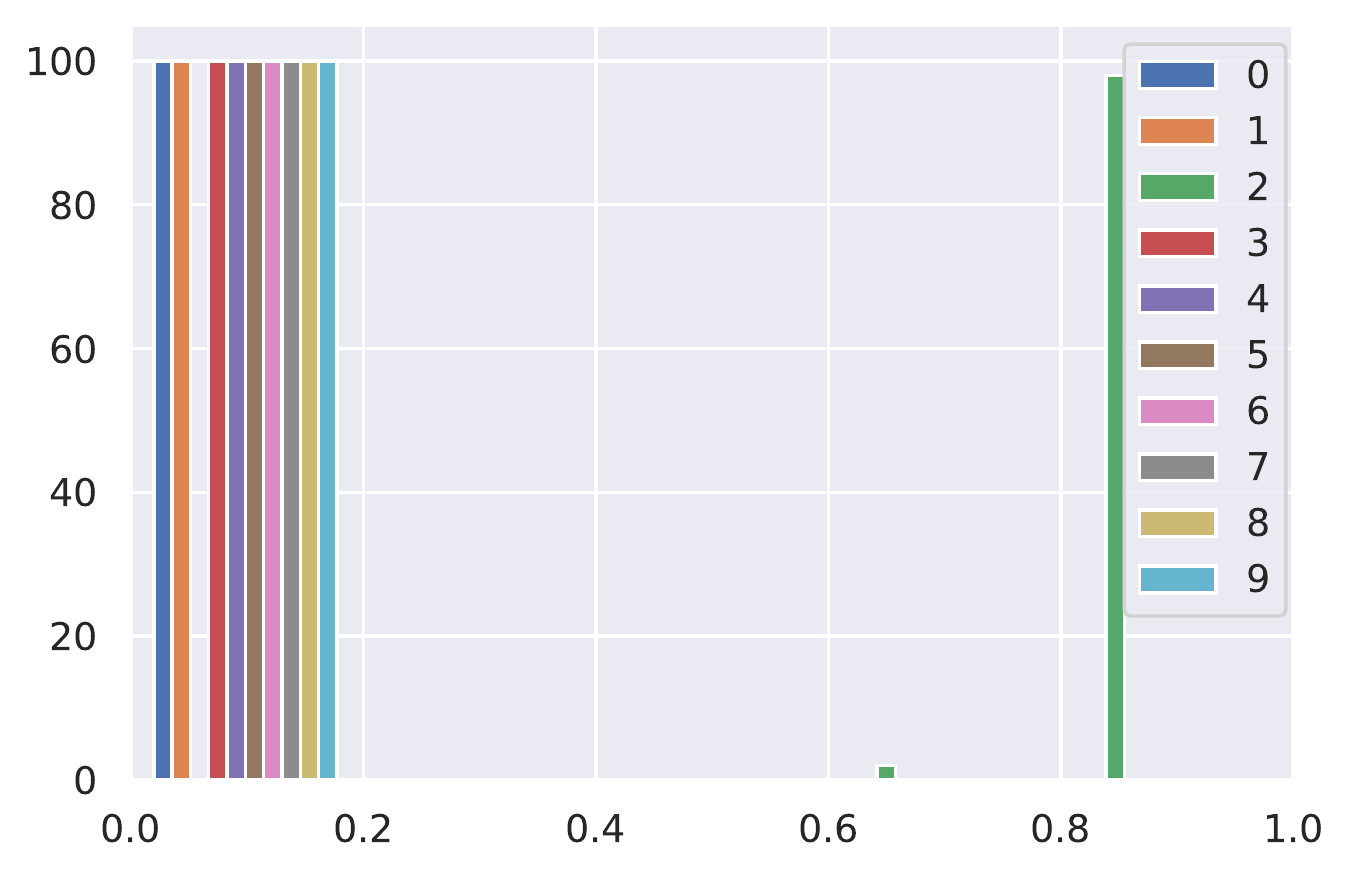}
\endminipage\vfill
\minipage{0.33\textwidth}
  \includegraphics[width=\linewidth]{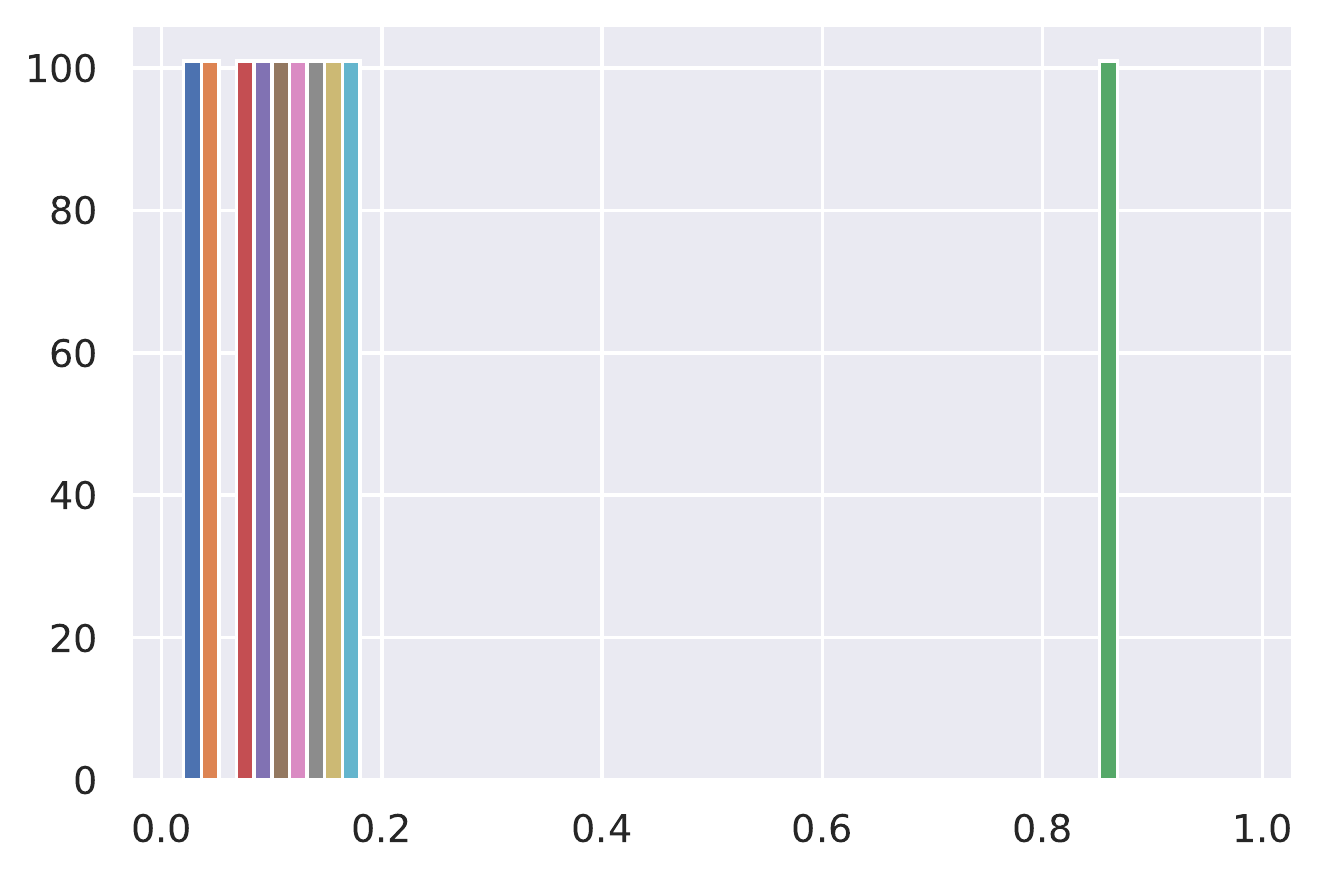}
\endminipage
\minipage{0.33\textwidth}%
  \includegraphics[width=\linewidth]{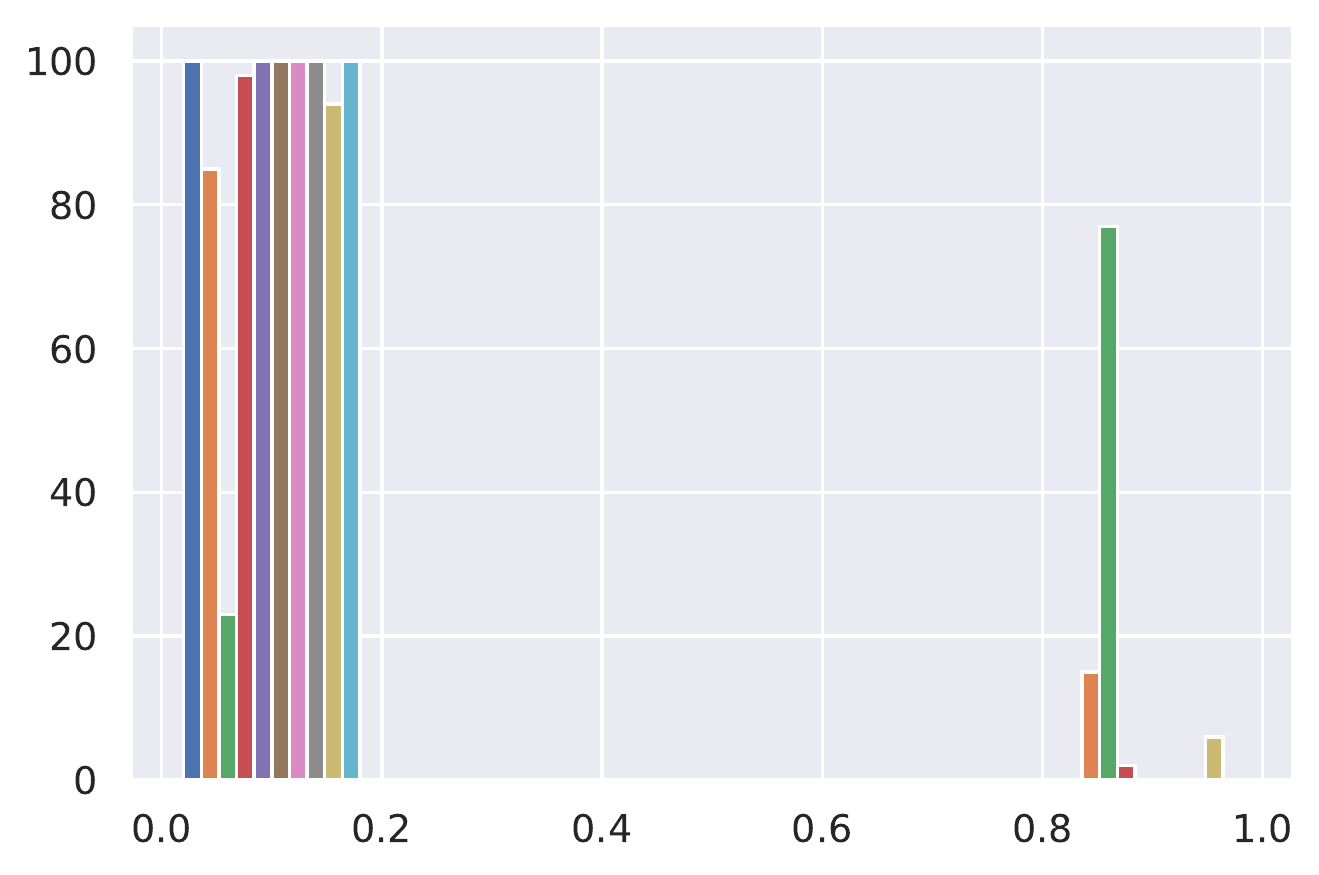}
\endminipage
\minipage{0.33\textwidth}%
  \includegraphics[width=\linewidth]{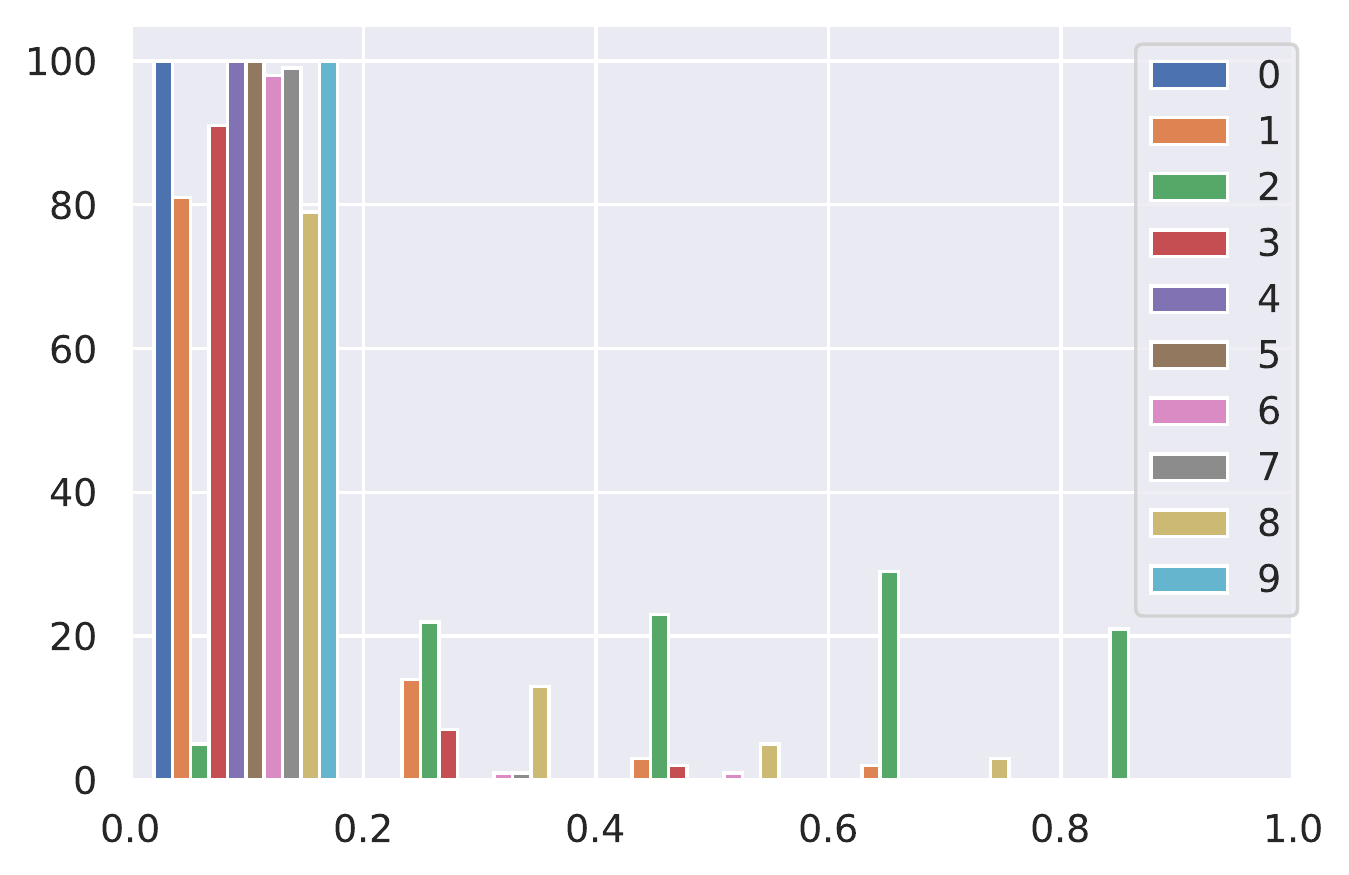}
\endminipage
\caption{y-axis refers to the frequency and x-axis refers to the prediction probability. Prediction distribution on MNIST over 100 repeated samplings (input image is $2$, see \Cref{Predicted_Labels}). Left to right: SGLD, BBP, MC Dropout. Upper: non-DP BNNs. Lower: DP-BNNs.}
\label{label_2}
\end{figure}

In \Cref{sec:classification}, we trained our algorithms on the MNIST digits dataset, in which each image is labelled with some number in between zero to nine. We consider two NNs with different architectures: multi-layer perceptron (MLP) and convolutional neural network (CNN). The softmax output layers have ten units, corresponding to possible labels. 

For MLP, there are two hidden layers, both containing 1200 units and activated by ReLU. For CNN, we use the benchmark architecture in \href{https://github.com/pytorch/opacus/blob/master/examples/mnist.py}{\texttt{Opacus}} and \href{https://github.com/tensorflow/privacy/blob/master/tutorials/walkthrough/mnist_scratch.py}{\texttt{Tensorflow Privacy}} libraries.

Given the specific structure of NN, DP-BBP and DP-SGLD assign either the Gaussian prior $\mathcal{N}(0,0.1^2)$ or the Laplacian prior $L(0,0.1)$ to each weight. For DP-MC Dropout, the dropout rate is $0.5$. After careful hyperparameter tuning, we select the following hyperparameters pairs (learning rate, batch size) for each method:  $(2\times 10^{-4}, 256)$ for DP-MC Dropout, $(0.25,256)$ for DP-BBP and $(5\times 10^{-6},256)$ for DP-SGLD. We set the privacy $\delta=10^{-5}$ and the clipping norm as $1.5$. For DP-BBP and DP-MC Dropout, the noise scale is 1.3. Experiments are conducted over $15$ epochs.

To show the prediction uncertainty in \Cref{label_3} and \Cref{label_2}, we select two digits (an easy-to-predict digit $2$ and a hard-to-predict digit $3$) from the test dataset, shown in Figure \ref{Predicted_Labels}. 

Fundamentally, the randomness of BNN comes from the weight uncertainty. For BBP, the weight uncertainty is from the posterior and we sample weights from their posterior distribution. For MC Dropout, the weight uncertainty is from the dropout and each time we randomly drop out the units of the trained NNs with the probability $0.5$. For SGLD, the weight uncertainty is from the weight updating rule and we record the last $100$ weight updates from the last epoch. Finally, for all methods, we collect the posterior probabilities of each class over $100$ independent predictions.

In \Cref{label_2}, we plot the empirical distribution of each class (denoted by different colors) over $100$ predictions. The $x$ axis refers to the predicted probability and the $y$ axis refers to the frequency. Similar as \Cref{label_3}, DP influences the posterior probability in different ways. However, even if three methods predict with different uncertainty, their predictions are correct no matter for DP or non-DP.

\subsection{Regression}
  For the heteroscedasticity regression problem in \Cref{sec:regression}, we used a network with two hidden layers of $200$ rectified linear units (ReLU), same as in the MNIST experiment. For both BBP and SGLD, we only consider the weights with the Gaussian prior. We generate $400$ data points, $250$ for training and $150$ for testing. Each input $x$ is sampled from the distribution $\text{Uniform}(-3,3)$,  while the output $\bm y = (y_1, y_2, \dots, y_{400})$ follows the multivariate normal distribution, whose covariance matrix is a function of $\bm x$. In the simulation study, this covariance matrix is the summation of the radial basis function kernel (RBF) with variance $1$, i.e. $\text{Kernel}(x,x')$ and a diagonal matrix whose diagonal element is $(0.3x+0.6)^2$. See the public notebook \url{https://github.com/JavierAntoran/Bayesian-Neural-Networks/blob/master/notebooks/regression/gp_homo_hetero.ipynb} for the code implementation. Notice that the output of the neural network has two elements: the prediction $\hat{y}_i$ and the noise estimation $\hat{\sigma}_i^2$.\footnote{See \url{https://github.com/JavierAntoran/Bayesian-Neural-Networks/blob/master/notebooks/regression/bbp_hetero.ipynb} for the network architecture.}
  
  In \Cref{regression}, we introduce two types of uncertainty. The data uncertainty is calculated by $\frac{1}{n}\sum_{i=1}^{n}\hat{\sigma}_i^2$. The posterior uncertainty is calculated by $ \dfrac{1}{n-1}\sum_{i=1}^n(\hat{y}_i - \bar{y})^2$, where $\bar{y} = \dfrac{1}{n}\sum_{i=1}^{n}\hat{y}_i$.
  
  \begin{itemize}
      \item MC Dropout: noise multiplier $\sigma=10$, clipping norm $C=2000$, learning rate 0.00005, dropout rate 0.5;
      \item SGLD: clipping norm $C=100$, learning rate 0.00025;
      \item BBP: noise multiplier $\sigma=10$, clipping norm $C=100$, learning rate 0.01;
  \end{itemize}

\subsection{Effects of batch size and learning rate}\label{app:experiment on training hyperparameter}
In \Cref{fig:dpsgld factors}, we empirically study the effects of batch size and learning rate on DP-SGLD and general DP-SGD. We use the standard DP CNN in \texttt{Opacus} library\footnote{See \url{https://github.com/pytorch/opacus/blob/master/examples/mnist.py}} and train with DP-SGD, which includes the DP-SGLD by \Cref{thm:dpsgld=dpsgd}. To be specific, for DP-SGLD, we set the number of epochs as 15, the clipping norm $C=1.5$, and the noise scale as $\sigma=\frac{|B|}{\sqrt{60000\times\eta}\times 1.5}$ in the DP-SGD; for general DP-SGD, we set the same number of epochs and clipping norm, but use a noise scale $\sigma=1.3$, which is the benchmark in \texttt{Opacus} and \texttt{Tensorflow Privacy} libraries, achieving around 95.0\% test accuracy with batch size 256.

When the batch size varies, we fix the learning rate at 0.25 for DP-SGD and 0.25/60000 for DP-SGLD; when the learning rate varies, we fix the batch size as 256.

\section{Additional Proofs}
\label{appendix: proof}

\subsection{Proof of \Cref{thm:dpsgld=dpsgd}} 
We start with stating the updating rules for both DP-SGD in \Cref{alg:DPSGD} and DP-SGLD in \Cref{alg:DPSGLD}. 

\begin{align*}
\w_t &= \w_{t-1}-\eta_t\left(\frac{1}{|B|}\sum_{i\in B} \widetilde{g}_i + \frac{\sigma \cdot C_t}{|B|}\cdot \mathcal{N}(0,I_d)+\nabla_{\w}r(\w_{t-1})\right), \quad \text{(DP-SGD)} \\
    \w_t &= \w_{t-1}-\eta_t\left(\frac{n}{|B|}\sum_{i\in B} \widetilde{g}_i + \frac{1}{\sqrt{\eta_t}}\mathcal{N}(0,I_d) +\nabla_{\w} r(\w_{t-1})\right), \quad \text{(DP-SGLD)} 
\end{align*}
where $|B|, C_t, \sigma, \widetilde{g}_i, \eta_t$ are defined in Section 2. By matching the coefficients of these updating rules, it is easy to see 
\begin{align*}
    \eta_\text{SGD} = \eta_\text{SGLD} \cdot n,  \qquad \eta_\text{SGD}\frac{\sigma_\text{SGD} C_\text{SGD}}{|B|} = \sqrt{\eta_\text{SGLD}}. 
\end{align*}
Additionally, the clipping is performed with the same gradient norm, hence $C_\text{SGLD}=C_\text{SGD}$. Therefore, we obtain
\begin{align*}
    &\textnormal{DP-SGLD}\left(\eta_\text{SGLD}=\eta,C_\text{SGLD}=C\right)=\textnormal{DP-SGD}\Big(\eta_\text{SGD} = \eta n, \sigma_\text{SGD} = \frac{\sqrt{\eta}|B|}{nC}, C_\text{SGD} = C \Big),
\\
&\textnormal{DP-SGD}\left(\eta_\text{SGD} = \eta, \sigma_\text{SGD} = \sigma, C_\text{SGD} = C \right)=\textnormal{DP-SGLD} \Big(\eta_\text{SGLD}= \frac{\eta}{n}, C_\text{SGLD} = C = \frac{|B|}{\sqrt{n\eta}\sigma}\Big).
\end{align*} 

\subsection{Proof of \Cref{thm:dpsgld privacy}}
Viewing DP-SGLD as DP-SGD, and plugging the coefficients $\sigma = \frac{\sqrt{\eta}|B|}{n  C}$ from \Cref{thm:dpsgld=dpsgd} into \Cref{thm:dpmc&dpbbp privacy}, we then get the desired result
$$
\sqrt{T(e^{1/\sigma^2}-1)}|B|/n = \sqrt{T(e^{n^2C^2/(\eta|B|^2)}-1)}|B|/n.
$$

\subsection{Larger batch size is more private in DP-SGLD}
From \Cref{thm:dpsgld privacy}, the privacy loss in GDP is $\sqrt{T(e^{n^2 C^2/(\eta|B|^2)}-1)}|B|/n$. Denoting $|B|/n$ as $0<x\leq 1$, we aim to show $\sqrt{T(e^{ C^2/\eta x^2}-1)}x$ is decreasing in $x$:
\begin{align*}
    \frac{d}{dx}\sqrt{T(e^{ C^2/\eta x^2}-1)}x 
    = \sqrt{T}\cdot\frac{\left(e^{C^2/\eta x^2} (1 - \frac{ C^2}{\eta x^2} ) - 1\right) }{\sqrt{(e^{C^2/\eta x^2}-1)}} \leq 0
\end{align*}
because $e^v (1 - v) - 1 \leq 0$ for any $v \in \mathbb{R}$. This fact can be checked by the derivative of $e^v (1 - v) - 1$ which indicates the only stationary point is $v=0$, and that $e^0(1-0)-1=0$. Hence $\sqrt{T(e^{n^2 C^2/\eta|B|^2}-1)}|B|/n$ is decreasing in $|B|$.

\subsection{Proof of \Cref{thm:dpsgld upperbound}}
In this section, we describe a road map to the convergence analysis of DP-SGLD. To be specific, we consider the following formula:
$$\w_t = \w_{t-1}-\eta_t\left(\frac{n}{|B|}\sum_{i\in B} \min\{1,\frac{C}{\|g_i\|_2+10^{-6}}\}\cdot g_i\right)+ \mathcal{N}(0,\eta_t).$$
In words, this DP-SGLD has no prior information and additionally incorporate a stability constant $10^{-6}$ as is implemented in the Opacus library\footnote{See line 400 of \url{https://github.com/pytorch/opacus/blob/main/opacus/optimizers/optimizer.py}.}.

We first view DP-SGLD as a specific case of DP-SGD through \Cref{thm:dpsgld=dpsgd}. This correspondence changes the hyperparamters from $(\eta, C)$ to $\eta_\textnormal{SGD} = \eta n, C_\textnormal{SGD}=C$ and  $\sigma_\textnormal{SGD} = \frac{\sqrt{\eta}|B|}{nC}$, where the subscript `SGD' refers to the hyperparamters of DP-SGD. 

Secondly, under \Cref{aspt: clipping happens} that all per-sample gradients are indeed clipped, it is obvious that
$$\min\left\{1,\frac{C}{\|g_i\|_2+10^{-6}}\right\}=\frac{C}{\|g_i\|_2+10^{-6}}, \forall i.$$

Thirdly, our DP-SGD reduces to the automatic DP-SGD \cite{bu2022automatic} (with learning rate $\eta_\textnormal{SGD}C$ and the same noise multiplier $\sigma_\textnormal{SGD}$). We show this reduction through \cite[Theorem 1]{bu2022automatic} which is restated as \Cref{lem:sgd vs auto sgd}.

Fourthly, it has been shown by \cite[Theorem 4]{bu2022automatic} and restated in \Cref{thm:dpsgd upperbound} that such automatic DP-SGD converges at an asymptotic rate $O(T^{-1/4})$ if
$$\eta_\textnormal{SGD}C_\textnormal{SGD}=\sqrt{\frac{\mathcal{L}_0}{TL(1+\sigma_\textnormal{SGD}^2 d/|B|^2)}}$$
which, in terms of the hyperparameters in DP-SGLD, can be written as
\begin{align}
\eta n C = \sqrt{\frac{\mathcal{L}_0}{TL(1+\frac{\eta d}{n^2 C^2})}}.
\label{eq:constraint to use automatic clipping}
\end{align}
This is further equivalent to
\begin{align}
\eta^2 n^2 C^2 TL + \eta^3 dTL = \mathcal{L}_0
\label{eq:what}
\end{align}
whose solution is the learning rate stated in our \Cref{thm:dpsgld upperbound}. Notice that the left two terms in \eqref{eq:what} must be positive and finite. The second term requires $\eta = O(T^{-1/3})$. Combined with first term, we obtain that $\eta C=O(T^{-1/2})$, which allows us to set $C=C_0 T^{-1/6}$ for some non-negative constant $C_0$. Now that the premise of \Cref{thm:dpsgd upperbound} is satisfied, we can leverage it to write
\begin{align}
\mathcal{X}=\frac{4}{\sqrt{T}}\sqrt{\mathcal{L}_0L\bigg( 1+ \frac{\sigma_\textnormal{SGD}^2 d}{|B|^2}\bigg)}= \frac{4}{\sqrt{T}}\sqrt{\mathcal{L}_0L\bigg( 1+ \frac{\eta d}{n^2 C^2}\bigg)} =  \frac{4\mathcal{L}_0}{T\eta n C},
\end{align}
in which the last equation follows from \eqref{eq:constraint to use automatic clipping}.

Lastly, substituting $\eta C=O(T^{-1/2})$, we have
$\mathcal{X}=O(1/\sqrt{T})$. Done.

\begin{lemma}[Theorem 1 in \citep{bu2022automatic}]
\label{lem:sgd vs auto sgd}
Under \Cref{aspt: clipping happens}, DP-SGD with learning rate $\eta_{\text{SGD}}$ is equivalent to an automatic
DP-SGD with learning rate $\eta_{\text{auto-SGD}} = \eta_{\text{SGD}} C$. In other words, DP-SGD
$$
\w_{t+1} = \w_t - \frac{\eta_\textnormal{SGD}}{|B_t|}\bigg( \sum_{i\in B_t} C\cdot \frac{\frac{\partial l_i}{\partial \w_t} }{\|\frac{\partial l_i}{\partial \w_t}\|+10^{-6}}+\eta_\textnormal{SGD} C\cdot\mathcal{N}(0,I)\bigg).
$$
is equivalent to the automatic DP-SGD
$$
\w_{t+1} = \w_t - \frac{\eta_\textnormal{auto-SGD}}{|B_t|}\bigg( \sum_{i\in B_t}\cdot \frac{\frac{\partial l_i}{\partial \w_t} }{\|\frac{\partial l_i}{\partial \w_t}\|+10^{-6}}+\eta_\textnormal{auto-SGD}\cdot\mathcal{N}(0,I)\bigg).
$$
with $\eta_\textnormal{SGD}C\equiv\eta_\textnormal{auto-SGD}$.
\end{lemma}

\begin{lemma}[Theorem 4 in \citep{bu2022automatic}]
\label{thm:dpsgd upperbound}
Under \Cref{aspt: smooth}, \ref{aspt: gn}, running the automatic DP-SGD for $T$ iterations, with learning rate $\eta= \sqrt{\frac{\mathcal{L}_0}{TL(1+\sigma^2 d/|B|^2)}}$, number of parameters $d$, initial loss $\mathcal{L}_0$, batch size $|B|$, Lipschitz smoothness constant $L$ and noise multiplier $\sigma$, gives 
\begin{equation}
    \min_{0\leq t \leq T} \mathbb{E}(\norm{g_t}) \leq \mathcal{G} \bigg(\mathcal{X};\xi \bigg)=O(T^{-1/4}),
\end{equation}
where $\mathcal{X}=\frac{4}{\sqrt{T}}\sqrt{\mathcal{L}_0L\bigg( 1+ \frac{\sigma^2 d}{B^2}\bigg)}$ and $\mathcal{G}(\cdot;\xi)$ is some increasing and positive function.
\end{lemma}

\end{document}